%% file: preprint.tex
\documentclass[nodate]{lucid-twocol}

\input{preamble}

\definecolor{cvprblue}{rgb}{0.21,0.49,0.74}
\definecolor{chrxcolor}{RGB}{1, 140, 116}

\usepackage[pagebackref,breaklinks,colorlinks,allcolors=cvprblue]{hyperref}

\input{macros}

\title[TellusLoom: Weaving Knowledge into Synthetic Worlds]%
  {TellusLoom: Weaving Knowledge into Synthetic Worlds\\for Dense Prediction in Remote Sensing}

\author[1,2]{Jian Song}\email{jian.song@riken.jp}
\author[1,2]{Hongruixuan Chen}
\author[1,2]{Pengyu Dai}
\author[2]{Yuzeng Chen}
\author[1]{Yafei Ou}
\author[1,2]{Weihao Xuan}
\author[3]{Jiaqi Yang}
\author[4]{Yiming Zhang}
\author[1]{Junshi Xia}
\author[1,2]{Naoto Yokoya}\email{yokoya@k.u-tokyo.ac.jp}

\affil[1]{RIKEN}
\affil[2]{The University of Tokyo}
\affil[3]{University of Wisconsin--Madison}
\affil[4]{University of Texas at Austin}

\logo*{\includegraphics[height=8.4mm]{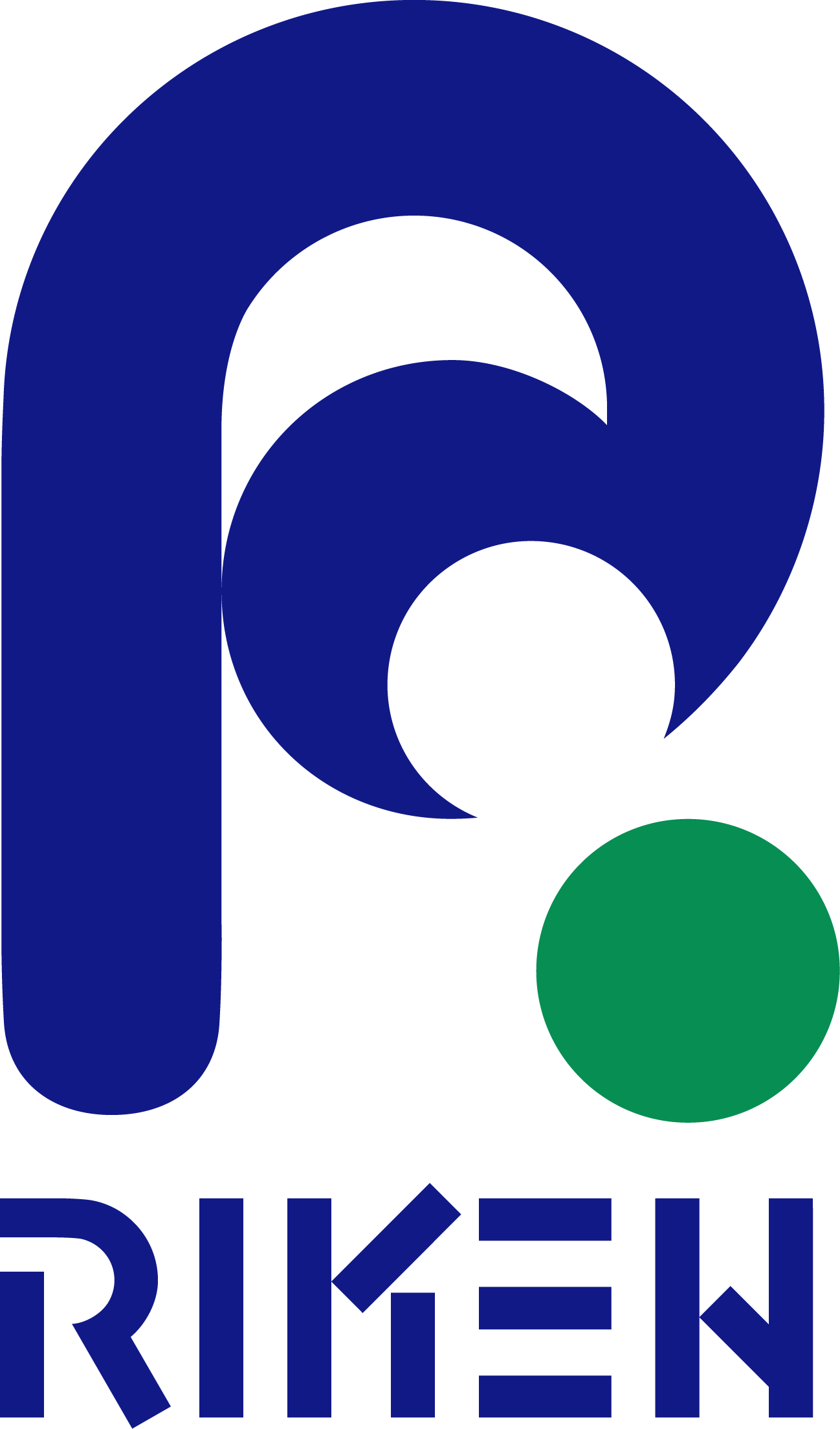}}
\logo*{\includegraphics[height=6.2mm]{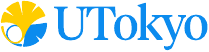}}

\input{supp_contents}
\makeatletter
\renewcommand\SuppTitleWithContents{%
  \newpage
  \twocolumn[{%
    \supplementhead
    \vskip \preprintsuppskip
    \edef\preprint@tocwd{\the\dimexpr\columnwidth-8pt\relax}%
    \begin{labpanel}[Contents]%
      \normalsize\normalfont\raggedright
      \setcounter{tocdepth}{2}%
      \renewcommand\@pnumwidth{2.2em}%
      \renewcommand\@tocrmarg{\dimexpr 2.55em-8pt\relax plus 1fil}%
      \renewcommand\@dotsep{6}%
      \hypersetup{linktoc=page}%
      \patchcmd\@dottedtocline{\leaders\hbox}{\color{labmuted!55!white}\leaders\hbox}{}{}%
      \patchcmd\@dottedtocline{\hfill \nobreak}{\hfill \normalcolor\nobreak}{}{}%
      \let\Supp@contentsline\contentsline
      \global\Supp@seccount=0 \global\Supp@lefttoks{}\global\Supp@righttoks{}%
      \def\contentsline##1##2##3##4{}%
      \def\SuppTocStart{\let\contentsline\Supp@collect}%
      \@starttoc{toc}%
      \noindent
      \begin{minipage}[t]{\preprint@tocwd}\the\Supp@lefttoks\end{minipage}\hfill
      \begin{minipage}[t]{\preprint@tocwd}\the\Supp@righttoks\end{minipage}\par
    \end{labpanel}%
    \vspace{\preprintsuppafter}%
  }]%
  \preprintpageheight{\preprintsuppextra}}
\newlength\preprintsuppskip 
\newlength\preprintsuppafter 
\newlength\preprintsuppextra 
\expandafter\let\expandafter\preprint@input\csname input\endcsname
\def\preprint@skip{supp_contents}
\long\expandafter\def\csname input\endcsname#1{\def\preprint@tmp{#1}\ifx\preprint@tmp\preprint@skip\else\preprint@input{#1}\fi}

\newcount\preprint@head \newcount\preprint@par \newcount\preprint@total \newcount\preprint@enl
\newcount\preprint@fhd \newcount\preprint@tol \newdimen\preprint@ems
\newif\ifpreprint@marks \newif\ifpreprint@loosened \newif\ifpreprint@main \newif\ifpreprint@stale \newif\ifpreprint@wasstale
\newbox\preprint@box
\newcommand\preprintparmarks{\preprint@markstrue}
\newcommand\preprintloose[2][1000]{\@namedef{preprint@loose@#2}{#1}}
\newcommand\preprintat[2]{\@namedef{preprint@at@#1}{#2}}
\newcommand\preprintfitfor[2]{\def\preprint@fitN{#1}\def\preprint@fitH{#2}}
\newcommand\preprintenlarge[2]{\@namedef{preprint@enl@#1}{#2}}
\let\preprintpageheight\enlargethispage
\renewcommand\enlargethispage[1]{\global\advance\preprint@enl\@ne
  \@ifundefined{preprint@enl@\the\preprint@enl}{\preprintpageheight{#1}}%
    {\expandafter\ifx\csname preprint@enl@\the\preprint@enl\endcsname\@empty\else
       \preprintpageheight{\@nameuse{preprint@enl@\the\preprint@enl}}\fi}}
\def\preprint@name{\the\preprint@head.\the\preprint@par}
\gdef\preprint@thisname{0.0}
\gdef\preprint@collist{}
\def\preprint@bib{thebibliography}
\let\preprint@logname\@gobble
\newcommand\preprintfitstale{\global\preprint@staletrue \global\preprint@wasstaletrue}
\@ifl@t@r\fmtversion{2021/06/01}{%
  \AddToHook{cmd/@startsection/before}{\global\advance\preprint@head\@ne \global\preprint@par\z@}
  \AddToHook{para/before}{\ifinner \global\preprint@mainfalse \else \global\preprint@maintrue \fi}
  \AddToHook{para/begin}{%
    \@ifundefined{@captype}{%
      \global\advance\preprint@par\@ne
      \ifpreprint@main \xdef\preprint@thisname{\preprint@name}\fi
      \preprint@logname{\preprint@name}%
      \ifpreprint@marks \rlap{\normalfont\tiny\color{red}[[\preprint@name]]}\fi
      \ifpreprint@stale\else
        \@ifundefined{preprint@loose@\preprint@name}{}%
          {\looseness=\@ne
           \global\preprint@loosenedtrue \preprint@fhd\finalhyphendemerits \preprint@tol\tolerance \preprint@ems\emergencystretch
           \finalhyphendemerits=1000000 \tolerance=\@nameuse{preprint@loose@\preprint@name}\relax \emergencystretch\z@}%
        \@ifundefined{preprint@at@\preprint@name}{}{\@nameuse{preprint@at@\preprint@name}}%
      \fi
    }{\preprint@logname{-}}}
  \AddToHook{para/after}{%
    \ifpreprint@loosened \global\preprint@loosenedfalse
      \finalhyphendemerits\preprint@fhd \tolerance\preprint@tol \emergencystretch\preprint@ems \fi
    \ifinner\else \@ifundefined{@captype}{%
      \global\advance\preprint@total\@ne
      \ifx\@currenvir\preprint@bib\else
        \xdef\preprint@collist{\preprint@collist\space\preprint@thisname:\the\prevgraf}\fi}{}\fi}
  \AtEndDocument{%
    \@ifundefined{preprint@fitN}{}{%
      \@tempswafalse
      \ifnum\preprint@total=\preprint@fitN\relax\else \@tempswatrue \fi
      \ifnum\preprint@head=\preprint@fitH\relax\else \@tempswatrue \fi
      \if@tempswa
        \immediate\write\@auxout{\string\providecommand\string\preprintfitstale{}\string\preprintfitstale}%
        \ifpreprint@wasstale
          \@latex@warning@no@line{preprint_fit.tex was made for another text (now \the\preprint@total\space paragraphs,
            \the\preprint@head\space headings) and is not applied: run scripts/preprint_fit.py}%
        \else \@latex@warning@no@line{The fit list does not belong to this text. Rerun to get the column feet without it}\fi
      \else
        \ifpreprint@wasstale \@latex@warning@no@line{The fit list belongs to this text again. Rerun to get the column feet right}\fi
      \fi}}
}{}
\@ifundefined{preprintfitrecords}{}{%
  \newwrite\preprint@cols \immediate\openout\preprint@cols=\jobname.cols
  \def\preprint@logname#1{\immediate\write\m@ne{PPFIT par #1}}%
  \AtBeginDocument{\hbadness=100 }%
  \let\preprint@makecol\@makecol
  \def\@makecol{%
    \setbox\preprint@box\vbox{\unvcopy\@cclv}%
    \dimen@\vsize \advance\dimen@-\ht\preprint@box
    \immediate\write\preprint@cols{col abs=\the\numexpr\ReadonlyShipoutCounter+1\relax\space
      first=\if@firstcolumn 1\else 0\fi\space gap=\the\dimen@\space pen=\the\outputpenalty\space
      special=\ifvbox\@kludgeins 1\else 0\fi\space two=\if@twocolumn 1\else 0\fi\space pars=\preprint@collist}%
    \gdef\preprint@collist{}%
    \preprint@makecol}
  \AtEndDocument{\immediate\write\preprint@cols{end paragraphs=\the\preprint@total\space headings=\the\preprint@head}}}
\makeatother
\input{preprint_fit}
\preprintenlarge{2}{}
\preprintat{49.1}{\ifnum\value{page}=12 \preprintpageheight{-204pt}\fi}

\begin{document}
\twocolumn[{%
\renewcommand\twocolumn[1][]{#1}%
\maketitle
\input{sections/1_teaser}
\input{sections/0_abstract}}]
\input{sections/1_intro}
\input{sections/2_related}
\input{sections/3_method}
\input{sections/4_protocol}
\input{sections/5_transfer}
\input{sections/7_properties}
\input{sections/7_conclusion}
\input{sections/8_llm_statement}
{
    \small
    \bibliographystyle{ieeenat_fullname}
    \bibliography{references}
}

\input{supp/supp}

\end{document}

%% file: preamble.tex
\usepackage[T1]{fontenc}
\usepackage{graphicx}
\usepackage{amsmath}
\usepackage{amssymb}
\usepackage{booktabs}
\usepackage{multirow}
\usepackage{xspace}
\usepackage{capt-of}
\usepackage{placeins}
\usepackage{adjustbox}
\usepackage{array}
\usepackage{tabularx}
\usepackage{makecell}
\usepackage{colortbl}
\usepackage{tikz}
\usetikzlibrary{arrows.meta,positioning,fit,calc,shapes.misc}

\newcommand{\subline}[1]{{\scriptsize\color{black!55}#1}}
\newcommand{\taskmetric}[2]{\makebox[2.95cm][l]{#1\hfill #2}}

%% file: macros.tex
\newcommand{\tl}{TellusLoom\xspace}
\newcommand{\tld}{TL-D\xspace}
\newcommand{\tlm}{TL-M\xspace}
\newcommand{\tlmix}{TL-Mix\xspace}
\newcommand{\photodr}{photometric randomization\xspace}
\newcommand{\uda}{UDA\xspace}
\newcommand{\NA}{{\color{gray}n/a}}
\definecolor{tlorange}{HTML}{EB6834}
\definecolor{mainhl}{RGB}{253,241,232}
\definecolor{gaingreen}{HTML}{1B7F3B}
\definecolor{realann}{RGB}{145,49,43}
\definecolor{unlabeled}{RGB}{52,88,108}
\definecolor{tableband}{gray}{0.96}
\definecolor{procband}{RGB}{233,242,236}\definecolor{procink}{RGB}{38,96,68}
\definecolor{diffband}{RGB}{236,236,248}\definecolor{diffink}{RGB}{66,62,140}
\definecolor{indom}{RGB}{99,99,99}
\newcommand{\INDOM}[1]{\textcolor{indom}{(#1\rlap{)}}}
\makeatletter
\newcommand{\reg}[1]{{\sffamily\fontsize{\strip@pt\dimexpr0.9\dimexpr\f@size pt\relax\relax}{\f@baselineskip}\selectfont #1}}
\makeatother

\newcommand{\odxw}{\textwidth}
\newcommand{\odxheadfive}{\makebox[0.0989\textwidth][c]{Image}\makebox[0.0989\textwidth][c]{Ground truth}\makebox[0.0989\textwidth][c]{Best OV}\makebox[0.0989\textwidth][c]{\textbf{Specialist}}\makebox[0.0989\textwidth][c]{\textbf{Unified}}}
\newcommand{\odxhead}{{\scriptsize\sffamily\makebox[\odxw]{\odxheadfive\hspace{0.011\textwidth}\odxheadfive}}\\[2pt]}
\newcommand{\odxlabel}[2]{{\footnotesize\sffamily\makebox[\odxw][l]{\textbf{#1}\quad best open-vocabulary method: #2}}\\[1pt]}

\newsavebox{\SuppTabBoxA}\newsavebox{\SuppTabBoxB}\newsavebox{\SuppTabBoxC}

\definecolor{rulep}{HTML}{E0A526}\definecolor{ruled}{HTML}{8B5CC4}\definecolor{rulec}{HTML}{3D6F2B}
\definecolor{ruletitle}{HTML}{F2F3F5}
\newcommand{\rulepill}[2]{\mbox{\setlength{\fboxsep}{1.3pt}\colorbox{#1!15}{\scriptsize\sffamily\textcolor{#1!68!black}{#2}}}}
\newcommand{\ruletag}[2]{\unskip\nobreak\hfill\penalty50\hskip0.8em\null\nobreak\hfill\rulepill{#1}{#2}\parfillskip=0pt\relax}
\newcommand{\ruleno}[1]{\textcolor{black!45}{\ttfamily\scriptsize #1}}
\newcommand{\mechP}{\rulepill{rulep}{P}}\newcommand{\mechD}{\rulepill{ruled}{D}}\newcommand{\mechC}{\rulepill{rulec}{C}}

%% file: preprint_fit.tex
\preprintfitfor{243}{95}
\preprintloose[5000]{3.1}
\preprintloose[5000]{9.3}
\preprintloose[5000]{8.3}
\preprintloose[5000]{6.1}
\preprintloose[5000]{15.4}
\preprintloose[5000]{22.1}
\preprintloose[5000]{21.1}
\preprintloose[5000]{20.1}
\preprintloose[5000]{73.1}
\preprintloose[5000]{88.2}
\preprintloose[5000]{93.1}

%% file: sections/1_teaser.tex
\begin{center}
\vspace{-4pt}
\includegraphics[width=\textwidth]{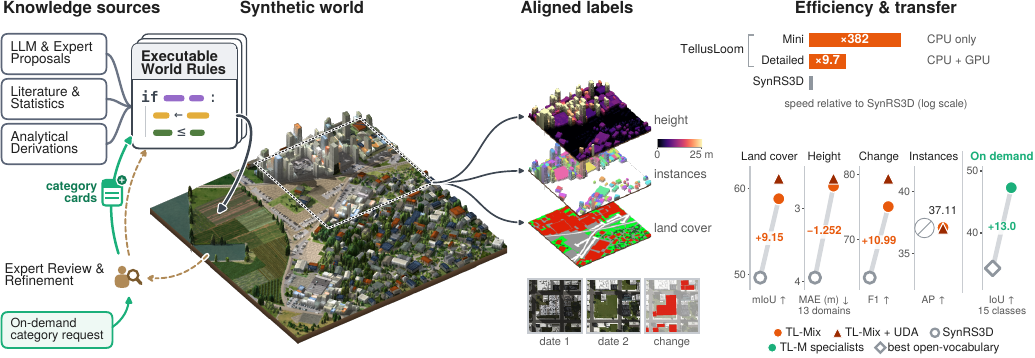}
\captionof{figure}{\textbf{From domain knowledge to transferable supervision.}
Through rule design and expert refinement, \tl formalizes domain knowledge into executable world rules to generate synthetic scenes with aligned labels for land cover, height, building instances, and change detection; category cards extend the rules on demand.
Right: generation throughput relative to SynRS3D~\cite{synrs3d}; transfer of \tlmix{} (an equal mix of the Detailed and Mini datasets) against SynRS3D, with and without adaptation on unlabeled target imagery; and \tlm specialists for 15 new classes against the best open-vocabulary method per category.
$\oslash$: SynRS3D has no instance labels.}
\label{fig:teaser}
\end{center}
\vspace{4pt}

%% file: sections/0_abstract.tex
\begin{abstract}
Dense prediction in very-high-resolution (VHR) optical remote sensing requires costly manual annotations.
Synthetic worlds provide scalable supervision, yet indiscriminate 3D asset and rendering complexity is costly and does not alone guarantee transfer to real imagery.
We introduce TellusLoom, a knowledge-driven framework that encodes domain knowledge as executable world rules---priors, conditional dependencies, and constraints---across five coupled dimensions of world design.
Two independent generators, the asset-rich TellusLoom-Detailed (\tld) and the lightweight, CPU-only TellusLoom-Mini (\tlm), generate imagery with aligned labels for land-cover mapping, height estimation, building change detection, and building instance segmentation, at approximately 9.7$\times$ and 382$\times$ the generation throughput of SynRS3D, respectively.
Their equal mixture (\tlmix) improves synthetic-only transfer over SynRS3D by 9.15~mIoU in land-cover mapping, reduces height mean absolute error (MAE) by 1.252~m, improves building-change F1 by 10.99 points, and reaches 37.11~mask AP for building instances; it also improves downstream performance when real training labels are scarce.
Beyond a fixed dataset, the throughput of \tlm turns procedural generation into an extensible and controllable platform.
\emph{Extensibility:} when new object categories are requested, lightweight category cards generate supervision on demand within hours, without manual 3D modeling or real labels; specialists for 15 such categories exceed the per-category best training-free open-vocabulary method by 13.0 mean IoU on iSAID.
\emph{Controllability:} rule ablations and targeted interventions show how executable rules, texture strength, scene-composition coverage, and observation geometry shape transfer.
Together, these findings favor preserving task-relevant world structure over indiscriminately scaling asset complexity.
Code and datasets will be released.
\end{abstract}

%% file: sections/1_intro.tex
\section{Introduction}
\label{sec:intro}

Dense prediction in very-high-resolution (VHR) optical remote sensing depends on costly annotations for land-cover mapping, height estimation, building change detection, and building instance segmentation.
Synthetic data offers scalable supervision, but its practical value hinges on how well models trained on it transfer to real imagery at manageable generation cost.

\begin{figure}[t]
\centering
\includegraphics[width=\columnwidth]{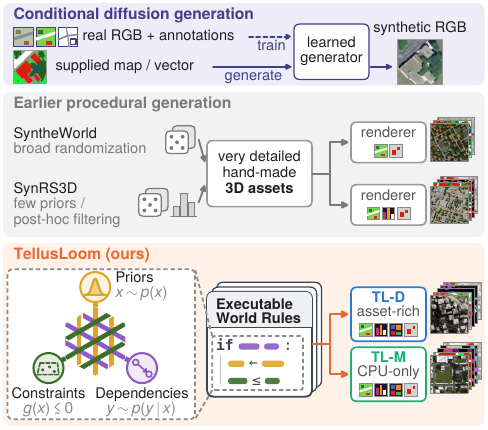}
\caption{\textbf{Rethinking remote sensing synthetic supervision.}
The evaluated conditional generative models rely on real RGB imagery paired with annotations or geospatial vectors to produce training-ready data.
Earlier procedural generators avoid this dependence but rely on elaborate 3D assets and relatively coarse world rules.
\tl instead emphasizes a rich library of coupled, executable rules, reducing reliance on detailed assets while generating aligned supervision through two independent generators.}
\label{fig:positioning}
\end{figure}

Two main routes have emerged (\cref{fig:positioning}).
\textbf{Conditional generative models} synthesize imagery from supplied maps or spatial conditions~\cite{seg2sat,changen2,vectorsynth}, serving as learned data augmentation but retaining a dependence on real RGB imagery paired with annotations or geospatial vectors.
\textbf{Procedural generators}, in contrast, construct virtual scenes and render imagery with aligned labels directly, without requiring real image--annotation pairs.
Earlier remote-sensing procedural generators progressed from broad randomization (SyntheWorld~\cite{syntheworld}) to real-world priors (SynRS3D~\cite{synrs3d}), yet rely on elaborate handcrafted assets and costly rendering while capturing only limited scene relationships.
This raises a central question: \textbf{Can executable world rules that encode task-relevant structure provide transferable supervision without indiscriminately scaling 3D asset and rendering complexity?}
Overhead imagery gives reason to expect so: seen from above, an object reveals only part of its 3D detail, whereas the image records how land uses, roads, parcels, and buildings are organized---regularities that can be written down as rules.

We introduce \textbf{TellusLoom}, a knowledge-driven procedural framework that prioritizes world design over asset complexity.
LLM and expert proposals, literature and statistics, and analytical derivations inform executable rules refined through expert review (\cref{sec:knowledge}; Supp.\ \cref{sec:s-world}).
These rules encode \textbf{priors} governing plausible distributions, \textbf{dependencies} relating variables to context, and \textbf{constraints} maintaining geometric and physical consistency across five coupled dimensions: composition and context, spatial organization, geometry and morphology, surface appearance, and observation formation (\cref{fig:method}).

To explore how much implementation complexity transfer actually requires, we realize this design in two independent generators (\cref{fig:teaser}): the asset-rich \textbf{TellusLoom-Detailed (\tld)} and the lightweight, CPU-only \textbf{TellusLoom-Mini (\tlm)}, which reach approximately 9.7$\times$ and 382$\times$ the generation throughput of SynRS3D, respectively.
Combining their datasets equally (\textbf{\tlmix}) improves synthetic-only transfer over SynRS3D across land-cover mapping (\ensuremath{+9.15}~mIoU), height estimation, and building change detection, while also boosting learning with scarce real labels.
The more expensive \tld does not consistently outperform \tlm across tasks and models.

Traditional procedural datasets are locked to fixed taxonomies because modeling 3D assets for new classes is slow, while training-free open-vocabulary foundation models~\cite{segearthov3,procsam3} often struggle with fine-grained overhead semantics.
Because \tlm decouples world rules from heavy 3D assets, it turns procedural generation into an \textbf{extensible and controllable platform}.
When novel object categories are requested, encoding their priors, dependencies, and constraints into lightweight \textbf{category cards} yields training data and a specialist model within hours on standard CPUs and a single GPU---without manual 3D modeling or any real labels---surpassing the per-category strongest training-free open-vocabulary baseline by 13.0 mean IoU across 15 iSAID~\cite{isaid} classes.
The same fast turnaround enables systematic rule ablations and targeted interventions on texture strength, scene composition, and observation geometry, revealing which properties of a synthetic world matter for transfer.

Our contributions are threefold:
\begin{enumerate}
\item \textbf{Knowledge-driven procedural world generation.} A framework of executable priors, dependencies, and constraints across five world dimensions, realized in two generators of contrasting complexity.
\item \textbf{Efficient, transferable, and open synthetic supervision.} Strong transfer across four tasks at up to 382$\times$ higher throughput, with code and datasets to be released.
\item \textbf{An extensible and controllable platform.} Rapid, label-free synthesis for newly requested categories via category cards ($+$13.0 mean IoU over the per-category best open-vocabulary method), and controlled studies of rules and scene properties to guide generator design.
\end{enumerate}

%% file: sections/2_related.tex
\section{Related work}
\label{sec:related}

\paragraph{Synthetic data generation for remote sensing.}
Conditional generative models synthesize overhead imagery from text, metadata, or spatial layouts~\cite{diffusionsat,geosynth,crsdiff}, including semantic maps or vectors (Seg2Sat~\cite{seg2sat}, EarthSynth~\cite{earthsynth}, VectorSynth~\cite{vectorsynth}), or synthesize bitemporal change data (ChangeAnywhere~\cite{changeanywhere}, Changen2~\cite{changen2}).
While effective for data augmentation, the conditional generators evaluated here rely on real RGB imagery paired with annotations or vectors for training, and some require real inputs at synthesis time.
Procedural datasets such as Synthinel~\cite{synthinel}, SyntheWorld~\cite{syntheworld}, and SynRS3D~\cite{synrs3d} instead render virtual scenes with aligned labels directly, avoiding real image--annotation pairs.
However, they typically depend on costly handcrafted 3D assets and fixed taxonomies while modeling relatively coarse scene relationships.
\tl retains annotation-free generation but shifts the focus from heavy assets to fine-grained, interconnected world rules.

\paragraph{Knowledge-guided procedural worlds.}
General-purpose frameworks such as Kubric~\cite{kubric} and Infinigen~\cite{infinigen} synthesize diverse 3D scenes with dense labels, while Holodeck~\cite{holodeck} and SceneCraft~\cite{scenecraft} introduce LLM-assisted spatial constraints and executable scene programs for indoor or object-centric environments.
Scaling procedural design to VHR remote sensing is particularly demanding: terrain, road networks, settlements, vegetation, surface materials, and imaging geometry interact across wide spatial scales.
\tl addresses this cross-scale coupling through executable priors, dependencies, and constraints.

\paragraph{Understanding synthetic-to-real transfer.}
Prior studies show that synthetic transfer depends critically on scene distributions rather than photorealism alone: structured domain randomization~\cite{sdr} and Meta-Sim~\cite{metasim} structure or learn scene distributions, while WinSyn~\cite{winsyn} examines geometry, materials, and rendering choices in building-window segmentation.
Yet it remains unclear which world properties justify higher generation cost in overhead imagery.
\tl pairs generation-throughput analysis with rule ablations and targeted interventions (\cref{sec:design}).

\paragraph{Open-vocabulary segmentation.}
Training-free open-vocabulary methods adapt vision--language and promptable foundation models~\cite{sam3} to segment novel overhead categories from text prompts alone~\cite{segearthov,segearthov3,procsam3,rosesam3}.
Because a class name alone conveys little of a category's overhead geometry, scale, and spatial context, \tl instead answers new category requests by synthesizing supervision on demand via category cards (\cref{sec:ondemand}).

%% file: sections/3_method.tex
\section{Knowledge-driven world generation}
\label{sec:method}

\begin{figure*}[t]
\centering
\includegraphics[width=\textwidth]{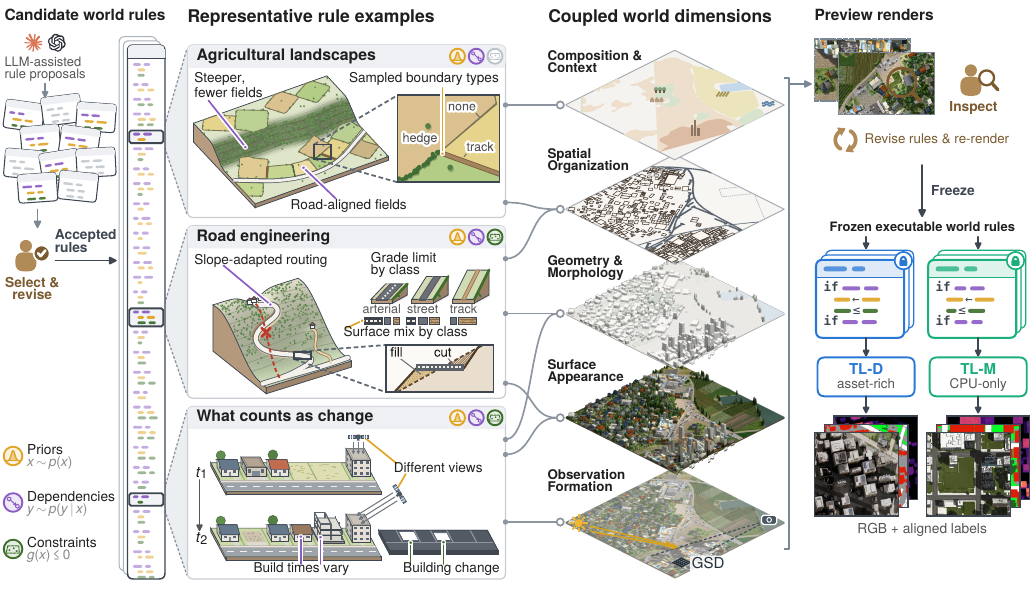}
\caption{\textbf{Representative rules couple world dimensions.} Three examples illustrate agricultural landscapes, road engineering, and building-change labeling. Colored icons denote priors, dependencies, and constraints, while links indicate how rules span five interconnected world dimensions. Experts inspect rendered previews and revise candidate rules, including LLM-assisted proposals, before freezing separate specifications for \tld and \tlm. Both produce RGB imagery with aligned task labels.}
\label{fig:method}
\end{figure*}

Generating coherent, large-scale VHR remote-sensing scenes requires governing not only individual objects but also their cross-scale interactions with surrounding environments and imaging geometry.
\tl formalizes this domain knowledge into interconnected, executable world rules (\cref{fig:method}), realized across two independent generators.

\subsection{Executable world rules}
\label{sec:structured-knowledge}

Rather than sampling scene parameters independently, \tl expresses domain knowledge through three coupled rule mechanisms: \textbf{priors} specify plausible property distributions; \textbf{conditional dependencies} tie variables to their spatial or semantic context; and \textbf{constraints} prevent or repair geometrically, physically, or semantically invalid configurations.
A single rule may combine all three mechanisms.

These rules weave together five world-design dimensions (\cref{fig:method}): \textbf{Composition \& Context} (scene types, land-use mixtures, object abundance); \textbf{Spatial Organization} (networks, parcels, adjacency, orientation); \textbf{Geometry \& Morphology} (terrain, footprints, heights, roofs, vegetation); \textbf{Surface Appearance} (materials, colors, texture scales); and \textbf{Observation Formation} (ground sampling distance, viewing angles, illumination, occlusion, and sensor effects).
Rules chain across these dimensions: road class restricts allowable grades, which govern how roads follow and cut or fill the terrain; the resulting road network orients adjacent agricultural fields, whose boundaries in turn organize hedgerows, tracks, and drainage ditches (\cref{sec:s-world}, \cref{tab:rulecards}).

\subsection{Rule grounding, refinement, and scale}
\label{sec:knowledge}

Constructing rules at this granularity requires complementary knowledge sources: LLM and expert proposals surface candidate variables and contextual dependencies; literature and public geospatial statistics anchor key distributions (such as building heights, tree height-to-crown allometry, and road-network statistics; \cref{sec:s-world}, \cref{tab:s-provenance}); and analytical derivations specify geometric, physical, and scale-dependent relations.
Most rules originate as proposals refined in expert review, with expert design where empirical measurements are unavailable (\cref{tab:s-sources}).
Candidate rules are iteratively refined through rendered previews: three remote sensing researchers (two postdoctoral researchers and one doctoral student, each with over three years of domain experience) inspect sampled scenes and aligned labels across ground sampling distances (GSDs), revising implausible layouts or observation geometries and selecting uncertain parameters via side-by-side comparisons.
Crucially, no downstream evaluation dataset is used during generator development; all rules and parameters are derived from the knowledge sources above, LLM cross-checks, and visual review before being frozen for production.

Counting each independent semantic decision as one rule (excluding parameters and helper code), \tlm comprises over 1\,000 executable rules and \tld over 500 (\cref{sec:s-rulescale}).
Priors, dependencies, and constraints populate all five dimensions, and roughly half of all rules depend on variables from a dimension other than the one they govern (51\,\% in \tlm, 48\,\% in \tld).
To cover diverse regions without overfitting to specific countries, rules are conditioned on regional archetypes---such as courtyard, self-built, or barrio settlement layouts, coastal, prairie, or semi-arid environment presets, and biome-specific vegetation ecoregions (\cref{sec:s-world}).
The same rule format also allows newly requested object categories to be specified on demand as plug-in category cards (\cref{sec:ondemand}).

\begin{table*}[t]
\caption{\textbf{Real-world transfer across tasks and generation paradigms (TEST).}
\emph{Real data}: $^t$, real images paired with annotations or geospatial vectors for generator training; $^g$, real images and corresponding annotations used during generation; $^u$, unlabeled real target images only, used for \uda.
S2L (S2Looking) evaluates large off-nadir views separately from the three-dataset building-change mean.
Parentheses indicate benchmarks used to train the generator (EarthSynth: OEM/LoveDA; Changen2-OEM: OEM).
($k$/13): domains better than SynRS3D.
Shading denotes \tl; bold marks the best procedural result per column.
--: unavailable.}
\label{tab:main}
\centering
\footnotesize
\setlength{\tabcolsep}{3.4pt}
\input{tables/main_tl}
\end{table*}

\subsection{Two generators and aligned supervision}
\label{sec:generators}
\label{sec:exporters}

To explore whether transferable world rules require heavy 3D assets, we instantiate our framework in two generators of contrasting complexity.
\textbf{TellusLoom-Detailed (\tld)} embeds fine-grained rules within an asset-rich Blender~\cite{blender} pipeline with GPU path tracing in Cycles, combining procedural terrain, road and river networks, biome-aware vegetation, and detailed city blocks and 3D models.
\textbf{TellusLoom-Mini (\tlm)} decouples rule execution from heavy 3D assets: all of its geometry is modeled procedurally, without pre-made 3D models, so that object shapes are themselves governed by rules and generation stays fast.
It implements its rule library using lightweight parametric representations in NumPy~\cite{numpy} and hybrid NumPy/Blender CPU rendering.
Following SyntheWorld~\cite{syntheworld} and SynRS3D~\cite{synrs3d}, both generators apply surface textures prepared offline with pretrained text-to-image models.

Because both generators project imagery and supervision from the same underlying 3D scene and camera state, task labels remain aligned even under oblique viewing angles and tall-structure occlusion.
Each scene yields eight-class OpenEarthMap land-cover maps~\cite{openearthmap}, above-ground height (nDSM), and visible building-instance masks.
For building change detection, simulated building-change events generate paired bitemporal scene states under date-specific illumination and viewing geometry, deriving change masks directly from visibility-filtered label passes.

%% file: tables/main_tl.tex
\renewcommand{\arraystretch}{1.0}
\begin{tabularx}{\textwidth}{>{\raggedright\arraybackslash}p{1.13in}>{\raggedright\arraybackslash}p{0.96in}>{\raggedleft\arraybackslash\hsize=0.8642\hsize\linewidth=\hsize}X>{\raggedleft\arraybackslash\hsize=0.8642\hsize\linewidth=\hsize}X>{\raggedleft\arraybackslash\hsize=1.3395\hsize\linewidth=\hsize}X>{\raggedleft\arraybackslash\hsize=0.8642\hsize\linewidth=\hsize}X>{\raggedleft\arraybackslash\hsize=0.8642\hsize\linewidth=\hsize}X>{\raggedleft\arraybackslash\hsize=0.8642\hsize\linewidth=\hsize}X>{\raggedleft\arraybackslash\hsize=1.3395\hsize\linewidth=\hsize}X}
\toprule
\multicolumn{2}{l}{\textbf{(a)}} & \multicolumn{3}{c}{\textbf{Land cover}\enspace mIoU\,$\uparrow$} & \multicolumn{4}{c}{\textbf{Building instances}\enspace mask AP\,$\uparrow$} \\
\cmidrule(lr){3-5}\cmidrule(l){6-9}
Training source & Real data & OEM & LoveDA & \textbf{Mean}\,$\pm$\,SD & xBD & WHU & BRIGHT & \textbf{Mean}\,$\pm$\,SD \\
\midrule
\multicolumn{9}{l}{\cellcolor{procband}\textcolor{procink}{\textbf{Procedural world generation}}} \\
SyntheWorld~\cite{syntheworld} & \textcolor{indom}{No real annot.} & 37.36 & 37.84 & \mbox{37.60\,{\scriptsize$\pm$0.66}} & -- & -- & -- & -- \\
SynRS3D~\cite{synrs3d} & \textcolor{indom}{No real annot.} & 47.09 & 52.18 & \mbox{49.64\,{\scriptsize$\pm$1.00}} & -- & -- & -- & -- \\
\addlinespace[2pt]
\rowcolor{mainhl}\textbf{\tld} & \textcolor{indom}{No real annot.} & 53.95 & 54.22 & \mbox{54.08\,{\scriptsize$\pm$0.41}} & \textbf{28.77} & 56.09 & 26.33 & \mbox{37.07\,{\scriptsize$\pm$0.13}} \\
\rowcolor{mainhl}\textbf{\tlm} & \textcolor{indom}{No real annot.} & 55.87 & 60.34 & \mbox{58.10\,{\scriptsize$\pm$0.22}} & 23.18 & 53.62 & 24.00 & \mbox{33.60\,{\scriptsize$\pm$0.56}} \\
\rowcolor{mainhl}\textbf{\tlmix} & \textcolor{indom}{No real annot.} & 57.47 & 60.11 & \mbox{58.79\,{\scriptsize$\pm$0.14}} & 27.82 & \textbf{56.78} & 26.74 & \mbox{\textbf{37.11}\,{\scriptsize$\pm$0.23}} \\
\rowcolor{mainhl}\textbf{\tlmix + \uda} & UDA$^{u}$ & \textbf{60.58} & \textbf{61.71} & \mbox{\textbf{61.15}\,{\scriptsize$\pm$0.09}} & 27.99 & 55.82 & \textbf{27.29} & \mbox{37.03\,{\scriptsize$\pm$0.11}} \\
\multicolumn{9}{l}{\cellcolor{diffband}\textcolor{diffink}{\textbf{Diffusion-based image generation}}} \\
Seg2Sat~\cite{seg2sat} & FLAIR~\cite{flair1}$^{t}$ & 27.31 & 26.71 & \mbox{27.01\,{\scriptsize$\pm$0.94}} & -- & -- & -- & -- \\
VectorSynth-COSA~\cite{vectorsynth} & RGB/OSM~\cite{osm}$^{t}$ & 15.58 & 14.90 & \mbox{15.24\,{\scriptsize$\pm$0.57}} & -- & -- & -- & -- \\
EarthSynth~\cite{earthsynth} & OEM, LoveDA, \dots$^{t}$ & \INDOM{24.48} & \INDOM{23.14} & \mbox{\INDOM{23.81\,{\scriptsize$\pm$0.76}}} & -- & -- & -- & -- \\
Changen2-OEM~\cite{changen2} & OEM$^{t}$ & \INDOM{61.33} & 56.51 & \mbox{\INDOM{58.92\,{\scriptsize$\pm$0.04}}} & -- & -- & -- & -- \\
Changen2-OEM + \uda & OEM$^{t}$, UDA$^{u}$ & \INDOM{64.19} & 58.19 & \mbox{\INDOM{61.19\,{\scriptsize$\pm$0.06}}} & -- & -- & -- & -- \\
\bottomrule
\end{tabularx}

\vspace{5pt}

\begin{tabularx}{\textwidth}{>{\raggedright\arraybackslash}p{1.13in}>{\raggedright\arraybackslash}p{0.96in}>{\raggedleft\arraybackslash\hsize=1.6500\hsize\linewidth=\hsize}X>{\raggedleft\arraybackslash\hsize=0.6750\hsize\linewidth=\hsize}X>{\raggedleft\arraybackslash\hsize=0.6750\hsize\linewidth=\hsize}X>{\raggedleft\arraybackslash\hsize=0.7875\hsize\linewidth=\hsize}X>{\raggedleft\arraybackslash\hsize=1.0500\hsize\linewidth=\hsize}X>{\raggedleft\arraybackslash\hsize=1.1625\hsize\linewidth=\hsize}X}
\toprule
\multicolumn{2}{l}{\textbf{(b)}} & \multicolumn{1}{c}{\textbf{Height}\enspace MAE (m)\,$\downarrow$} & \multicolumn{5}{c}{\textbf{Building change}\enspace F1\,$\uparrow$} \\
\cmidrule(lr){3-3}\cmidrule(l){4-8}
Training source & Real data & \multicolumn{1}{c}{13 domains} & LEVIR & WHU & SECOND & \textbf{Mean}\,$\pm$\,SD & S2L (off-nadir) \\
\midrule
\multicolumn{8}{l}{\cellcolor{procband}\textcolor{procink}{\textbf{Procedural world generation}}} \\
SyntheWorld~\cite{syntheworld} & \textcolor{indom}{No real annot.} & --\phantom{\,{\scriptsize(13/13)}} & 77.19 & 69.87 & 56.42 & \mbox{67.82\,{\scriptsize$\pm$2.44}} & 39.73 \\
SynRS3D~\cite{synrs3d} & \textcolor{indom}{No real annot.} & \mbox{3.945\,{\scriptsize$\pm$0.043}\phantom{\,{\scriptsize(13/13)}}} & 59.33 & 71.42 & 61.47 & \mbox{64.08\,{\scriptsize$\pm$3.03}} & 36.96 \\
\addlinespace[2pt]
\rowcolor{mainhl}\textbf{\tld} & \textcolor{indom}{No real annot.} & \mbox{2.867\,{\scriptsize$\pm$0.028}\,{\scriptsize(12/13)}} & 77.71 & 70.70 & 70.37 & \mbox{72.93\,{\scriptsize$\pm$0.33}} & 50.81 \\
\rowcolor{mainhl}\textbf{\tlm} & \textcolor{indom}{No real annot.} & \mbox{2.899\,{\scriptsize$\pm$0.007}\,{\scriptsize(13/13)}} & 66.92 & 77.47 & 61.75 & \mbox{68.72\,{\scriptsize$\pm$1.78}} & 47.73 \\
\rowcolor{mainhl}\textbf{\tlmix} & \textcolor{indom}{No real annot.} & \mbox{2.693\,{\scriptsize$\pm$0.008}\,{\scriptsize(13/13)}} & 78.75 & 75.33 & 71.12 & \mbox{75.07\,{\scriptsize$\pm$1.90}} & 53.40 \\
\rowcolor{mainhl}\textbf{\tlmix + \uda} & UDA$^{u}$ & \mbox{\textbf{2.587}\,{\scriptsize$\pm$0.006}\,{\scriptsize(13/13)}} & \textbf{83.45} & \textbf{80.94} & \textbf{73.93} & \mbox{\textbf{79.44}\,{\scriptsize$\pm$0.54}} & \textbf{56.18} \\
\multicolumn{8}{l}{\cellcolor{diffband}\textcolor{diffink}{\textbf{Diffusion-based image generation}}} \\
Changen2 S1~\cite{changen2} & xView2~\cite{xbd}$^{t,g}$ & --\phantom{\,{\scriptsize(13/13)}} & 84.64 & 87.55 & 73.99 & \mbox{82.06\,{\scriptsize$\pm$0.10}} & 57.86 \\
Changen2 S9 & OEM$^{t,g}$ & --\phantom{\,{\scriptsize(13/13)}} & 83.87 & 89.91 & 71.36 & \mbox{81.71\,{\scriptsize$\pm$0.49}} & 54.65 \\
\bottomrule
\end{tabularx}

%% file: sections/4_protocol.tex
\section{Experimental setup}
\label{sec:protocol}
\begin{figure*}[t]
\centering
\includegraphics[width=\textwidth]{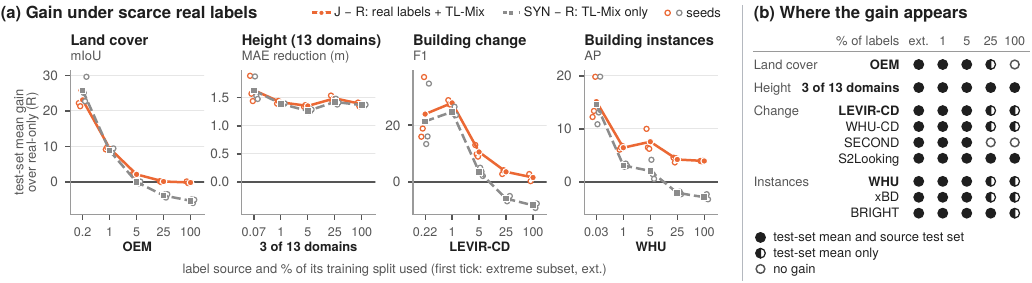}
\caption{\textbf{Learning with scarce real labels (TEST).}
(a) Test-set mean gain over real-only training R (\reg{REAL@p}) for joint training J (\reg{REAL@p+SYN}: the same labels plus \tlmix) and for \tlmix alone (\reg{SYN}).
Test sets as in \cref{tab:setup}.
Height gains are MAE reductions (m).
(b) All nine label sources (bold: shown in a): where J's mean gain over R is positive.
Protocol: \cref{sec:s-p3-config}.}
\label{fig:scarce}
\label{fig:fewlabels}
\end{figure*}
\begin{table}[t]
\caption{\textbf{Tasks, models, real test benchmarks, and metrics} (\cref{tab:main}, \cref{fig:scarce}).}
\label{tab:setup}
\centering
\footnotesize
\setlength{\tabcolsep}{0pt}
\begin{tabular*}{\columnwidth}{@{}l@{\hspace{5pt}}@{\extracolsep{\fill}}l@{}}
\toprule
\taskmetric{Task}{Metric} & Real evaluation benchmarks (TEST) \\[-1pt]
\subline{model} & \subline{details} \\
\midrule
\taskmetric{\textbf{Land cover}}{mIoU $\uparrow$} & OEM~\cite{openearthmap}, LoveDA~\cite{loveda} \\[-1pt]
\subline{DINOv3-L + DPT~\cite{dpt,dinov3}} & \subline{eight classes (OEM); six shared (LoveDA val)} \\
\addlinespace[1.5pt]
\taskmetric{\textbf{Height}}{MAE (m) $\downarrow$} & 13 geographic domains (\cref{tab:s-hdomains}) \\[-1pt]
\subline{DINOv3-L + DPT~\cite{dpt}} & \subline{6 from~\cite{song2026sparse}; labels in \cref{fig:scarce} from~\cite{isprs,dfc18}} \\
\addlinespace[1.5pt]
\taskmetric{\textbf{Bldg.\ change}}{F1 $\uparrow$} & LEVIR~\cite{levircd}, WHU-CD~\cite{whubuildings}, SECOND~\cite{second} \\[-1pt]
\subline{DINOv3-L + ChangeMamba~\cite{changemamba}} & \subline{plus off-nadir S2Looking~\cite{s2looking} separately} \\
\addlinespace[1.5pt]
\taskmetric{\textbf{Bldg.\ inst.}}{AP~\cite{coco} $\uparrow$} & xBD~\cite{xbd}, WHU~\cite{whubuildings}, BRIGHT~\cite{bright} \\[-1pt]
\subline{DINOv3-L + Mask R-CNN~\cite{maskrcnn}} & \subline{urban \& disaster; $\le$\,1\,500 det./image} \\
\bottomrule
\end{tabular*}
\end{table}

\paragraph{Training data and regimes.}
\label{sec:training}
\Cref{tab:setup} summarizes the four dense-prediction tasks with their models, real benchmarks, and metrics.
The main synthetic training pools contain up to 30\,000 \tld and 26\,985 \tlm samples; \tlmix draws equally from the two generators, with up to 53\,970 samples in total (\cref{tab:s-pools}).
\reg{SYN} trains exclusively on synthetic image--label pairs, while \reg{SYN+UD} additionally exploits unlabeled target imagery through iterative pseudo-label distillation~\cite{noisystudent}.
For scarce-label learning, \reg{REAL@p} uses $p$\,\% of a real training split, and \reg{REAL@p+SYN} supplements the same annotations through joint training or synthetic pretraining followed by real-data fine-tuning.
For building change detection, training retains only synthetic pairs containing labeled change (26\,154 for \tld, 8\,234 for \tlm, and 16\,468 for \tlmix); building instance segmentation uses 13\,401 tiles from each generator.

\paragraph{Evaluation protocol.}
Within each comparison, downstream architectures, training schedules, and checkpoint readout protocols are fixed across training sources.
The main transfer results are evaluated on real test (TEST) sets.
Unless otherwise specified, synthetic training uses label-preserving photometric and sensor randomization~\cite{tobin2017domain}, for our sources and for the procedural and generative baselines of \cref{tab:main} alike.
For land cover and height, SynRS3D is trained on a draw of its tiles matched in size to \tlm's pool of 26\,985 (\tld: 30\,000; \tlmix: 53\,970), at the same number of updates (\cref{tab:s-pools}).
Full protocols are in \cref{sec:s-protocols}.

%% file: sections/5_transfer.tex
\section{Results}
\label{sec:supervision}

\subsection{Generation efficiency and transfer}
\label{sec:transfer}
\label{sec:complexity}
\label{sec:cost}

Throughput is measured for all three generators on the same hardware, as rendered images per GPU-hour (one GB200 GPU with its 32 CPU threads, of which \tlm uses only the CPU threads; \cref{sec:s-cost}).
On this basis, \tlm achieves \textbf{38.7{\boldmath$\times$}} the generation throughput of \tld, while \tlm and \tld run approximately \textbf{382{\boldmath$\times$}} and \textbf{9.7{\boldmath$\times$}} faster than SynRS3D, respectively (\cref{fig:teaser}).

Despite this large gap in generation cost, the asset-heavy \tld does not consistently outperform the lightweight \tlm on real imagery (\cref{tab:main}).
Under synthetic-only training, \tlm leads in land-cover mapping (58.10 vs.\ 54.08~mIoU), whereas \tld performs better in building change detection and instance segmentation, and both perform similarly in height estimation.
\tld's handcrafted 3D building models may aid building-centric tasks, suggesting that the value of geometric and visual detail depends on the downstream task rather than on generation complexity alone.

The two generators are complementary: their equal mixture, \tlmix, improves land-cover mapping, height estimation, and building change detection over either generator alone while matching \tld in instance segmentation.
Compared with SynRS3D, \tlmix improves land-cover mIoU by \textbf{9.15 points}, reduces height MAE by \textbf{1.252~m}, and boosts building-change F1 by \textbf{10.99 points}, while reaching \textbf{37.11 mask AP} on building instances (which SynRS3D does not support).
In change detection, where SyntheWorld is the stronger prior procedural baseline, \tlmix still leads by 7.24~F1.
Adapting to unlabeled target imagery via \uda further improves the first three tasks while maintaining instance segmentation accuracy.
Beyond these benchmarks, the same models approach an xBD-supervised model in recognizing destroyed buildings without any damage labels and achieve higher building object F1 than global building products~\cite{sirko2021continental,msfootprints,overture} on xBD and BRIGHT (\cref{sec:s-utility}).

\subsection{Learning with scarce real labels}
\label{sec:scarce}
\label{sec:fewlabels}

Supplementing limited real data with \tlmix yields positive test-set mean gains across all nine label sources at budgets up to 5\,\% (\cref{fig:scarce}), and remains positive for eight and seven sources at 25\,\% and 100\,\% labels, respectively.
At higher label budgets, gains increasingly stem from better cross-dataset generalization beyond the labeled source domain, whereas under extreme label scarcity, synthetic-only training (\reg{SYN}) alone substantially outperforms real-only training.

\subsection{Comparison with conditional generators}
\label{sec:conditional}

\begin{figure*}[t]
\centering
\includegraphics[width=\textwidth]{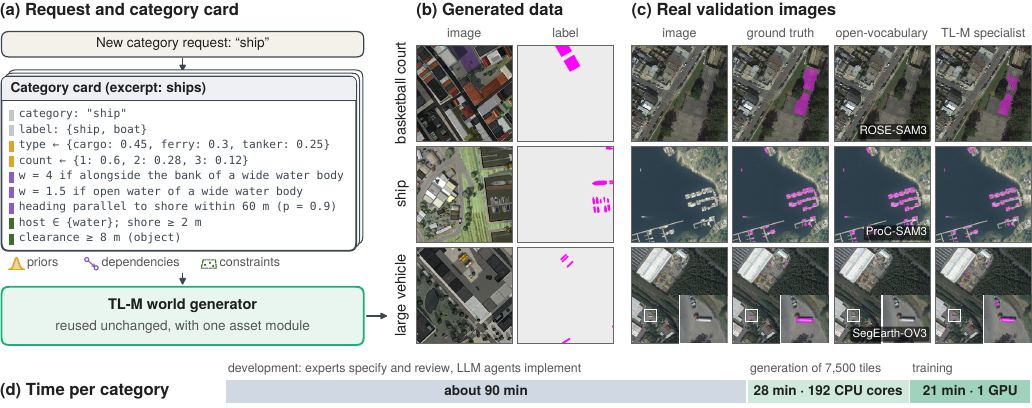}
\caption{\textbf{On-demand category extension.}
(a) A requested category is written as a card of priors, dependencies, and constraints that the unchanged \tlm generator executes.
(b) Generated images and labels and (c) real validation images with the best open-vocabulary method and the \tlm specialist, for the categories with the largest, median, and smallest difference between the two; the inset magnifies the boxed region.
(d) Time needed for one category.}
\label{fig:ondemand}
\end{figure*}

Unlike Seg2Sat, EarthSynth, VectorSynth-COSA, and Changen2---which train on real RGB imagery paired with annotations or geospatial vectors---\tl generates both imagery and supervision procedurally without any real image--annotation pairs (\cref{tab:main}).
Even without real training data, \tlmix substantially outperforms Seg2Sat, EarthSynth, and VectorSynth-COSA in land-cover mapping (58.79 vs.\ 15.24--27.01~mIoU).
Moreover, while Changen2-OEM is trained directly on OEM, annotation-free \tlmix surpasses it on the unseen LoveDA benchmark (60.11 vs.\ 56.51~mIoU).
Changen2 S1 and S9 reach higher building-change F1 (82.06 and 81.71 vs.\ 75.07 for \tlmix and 79.44 with \uda), but do so by conditioning synthesis on real images and building or land-cover annotations.

\begin{figure}[t]
\centering
\includegraphics[width=\columnwidth]{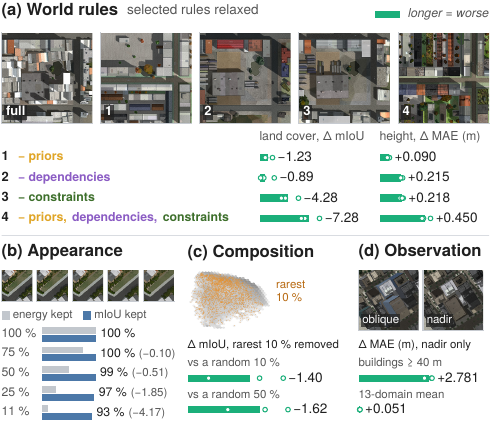}
\caption{\textbf{Dissecting world rules and scene properties.}
(a) Selected priors, dependencies, and constraints of \tlm relaxed alone or together (\emph{full}: the intact generator).
(b) Appearance: texture energy and land-cover mIoU kept under texture fading.
(c) Composition: uncommon scene mixtures versus scene count.
(d) Observation: oblique training views.
Hollow markers denote seeds.}
\label{fig:properties}
\label{fig:core}
\end{figure}

\begin{table*}[t]
\caption{\textbf{On-demand category extension on iSAID (validation split, foreground IoU, \%).} Best OV: the best of SegEarth-OV3 (S), ProC-SAM3 (P), and ROSE-SAM3 (R) for each category. $\Delta$: specialist minus best OV. Real: the same network trained on all real training labels. ST: storage tank, TC: tennis court, PL: plane, SH: ship, SP: swimming pool, BD: baseball diamond, BC: basketball court, RA: roundabout, SBF: soccer ball field, GTF: ground track field, SV: small vehicle, LV: large vehicle, BR: bridge, HA: harbor, HC: helicopter.}
\label{tab:ondemand}
\centering
\footnotesize
\input{tables/ondemand_wide}
\end{table*}

%% file: tables/ondemand_wide.tex
\setlength{\tabcolsep}{2.9pt}
\begin{tabular}{@{}lrrrrrrrrrrrrrrr@{\hspace{8pt}}r@{}}
\toprule
 & ST & TC & PL & SH & SP & BD & BC & RA & SBF & GTF & SV & LV & BR & HA & HC & Mean \\
\midrule
Best OV & 62.3\textsuperscript{P} & 44.6\textsuperscript{R} & 72.8\textsuperscript{R} & 49.9\textsuperscript{P} & \textbf{27.2}\textsuperscript{S} & 6.7\textsuperscript{P} & 8.0\textsuperscript{R} & 19.8\textsuperscript{R} & 29.6\textsuperscript{R} & 27.1\textsuperscript{P} & \textbf{47.0}\textsuperscript{P} & \textbf{59.5}\textsuperscript{S} & 17.8\textsuperscript{S} & 18.8\textsuperscript{R} & \textbf{24.2}\textsuperscript{P} & 34.3 \\
\rowcolor{mainhl}\tlm specialist & \textbf{69.5} & \textbf{81.1} & \textbf{82.6} & \textbf{60.2} & \textbf{27.2} & \textbf{41.8} & \textbf{47.3} & \textbf{30.5} & \textbf{67.5} & \textbf{43.6} & 35.0 & 43.5 & \textbf{22.1} & \textbf{40.7} & 17.2 & \textbf{47.3} \\
$\Delta$ & \textcolor{gaingreen}{+7.1} & \textcolor{gaingreen}{+36.6} & \textcolor{gaingreen}{+9.8} & \textcolor{gaingreen}{+10.3} & 0.0 & \textcolor{gaingreen}{+35.1} & \textcolor{gaingreen}{+39.4} & \textcolor{gaingreen}{+10.8} & \textcolor{gaingreen}{+38.0} & \textcolor{gaingreen}{+16.5} & $-$12.0 & $-$15.9 & \textcolor{gaingreen}{+4.3} & \textcolor{gaingreen}{+21.9} & $-$7.0 & \textcolor{gaingreen}{+13.0} \\
\rowcolor{mainhl}\tlm unified & 69.8 & 72.7 & 81.1 & 54.0 & 26.0 & 27.2 & 52.4 & 32.0 & 66.7 & 30.1 & 32.5 & 39.4 & 22.2 & 37.3 & 23.4 & 44.5 \\
\midrule
Real & 67.7 & 89.5 & 83.6 & 77.9 & 31.8 & 63.2 & 57.5 & 42.0 & 77.6 & 64.9 & 54.1 & 69.4 & 48.3 & 63.4 & 48.2 & 62.6 \\
\bottomrule
\end{tabular}

%% file: sections/7_properties.tex
\section{Extending and understanding synthetic worlds}
\label{sec:properties}
\label{sec:platform}

\subsection{On-demand category extension}
\label{sec:ondemand}

Because \tlm decouples world rules from heavy 3D assets, it can answer new recognition requests by synthesizing task supervision on demand (\cref{fig:ondemand}).
Given a request specifying the target categories, their scope, and basic imaging properties---here, color and panchromatic imagery at about 0.1--1~m GSD---\tlm executes a lightweight \emph{category card} for each class, and a specialist is trained on the generated tiles without any real labels.

\paragraph{Category cards.}
Each card is a machine-readable specification encoding a category's priors, dependencies, and constraints (\cref{sec:method})---covering its shapes, dimensions, materials, host scene contexts, spatial relations to nearby objects, placement constraints, frequency, and labeling rules---paired with a lightweight parametric asset module (\cref{fig:ondemand}a).
Two remote sensing experts defined the required knowledge schema and granularity; LLM agents then researched each requested category and implemented its card and asset module, which the experts reviewed together with rendered samples.
As with the core generators, no evaluation dataset is accessed during card development, and the underlying \tlm world generator is reused unchanged.

\paragraph{Setup.}
For each of the 15 object categories in iSAID~\cite{isaid}, \tlm renders 7\,500 tiles, on which a binary segmenter is trained for 8k updates with the land-cover model and optimizer settings of \cref{sec:protocol} (\cref{tab:setup}, \cref{tab:recipes}; task-specific sampling and augmentation in \cref{sec:s-ondemand}); we also train a single 16-class unified model on the union of all tiles to handle multi-category requests with one network.
We compare against three recent SAM\,3-based training-free open-vocabulary methods---SegEarth-OV3~\cite{segearthov3}, ProC-SAM3~\cite{procsam3}, and ROSE-SAM3~\cite{rosesam3}---using their released prompts and settings, benchmarking each category against the strongest of the three.
We report foreground IoU on the full iSAID validation split (\cref{sec:s-ondemand}).

\paragraph{Results.}
Trained without a single real label, \tlm specialists achieve a mean IoU of 47.3 (versus 62.6 when trained on all real iSAID labels), surpassing the per-category best open-vocabulary baseline (34.3) by 13.0 points and winning on 11 of the 15 categories (\cref{tab:ondemand}).
Gains are largest on categories governed by strong geometric regularity and spatial context---reaching $+$39.4 IoU on basketball courts---whereas open-vocabulary foundation models remain ahead on compact mobile targets (small and large vehicles, down to $-$15.9, and helicopters).
A single unified network reaches 44.5 mean IoU, within 2.9 points of the per-class specialists and $+$10.1 points above the per-category best open-vocabulary baseline.
Crucially, adding a new category takes just over two hours end-to-end (\cref{fig:ondemand}d): roughly 90~min for expert-guided card and module development, 28~min to render 7\,500 tiles on up to 192 CPU cores, and 21~min to train a specialist on a single GPU of the reference GB200 node.
The workflow is not tied to these categories or to semantic segmentation: cards for other categories follow the same schema, and the generated tiles also carry instance identifiers for instance-level tasks.

\subsection{Dissecting world rules and scene properties}
\label{sec:design}

We use \tlm to examine how its world rules and scene properties contribute to real-world transfer (\cref{fig:properties}, \cref{sec:s-atlas}).

\paragraph{World rules.}
Each ablation relaxes selected rules of one kind while keeping the rest of the generator intact: priors become uniform over their valid ranges, conditional dependencies become independent draws with unchanged marginals, and constraints are relaxed (8k updates, no photometric randomization; \cref{sec:s-mabl}).
Most of the library, including the rules that construct buildings, vegetation, and terrain, stays in force in every world.
All three mechanisms contribute to transfer, with task-dependent sensitivity (\cref{fig:properties}a): relaxing priors or dependencies lowers land-cover mIoU by about one point each ($-$1.23 and $-$0.89 points), whereas relaxing dependencies hurts height MAE more than twice as much as priors ($+$0.215 vs.\ $+$0.090~m), and relaxing constraints substantially degrades both ($-$4.28 mIoU, $+$0.218~m MAE).
Relaxing all three together compounds the degradation to $-$7.28 mIoU and $+$0.450~m MAE, the largest losses on both tasks.

\paragraph{Appearance, composition, and observation.}
Fading within-region texture to 50\,\% of its energy lowers land-cover mIoU by only 0.51 points, and fading to 11\,\% still retains 93\,\% of the full-texture mIoU ($-$4.17 points; \cref{fig:properties}b): texture matters, but its strength can be reduced far below the generator's default at moderate cost.
Removing the 10\,\% of scenes with the rarest class-area compositions causes 1.62~points more mIoU loss than randomly removing half of all scenes (\cref{fig:properties}c): diverse scene compositions can matter more than raw scene count.
Finally, removing oblique training views increases height MAE by 2.781~m for buildings at least 40~m tall while raising the overall 13-domain MAE by only 0.051~m (\cref{fig:properties}d), a benefit for tall structures that aggregate metrics largely hide.

%% file: sections/7_conclusion.tex
\section{Conclusion}
\label{sec:conclusion}

\tl translates domain knowledge into executable priors, dependencies, and constraints for procedural remote-sensing data generation.
\tld realizes this design with rich 3D assets, while \tlm achieves strong real-world transfer at high throughput with lightweight parametric representations and without pre-made 3D models.
Their combination (\tlmix) supports four dense-prediction tasks, substantially advances synthetic-to-real transfer over prior procedural datasets, boosts learning with scarce real labels, and remains competitive with diffusion-based generators that rely on real annotated imagery.
Beyond a fixed dataset, \tlm serves as an extensible and controllable platform: category cards extend it on demand to newly requested object classes without real labels---exceeding the best training-free open-vocabulary method on average---while rule ablations and targeted interventions reveal how world rules, texture strength, scene composition, and observation geometry shape transfer.
Together, these results favor preserving task-relevant world structure over indiscriminately scaling asset and rendering complexity.
Code and datasets will be released.

%% file: sections/8_llm_statement.tex
\section*{LLM usage statement}
We used LLM-based assistants (Claude Opus~5.5 and Codex with GPT-6~Astra) to survey the literature and public statistics behind the world rules, to propose candidate rules and parameters, to implement generator code, and to edit the manuscript.
Claude Opus~5.5 mainly carried out the implementation, and GPT-6~Astra cross-checked the proposed rules, parameters, and code.
The authors reviewed all LLM outputs, refined the rules, and take full responsibility for the manuscript.
\par

%% file: supp/supp.tex
\clearpage
\renewcommand{\thepage}{S\arabic{page}}
\renewcommand{\theHsection}{supp.\arabic{section}}
\renewcommand{\theHsubsection}{\theHsection.\arabic{subsection}}
\renewcommand{\theHtable}{supp.\arabic{table}}
\renewcommand{\theHfigure}{supp.\arabic{figure}}
\setcounter{page}{1}
\input{supp_contents}
\SuppTitleWithContents
\renewcommand{\thesection}{S\arabic{section}}
\renewcommand{\thetable}{S\arabic{table}}
\renewcommand{\thefigure}{S\arabic{figure}}
\setcounter{section}{0}
\setcounter{table}{0}
\setcounter{figure}{0}
\makeatletter\setlength{\@fptop}{0pt}\setlength{\@fpsep}{12pt plus 1fil}\setlength{\@fpbot}{0pt plus 1fil}
\setlength{\@dblfptop}{0pt plus 1fil}\setlength{\@dblfpsep}{12pt plus 2pt}\setlength{\@dblfpbot}{0pt plus 1fil}\makeatother

\SuppContentsSetup
\enlargethispage{-100pt}

This supplementary material provides additional examples of synthetic worlds and aligned labels, implementation details, experimental protocols, and extended results supporting the main paper.

\input{supp/s1_world}
\input{supp/s2_protocols}
\input{supp/s3_transfer}
\input{supp/s6_cost}
\FloatBarrier
\input{supp/s_generators}
\FloatBarrier
\input{supp/s5_real_labels}
\FloatBarrier
\clearpage
\clearpage
\input{supp/s4_interventions}
\FloatBarrier
\input{supp/s7_damage}
\input{supp/s8_products}
\FloatBarrier
\clearpage
\clearpage
\input{supp/s_ondemand}
\input{supp/s9_limitations}

%% file: supp/s1_world.tex
\section{Synthetic worlds, rules, and supervision}
\label{sec:s-world}
\label{sec:s-rules}

\tl comprises two independently developed generators, TellusLoom-Detailed (\tld) and TellusLoom-Mini (\tlm), which implement knowledge-driven world design using different scene representations and rendering pipelines.
This section presents their generated scenes and aligned supervision, summarizes the scale and composition of their executable rule libraries, and describes representative rules, label generation, and implementation differences.

\subsection{Scene gallery with aligned labels}
\label{sec:s-gallery}

\begin{figure*}[p]
\centering
\includegraphics[width=\textwidth]{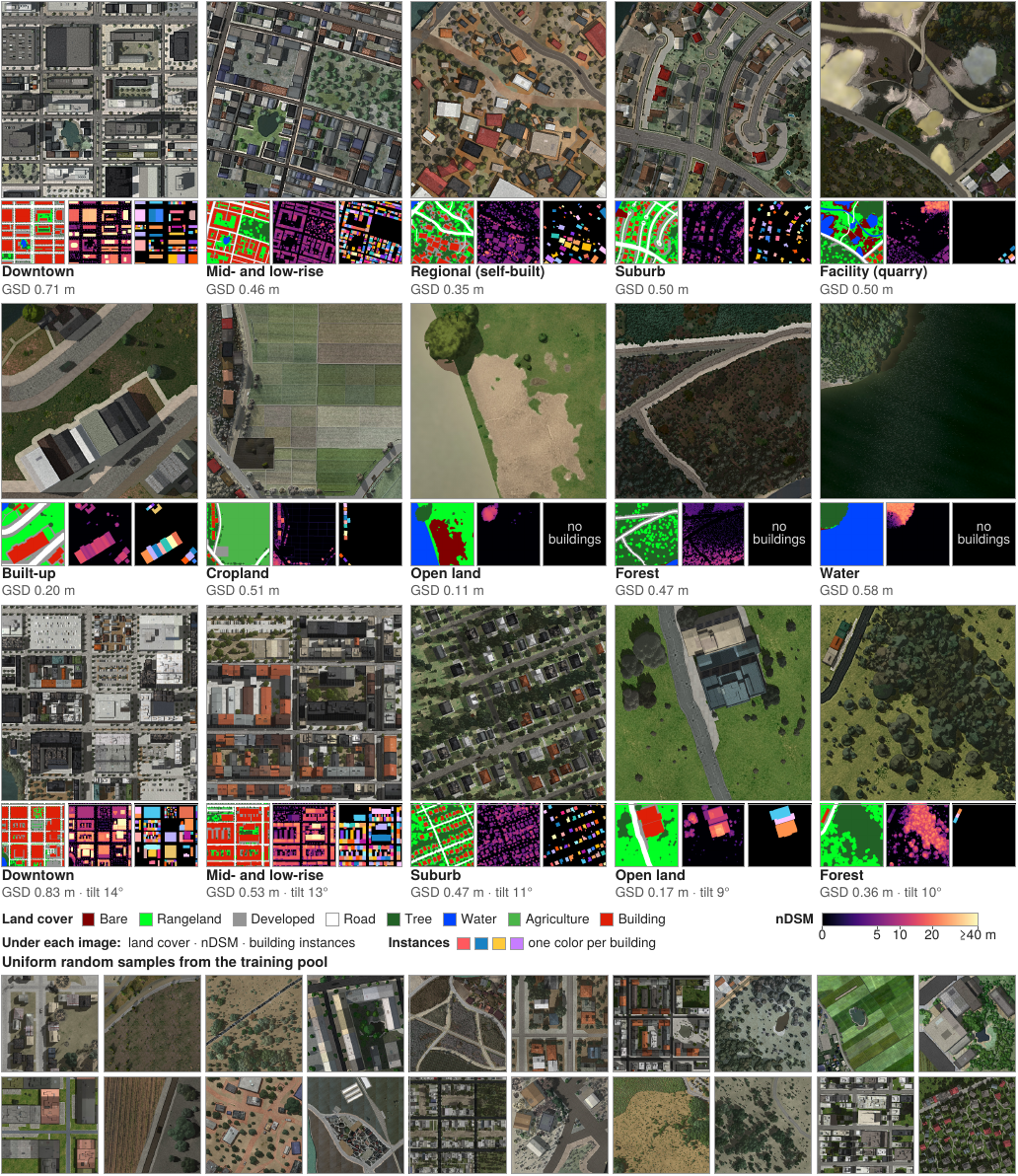}
\caption{\textbf{Synthetic scenes and aligned labels from \tlm.} One training tile per scene kind of the nadir scene program (first two rows) and of the tilting scene program (third row, tilted views), each with its aligned land-cover, height (nDSM), and building-instance labels and its ground sampling distance (GSD). The kinds of the first row and of the first three tiles of the third row are scene archetypes of the generator; the other tiles have no archetype and are named after the land-cover group that covers at least half of the tile (built-up: building, developed, and road; open land: rangeland and bare land). Bottom two rows: 20 tiles drawn uniformly at random from the training pool.}
\label{fig:s-gallery-tlm}
\end{figure*}

\begin{figure*}[p]
\centering
\includegraphics[width=\textwidth]{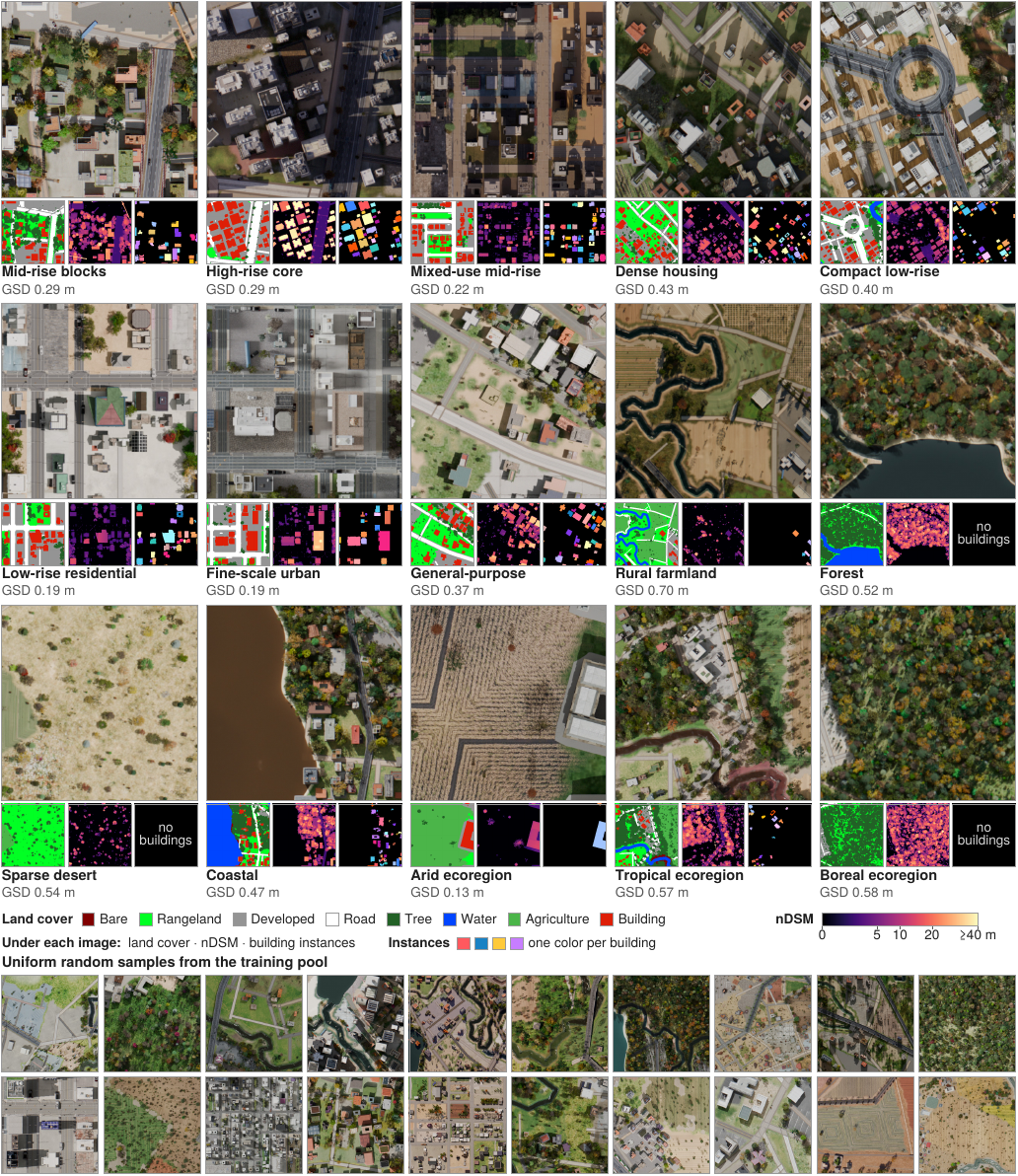}
\caption{\textbf{Synthetic scenes and aligned labels from \tld.} One training tile for each of the 12 scene families of the generator and, as the last three tiles, one each from its arid, tropical, and boreal ecoregions, with aligned land-cover, height (nDSM), and building-instance labels and the GSD. A title names the scene family, a set of generation settings, or the ecoregion from which the tile is drawn, not a class of the single tile. Bottom two rows: 20 tiles drawn uniformly at random from the training pool. All training scenes use near-nadir views.}
\label{fig:s-gallery-tld}
\end{figure*}

\Cref{fig:s-gallery-tlm,fig:s-gallery-tld} showcase the diversity of worlds generated by \tlm and \tld, respectively.
The examples cover urban, rural, agricultural, forested, and other landscapes at different ground sampling distances, together with their aligned land-cover maps, height maps (nDSM), and building-instance masks.

\begin{figure*}[p]
\centering
\includegraphics[width=\textwidth]{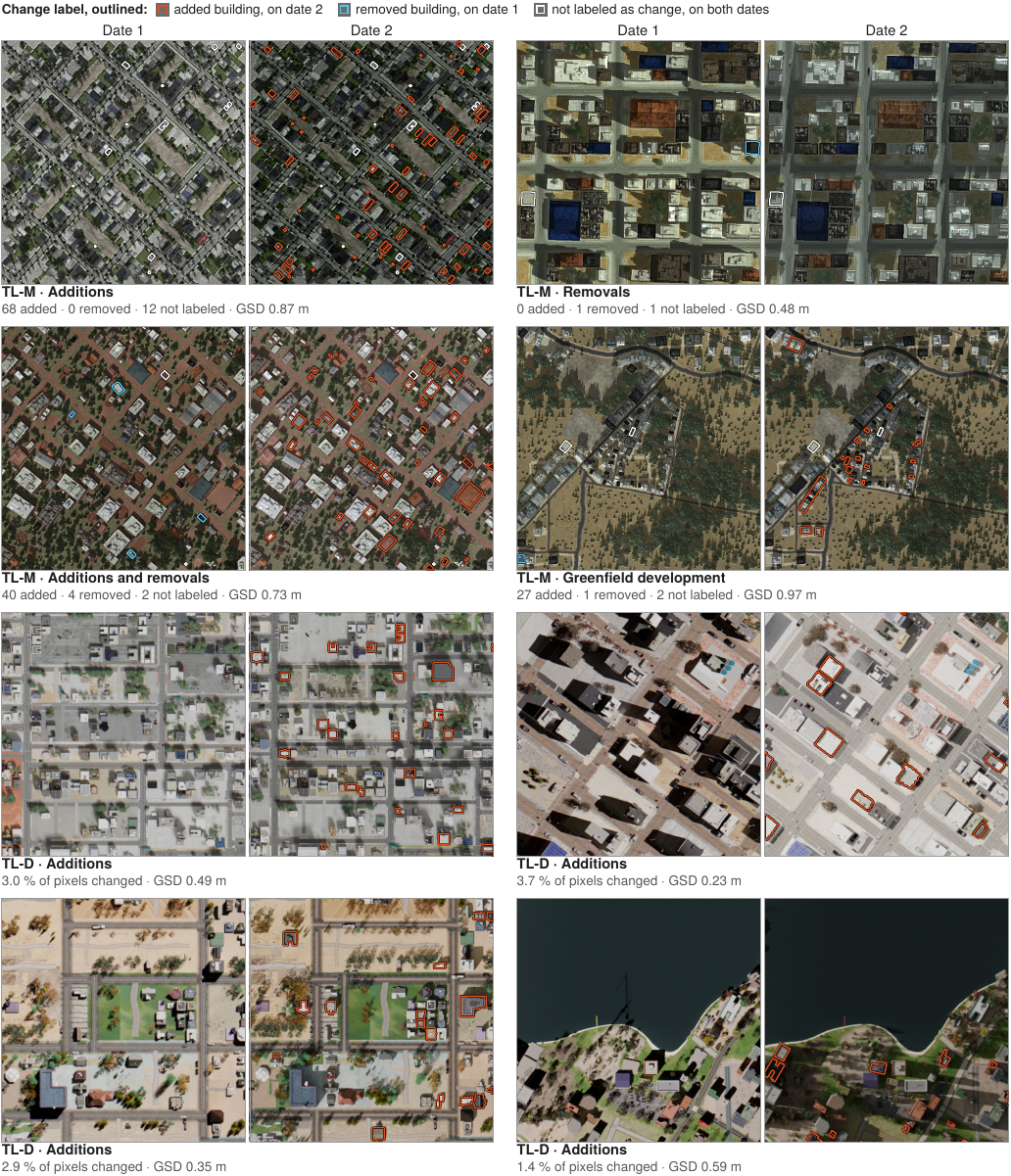}
\caption{\textbf{Two-date building-change pairs.} Pairs from \tlm (first two rows) and \tld (last two rows) illustrate building additions, removals, and greenfield development. Outlines distinguish labeled changes from events excluded by the labeling criteria (\cref{sec:s-observations}).}
\label{fig:s-gallery-events}
\end{figure*}

\Cref{fig:s-gallery-events} shows two-date building-change pairs of both generators (\cref{sec:s-observations}).

\FloatBarrier
\clearpage
\clearpage

\subsection{Scale and coverage of world rules}
\label{sec:s-rulescale}

The frozen \tlm and \tld generators contain 1\,266 and 797 executable world rules, respectively (\cref{tab:s-rules}).
A rule is counted as a distinct semantic decision governing scene construction or observation, rather than an individual parameter, code branch, or helper function (\cref{fig:s-rule-listing} writes out examples in full).
\tlm builds all of its geometry procedurally, including the buildings and trees for which \tld uses prepared 3D assets (\cref{tab:s-implementation}), and therefore holds more rules.
\tlm's separate two-date driver holds another 85 rules, which \cref{tab:s-sources} counts with \tlm; \tld's temporal rules are included in its count.

\begin{table}[t]
\caption{\textbf{Scale and coverage of executable world rules.} (a)~Rules by mechanism and world-design dimension. A rule may involve several mechanisms and dimensions, so cells overlap; the column \emph{Rules} and the rows \emph{All rules} count each rule once; \tlm's two-date driver is not included. (b)~Percentages of rules that span multiple dimensions, counting the dimensions a rule reads from and those it decides, or that depend on variables from other dimensions. CC: Composition \& Context; SO: Spatial Organization; GM: Geometry \& Morphology; SA: Surface Appearance; OF: Observation Formation.}
\label{tab:s-rules}
\centering
\footnotesize
\setlength{\tabcolsep}{3pt}
\input{tables/supp_rules}
\end{table}

\begin{figure*}[!t]
\footnotesize
{\centering\rulepill{rulep}{priors}\quad\rulepill{ruled}{dependencies}\quad\rulepill{rulec}{constraints}\par}
\vspace{4pt}
\begin{minipage}[t]{\columnwidth}\vspace{0pt}\centering
\input{tables/supp_listing_chain}\\[2pt]
(a)
\end{minipage}\hfill
\begin{minipage}[t]{\columnwidth}\vspace{0pt}\centering
\input{tables/supp_listing_rules}\\[2pt]
(b)
\end{minipage}
\caption{\textbf{Four rows of \cref{tab:rulecards} written out as the generators execute them}, with the values of the frozen generators' configurations. (a)~Its first three rows, the chain of \cref{sec:structured-knowledge} in \tld: road classes limit grades, roads are graded into the terrain, fields take the direction of the nearest road, and each pair of neighboring parcels draws its boundary. (b)~Its row \emph{Heights in context}, a guard of \tlm: a prior draws the heights, conditional dependencies describe each building's context, and constraints bound its height; where a limit is exceeded, the generator enforces it by swapping heights with a nearby building or by clipping.}
\label{fig:s-rule-listing}
\end{figure*}

Both generators implement priors, conditional dependencies, and constraints across all five world-design dimensions.
Most rules span multiple dimensions (71\,\% in \tlm and 59\,\% in \tld), and approximately half depend on variables outside the dimension they govern (51\,\% and 48\,\%, respectively).
For example, terrain influences road routing, road geometry determines surface markings, and building density affects vegetation placement.
These statistics characterize the scale and interconnected structure of the rule libraries.

\subsection{Rule examples, knowledge sources, and authoring}
\label{sec:s-provenance}
\label{sec:s-parameters}

\Cref{tab:rulecards} lists fourteen example rules in plain language, each with its generator, its mechanisms, and the dimensions it involves.
The first three form the chain of \cref{sec:structured-knowledge}, from road classes and grades through cut-and-fill and field directions to the boundaries between parcels; the table also holds the view law of \tlm's tilting scene program.
\Cref{fig:s-rule-listing} lists four of these rows step by step, as two listings in the order in which the generators execute them and with their values.

\Cref{tab:s-sources} counts the rules of both generators by knowledge source; a rule may draw on several sources, so the shares sum to more than one hundred percent.
Expert design and selection contributes to 57\,\% of the rules and LLM proposals to 37\,\%.
Analytical derivations are most frequent in observation formation (42\,\% of its rules), where viewing and illumination geometry determine the image, while literature and statistics anchor a smaller set of quantitative rules.
A source's share of the rule count does not measure its scope: the rules grounded in literature and statistics fix scene-wide quantities such as building-height distributions, tree height-to-crown allometry, and road-network statistics (\cref{tab:s-provenance}, which lists these quantities with their sources), while expert design covers the many contextual decisions where empirical measurements are unavailable.

In writing a scene type, object class, or rule family, a researcher first set out what was known, what was uncertain, and which knowledge was missing.
The researchers then used Claude Opus~5.5 to survey literature and public statistics, to propose candidate rules and analytical relations, and to draft the code, and Codex with GPT-6~Astra to cross-check the proposed rules, parameters, and code.
They reviewed the rules and the cross-check comments together with rendered previews and aligned labels across ground sampling distances, and decided what was accepted, revised, or frozen; uncertain parameters were selected in side-by-side comparisons of rendered candidates that differ in the value under review.
Domain judgment thus guides how a rule is written and checked, and every accepted decision is an explicit, executable rule (as in \cref{fig:s-rule-listing}) that can be inspected, ablated, and changed in the code.

\begin{table}[t]
\caption{\textbf{Executable world rules by knowledge source and world-design dimension.} Rules of \tlm (with its two-date driver) and \tld together. A rule may draw on several sources and decide several dimensions, so cells overlap; the column \emph{Rules} and the row \emph{All rules} count each rule once. Abbreviations as in \cref{tab:s-rules}.}
\label{tab:s-sources}
\centering
\footnotesize
\setlength{\tabcolsep}{3pt}
\input{tables/supp_rule_sources}
\end{table}

\begin{table*}[!tp]
\caption{\textbf{Example executable world rules.} Each row states in plain language what a rule or rule family of the frozen generators does. Mechanisms: \mechP~prior, \mechD~conditional dependency, \mechC~constraint; dimensions as in \cref{tab:s-rules}. The first three rows are the chain of \cref{sec:structured-knowledge}; \emph{View law} is the rule that draws the view of a tile.}
\label{tab:rulecards}
\centering
\footnotesize
\renewcommand{\arraystretch}{1.1}
\input{tables/supp_rulecards}
\end{table*}

\begin{table*}[!tp]
\caption{\textbf{Quantities and methods that rest on literature and public data.} For each, the generator, the sources, and how it enters the generator. The other rules come from analytical derivation, LLM proposals, and expert design (\cref{tab:s-sources}).}
\label{tab:s-provenance}
\label{tab:s-paramgroups}
\centering
\footnotesize
\renewcommand{\arraystretch}{1.1}
\input{tables/supp_provenance}
\end{table*}

\subsection{Visible-surface and two-date change labels}
\label{sec:s-observations}

\paragraph{Visible-surface supervision.}
Both generators derive imagery and task labels from the same scene state and observation geometry.
Land-cover maps follow the OpenEarthMap (OEM) taxonomy~\cite{openearthmap}, height labels represent above-ground height (nDSM), and building-instance masks preserve individual identities.
Under \tlm's oblique views (third row of \cref{fig:s-gallery-tlm}), labels follow visible surfaces, including building facades and occlusions.
Changing the camera therefore produces a new image and its corresponding labels, rather than transforming labels from a nadir view.

\paragraph{Two-date building changes.}
\label{sec:s-events}
Building-change pairs are generated from two scene states (\cref{fig:s-gallery-events}): \tld obtains the earlier state by leaving out a share of the constructed scene's buildings and trees, so its building changes are additions, whereas \tlm's two-date driver also removes buildings.
The dates may also differ in camera position, illumination, and surface appearance, and in \tlm also in season.
Change masks are derived from the visible labels of the two scene states, so changes hidden under tree crowns or taller roofs are not labeled; \tlm additionally labels a building event only when its ground area is at least 10~m$^2$, its footprint in the image spans at least 4~pixels, and at least a share of 0.5 of it is visible.
These distinctions allow the training pairs to contain realistic observation differences without treating every appearance change as a building change.

\subsection{Training pools and generator differences}
\label{sec:s-pools}
\label{sec:s-implementation}

\paragraph{Training pools.}
The main synthetic pools contain 30\,000 \tld and 26\,985 \tlm samples (\cref{tab:s-pools}).
\tlmix combines equal numbers from both sources, yielding 53\,970 samples.
Neither pool is curated: \tlm's pool consists of all tiles that the generator produced, and \tld's is a uniform random draw, with a fixed seed, from the training tiles that the generator produced; no sample is selected by its content.
Every sample is determined by the frozen generator configuration and its seed, so both pools can be regenerated.
The two pools differ in scene composition, ground sampling distance, and observation geometry (\cref{tab:s-pools}): \tld uses near-nadir views, while \tlm additionally includes tilted observations, and \tlmix combines both.
These pools train land cover and height; building change detection uses each generator's two-date pairs that contain labeled change (26\,154 for \tld and 8\,234 for \tlm), and instance segmentation uses 13\,401 building-instance tiles from each generator (\cref{sec:training}).
The change pairs are selected from all two-date samples of a generator and are not a subset of the pools above; \cref{tab:s-size} compares the sources at the same number of pairs.

\begin{table*}[!tp]
\caption{\textbf{Composition of the synthetic training pools of land cover and height.} Number of samples, GSD distributions, tilted-view proportions, and land-cover class frequencies for \tld, \tlm, and their equal mixture (\tlmix). Class frequencies are computed over labeled pixels.}
\label{tab:s-pools}
\centering
\footnotesize
\setlength{\tabcolsep}{3pt}
\input{tables/supp_pools}
\end{table*}

\paragraph{Generator differences.}
\tld constructs detailed 3D scenes in Blender~\cite{blender}, combining procedural terrain, infrastructure, vegetation, and prepared assets with GPU path tracing.
\tlm instead generates all of its geometry, including buildings and trees, procedurally in NumPy~\cite{numpy}, supplemented by lightweight CPU-based Blender rendering.
\Cref{tab:s-implementation} summarizes their main implementation differences.
In \tld, the fine-grained geometry and local appearance of objects such as individual buildings and trees are largely encapsulated in prepared 3D assets and asset-level Blender modules: how these objects are shaped is fixed by the assets and is not set by our rules.
\tld's executable rules therefore govern scene composition, layout, road and river networks, the placement of objects in their context, and observation formation, but not the internal construction of these objects.
\tlm, by contrast, constructs all geometry procedurally without pre-made 3D assets, so its rules govern both scene organization and object shapes.
Because the two generators differ in this way, the contribution of world rules is measured by ablations within \tlm (\cref{sec:s-mabl}).

\begin{table}[!htb]
\caption{\textbf{Implementation differences between \tld and \tlm.} Scene construction, mesh-asset usage, rendering backend, and generated labels. Mesh counts indicate the total asset library and the number of files opened per scene. Generation resources are compared in \cref{tab:cost}.}
\label{tab:s-implementation}
\centering
\footnotesize
\setlength{\tabcolsep}{3pt}
\begin{tabular}{@{}>{\raggedright\arraybackslash}p{2.65cm}>{\centering\arraybackslash}p{2.5cm}>{\centering\arraybackslash}p{2.5cm}@{}}
\toprule
 & \tld & \tlm \\
\midrule
Scene construction & Procedural geometry and prepared assets & Procedural geometry \\
Mesh files (library / opened per scene) & 957 / 217 & 0 / 0 \\
Renderer & Blender (Cycles), GPU path tracing & NumPy and Blender (Cycles), CPU \\
Labels & \multicolumn{2}{>{\centering\arraybackslash}p{5.2cm}}{Land cover, nDSM, building instances, two-date change} \\
\bottomrule
\end{tabular}
\end{table}

%% file: tables/supp_rules.tex
\begin{tabular}{@{}llrrrrrr@{}}
\toprule
\multicolumn{8}{@{}l}{\textit{(a) Rules by mechanism and decided dimension}} \\
 & & CC & SO & GM & SA & OF & Rules \\
\midrule
\tlm & \rulepill{rulep}{Prior} & 192 & 268 & 249 & 136 & 22 & 589 \\
 & \rulepill{ruled}{Dependency} & 191 & 286 & 260 & 225 & 62 & 766 \\
 & \rulepill{rulec}{Constraint} & 154 & 314 & 221 & 138 & 75 & 649 \\
 & All rules & 313 & 513 & 435 & 307 & 119 & 1\,266 \\
\addlinespace
\tld & \rulepill{rulep}{Prior} & 87 & 77 & 102 & 66 & 35 & 302 \\
 & \rulepill{ruled}{Dependency} & 129 & 173 & 135 & 84 & 59 & 473 \\
 & \rulepill{rulec}{Constraint} & 46 & 155 & 91 & 32 & 68 & 341 \\
 & All rules & 181 & 285 & 207 & 127 & 136 & 797 \\
\midrule
\multicolumn{2}{@{}l}{\textit{(b) Share of rules}} & \multicolumn{3}{r}{\tlm} & \multicolumn{3}{r@{}}{\tld} \\
\midrule
\multicolumn{2}{@{}l}{Span two or more dimensions} & \multicolumn{3}{r}{71\,\%} & \multicolumn{3}{r@{}}{59\,\%} \\
\multicolumn{2}{@{}l}{Depend on another dimension} & \multicolumn{3}{r}{51\,\%} & \multicolumn{3}{r@{}}{48\,\%} \\
\bottomrule
\end{tabular}

%% file: tables/supp_listing_chain.tex
\setlength{\aboverulesep}{0pt}\setlength{\belowrulesep}{0pt}\renewcommand{\arraystretch}{1.15}%
\begin{tabular}{@{}p{2pt}@{\hspace{4pt}}>{\raggedleft\arraybackslash}p{9pt}@{\hspace{5pt}}>{\raggedright\arraybackslash}p{\dimexpr\linewidth-20pt\relax}@{}}
\toprule
\multicolumn{3}{@{}>{\columncolor{ruletitle}[0pt][0pt]}p{\linewidth}@{}}{\rule{0pt}{9.5pt}\hspace{4pt}\textbf{Roads, terrain, fields, and field boundaries (\tld)}\rule[-3.5pt]{0pt}{0pt}} \\
\midrule
\addlinespace[3pt]
 & \ruleno{01} & \hangindent=0.9em\hangafter=1\relax \textbf{input:} the terrain and the road classes of a scene \\
\addlinespace[3pt]
\multicolumn{1}{@{}>{\columncolor{rulec}[0pt][0pt]}p{2pt}@{\hspace{4pt}}}{} & \ruleno{02} & \hangindent=0.9em\hangafter=1\relax \textbf{for each} road of class $c$, grown segment by segment:\ruletag{rulec}{constraints} \\
\multicolumn{1}{@{}>{\columncolor{rulec}[0pt][0pt]}p{2pt}@{\hspace{4pt}}}{} & \ruleno{03} & \leftskip=1.1em\hangindent=0.9em\hangafter=1\relax grade $\leq g(c)$: 5\,\% expressway, 7\,\% arterial, 10\,\% collector, 14\,\% local street, 16\,\% service road, 20\,\% track \\
\addlinespace[3pt]
\multicolumn{1}{@{}>{\columncolor{ruled}[0pt][0pt]}p{2pt}@{\hspace{4pt}}}{} & \ruleno{04} & \hangindent=0.9em\hangafter=1\relax \textbf{for each} road:\ruletag{ruled}{dependencies} \\
\multicolumn{1}{@{}>{\columncolor{ruled}[0pt][0pt]}p{2pt}@{\hspace{4pt}}}{} & \ruleno{05} & \leftskip=1.1em\hangindent=0.9em\hangafter=1\relax cut or fill the terrain toward the road profile within half the road width $+$ 6 m \\
\addlinespace[3pt]
\multicolumn{1}{@{}>{\columncolor{rulec}[0pt][0pt]}p{2pt}@{\hspace{4pt}}}{} & \ruleno{06} & \hangindent=0.9em\hangafter=1\relax \textbf{fields:}\ruletag{rulec}{constraints} \\
\multicolumn{1}{@{}>{\columncolor{rulec}[0pt][0pt]}p{2pt}@{\hspace{4pt}}}{} & \ruleno{07} & \leftskip=1.1em\hangindent=0.9em\hangafter=1\relax only on ground with a slope below 37$^\circ$ \\
\addlinespace[3pt]
\multicolumn{1}{@{}>{\columncolor{ruled}[0pt][0pt]}p{2pt}@{\hspace{4pt}}}{} & \ruleno{08} & \hangindent=0.9em\hangafter=1\relax \textbf{for each} field seed:\ruletag{ruled}{dependencies} \\
\multicolumn{1}{@{}>{\columncolor{ruled}[0pt][0pt]}p{2pt}@{\hspace{4pt}}}{} & \ruleno{09} & \leftskip=1.1em\hangindent=0.9em\hangafter=1\relax field $\leftarrow$ rounded rectangle of 200--500 m by 150--400 m whose axis is the direction of the nearest road \\
\addlinespace[3pt]
\multicolumn{1}{@{}>{\columncolor{rulep}[0pt][0pt]}p{2pt}@{\hspace{4pt}}}{} & \ruleno{10} & \hangindent=0.9em\hangafter=1\relax \textbf{for each} pair of neighboring parcels:\ruletag{rulep}{priors} \\
\multicolumn{1}{@{}>{\columncolor{rulep}[0pt][0pt]}p{2pt}@{\hspace{4pt}}}{} & \ruleno{11} & \leftskip=1.1em\hangindent=0.9em\hangafter=1\relax boundary $\leftarrow$ hedgerow, ditch, dirt track, or none (two crop fields: 0.15, 0.12, 0.22, 0.51) \\
\addlinespace[3pt]
\multicolumn{1}{@{}>{\columncolor{ruled}[0pt][0pt]}p{2pt}@{\hspace{4pt}}}{} & \ruleno{12} & \hangindent=0.9em\hangafter=1\relax \textbf{labels} of a boundary:\ruletag{ruled}{dependencies} \\
\multicolumn{1}{@{}>{\columncolor{ruled}[0pt][0pt]}p{2pt}@{\hspace{4pt}}}{} & \ruleno{13} & \leftskip=1.1em\hangindent=0.9em\hangafter=1\relax hedgerow $\Rightarrow$ a dense row of trees, labeled tree; dirt track $\Rightarrow$ labeled road \\
\addlinespace[3pt]
\multicolumn{1}{@{}>{\columncolor{rulec}[0pt][0pt]}p{2pt}@{\hspace{4pt}}}{} & \ruleno{14} & \hangindent=0.9em\hangafter=1\relax \textbf{per scene:}\ruletag{rulec}{constraints} \\
\multicolumn{1}{@{}>{\columncolor{rulec}[0pt][0pt]}p{2pt}@{\hspace{4pt}}}{} & \ruleno{15} & \leftskip=1.1em\hangindent=0.9em\hangafter=1\relax at most 40 hedgerows, 8 ditches, and 32 dirt tracks \\
\addlinespace[3pt]
\bottomrule
\end{tabular}

%% file: tables/supp_listing_rules.tex
\setlength{\aboverulesep}{0pt}\setlength{\belowrulesep}{0pt}\renewcommand{\arraystretch}{1.15}%
\begin{tabular}{@{}p{2pt}@{\hspace{4pt}}>{\raggedleft\arraybackslash}p{9pt}@{\hspace{5pt}}>{\raggedright\arraybackslash}p{\dimexpr\linewidth-20pt\relax}@{}}
\toprule
\multicolumn{3}{@{}>{\columncolor{ruletitle}[0pt][0pt]}p{\linewidth}@{}}{\rule{0pt}{9.5pt}\hspace{4pt}\textbf{Context-coherent building heights (\tlm)}\rule[-3.5pt]{0pt}{0pt}} \\
\midrule
\addlinespace[3pt]
 & \ruleno{01} & \hangindent=0.9em\hangafter=1\relax \textbf{input:} the buildings of a tile, with heights drawn from its height prior \\
\addlinespace[3pt]
\multicolumn{1}{@{}>{\columncolor{ruled}[0pt][0pt]}p{2pt}@{\hspace{4pt}}}{} & \ruleno{02} & \hangindent=0.9em\hangafter=1\relax \textbf{for each} building $b$:\ruletag{ruled}{dependencies} \\
\multicolumn{1}{@{}>{\columncolor{ruled}[0pt][0pt]}p{2pt}@{\hspace{4pt}}}{} & \ruleno{03} & \leftskip=1.1em\hangindent=0.9em\hangafter=1\relax $\rho(b) \leftarrow$ local built coverage around $b$ \\
\multicolumn{1}{@{}>{\columncolor{ruled}[0pt][0pt]}p{2pt}@{\hspace{4pt}}}{} & \ruleno{04} & \leftskip=1.1em\hangindent=0.9em\hangafter=1\relax \textit{house}$(b) \leftarrow$ footprint $<$ 250 m$^2$ and $\rho(b) <$ 0.3 \\
\multicolumn{1}{@{}>{\columncolor{ruled}[0pt][0pt]}p{2pt}@{\hspace{4pt}}}{} & \ruleno{05} & \leftskip=1.1em\hangindent=0.9em\hangafter=1\relax $s(b) \leftarrow$ 3.5 for footprints up to 400 m$^2$, rising linearly to 5.5 at 1\,200 m$^2$ and above \\
\multicolumn{1}{@{}>{\columncolor{ruled}[0pt][0pt]}p{2pt}@{\hspace{4pt}}}{} & \ruleno{06} & \leftskip=1.1em\hangindent=0.9em\hangafter=1\relax $S(b) \leftarrow$ height of the second tallest other building within 150 m \\
\addlinespace[3pt]
\multicolumn{1}{@{}>{\columncolor{rulec}[0pt][0pt]}p{2pt}@{\hspace{4pt}}}{} & \ruleno{07} & \hangindent=0.9em\hangafter=1\relax \textbf{limits} of $b$:\ruletag{rulec}{constraints} \\
\multicolumn{1}{@{}>{\columncolor{rulec}[0pt][0pt]}p{2pt}@{\hspace{4pt}}}{} & \ruleno{08} & \leftskip=1.1em\hangindent=0.9em\hangafter=1\relax house cap: \textit{house}$(b) \Rightarrow h(b) \leq$ 15 m \\
\multicolumn{1}{@{}>{\columncolor{rulec}[0pt][0pt]}p{2pt}@{\hspace{4pt}}}{} & \ruleno{09} & \leftskip=1.1em\hangindent=0.9em\hangafter=1\relax slenderness: $h(b) \leq s(b)\,\times$ short side of the footprint \\
\multicolumn{1}{@{}>{\columncolor{rulec}[0pt][0pt]}p{2pt}@{\hspace{4pt}}}{} & \ruleno{10} & \leftskip=1.1em\hangindent=0.9em\hangafter=1\relax support: if $h(b) \geq$ 20 m and three or more other buildings lie within 150 m, $h(b) \leq 2.5 \times \max(S(b),$ 8 m$)$ \\
\multicolumn{1}{@{}>{\columncolor{rulec}[0pt][0pt]}p{2pt}@{\hspace{4pt}}}{} & \ruleno{11} & \hangindent=0.9em\hangafter=1\relax \textbf{for each} $b$ above a limit, tallest first: \\
\multicolumn{1}{@{}>{\columncolor{rulec}[0pt][0pt]}p{2pt}@{\hspace{4pt}}}{} & \ruleno{12} & \leftskip=1.1em\hangindent=0.9em\hangafter=1\relax swap heights with the nearest lower building that is not a house and whose limits admit $h(b)$; if there is none, clip $h(b)$ to its limit \\
\addlinespace[3pt]
\bottomrule
\end{tabular}

%% file: tables/supp_rule_sources.tex
\begin{tabular}{@{}lrrrrrrr@{}}
\toprule
Knowledge source & CC & SO & GM & SA & OF & Rules & \% \\
\midrule
Literature and statistics & 29 & 52 & 32 & 8 & 14 & 115 & 5 \\
Analytical derivation & 28 & 99 & 88 & 33 & 115 & 295 & 14 \\
LLM proposal & 194 & 307 & 297 & 171 & 72 & 800 & 37 \\
Expert design and selection & 325 & 481 & 321 & 274 & 135 & 1\,226 & 57 \\
\midrule
All rules & 534 & 814 & 646 & 453 & 274 & 2\,148 & 100 \\
\bottomrule
\end{tabular}

%% file: tables/supp_rulecards.tex
\setlength{\tabcolsep}{4pt}
\begin{tabularx}{\textwidth}{@{}>{\raggedright\arraybackslash}p{2.75cm}>{\raggedright\arraybackslash}p{1.75cm}ll>{\raggedright\arraybackslash}X@{}}
\toprule
Rule & Generator & Mech. & Dim. & What the generator does \\
\midrule
\multicolumn{5}{@{}l}{\cellcolor{tableband}\textit{The chain of \cref{sec:structured-knowledge}: roads, terrain, fields, and field boundaries}} \\
Road grades and cut-and-fill & \tld & \mechD{}\,\mechC{} & SO, GM & Each road class has a grade limit, from 5\,\% for the flattest class to 20\,\%; roads are routed within it and graded into the terrain by cut and fill. \\
Road-aligned fields & \tld & \mechD{}\,\mechC{} & SO, GM & Fields grow only on ground below a slope limit, and each field takes the direction of the nearest road as its axis. \\
Field boundaries & \tld & \mechP{}\,\mechD{}\,\mechC{} & CC, SO, GM & Each pair of neighboring parcels draws one boundary form: a hedgerow, which is labeled tree, a dirt track, which is labeled road, a ditch, or none. \\
\multicolumn{5}{@{}l}{\cellcolor{tableband}\textit{Observation}} \\
View law & \tlm & \mechP{}\,\mechD{}\,\mechC{} & SO, GM, OF & Each tile of the tilting scene program draws its view, tilted by up to 15$^\circ$; the ground keeps its map position while raised surfaces lean in proportion to their height, so facades appear; image and labels are rendered by the same camera. \\
Tree shadows & \tlm & \mechP{}\,\mechD{}\,\mechC{} & GM, SA, OF & Under a low sun, the shadow of a crown detaches from its tree and a thin trunk shadow joins it. \\
Sun glint & \tlm & \mechP{}\,\mechD{}\,\mechC{} & SA, OF & Sun glint appears on water only with wind, a high sun, and fine pixels. \\
\multicolumn{5}{@{}l}{\cellcolor{tableband}\textit{Terrain, water, and layout}} \\
Buildings on slopes & \tld & \mechD{}\,\mechC{} & SO, GM, SA & A building on sloping ground is sunk into it, set on a plinth, or terraced, and is left out where the relief is too large. \\
River corridors & \tld & \mechP{}\,\mechD{}\,\mechC{} & CC, SO, SA & Along a river, fields and buildings keep their distance, a belt of riparian trees follows the banks, and the wet strip carries no trees. \\
Heights in context & \tlm & \mechD{}\,\mechC{} & SO, GM & A tall building does not stand alone among low ones: a height that its footprint or its neighbors do not support is swapped with that of a nearby building whose limits admit it, or clipped. \\
Lake level & \tlm & \mechD{}\,\mechC{} & SO, GM & A lake is filled only up to the lowest point of its shore, and the terrain around it is shaped so that its banks stay above the water. \\
Streets meeting water & \tlm & \mechP{}\,\mechD{}\,\mechC{} & CC, SO, GM & Where a street meets water, the kind and width of the water and the rank of the street decide between a culvert, a bridge, and a dead end. \\
Joined roofs & \tlm & \mechD{}\,\mechC{} & SO, GM, OF & Row houses under one continuous roof are separate buildings: each has its own height, roof, and instance label. \\
\multicolumn{5}{@{}l}{\cellcolor{tableband}\textit{Two-date change}} \\
New neighborhoods & \tlm, two-date & \mechP{}\,\mechD{}\,\mechC{} & CC, SO, SA & A new neighborhood appears as a whole: where the later date shows streets, houses, and yards, the earlier date shows the land before its development. \\
Whole-building change & \tlm, two-date & \mechD{}\,\mechC{} & GM, OF & Buildings appear and disappear whole: a tower with its podium, a house with its garage. \\
\bottomrule
\end{tabularx}

%% file: tables/supp_provenance.tex
\setlength{\tabcolsep}{4pt}
\begin{tabularx}{\textwidth}{@{}>{\raggedright\arraybackslash}p{5.3cm}>{\raggedright\arraybackslash}p{1.75cm}>{\raggedright\arraybackslash}X>{\raggedright\arraybackslash}p{4.3cm}@{}}
\toprule
Quantity or method & Generator & Source & How it enters \\
\midrule
\multicolumn{4}{@{}l}{\cellcolor{tableband}\textit{Public geospatial data and statistics}} \\
Land-cover, building-footprint, and road shares per GSD band (nadir scene program) & \tlm & WorldCover~\cite{worldcover2021}, GHSL~\cite{ghsl2023,pesaresi2024ghsl}, Global Surface Water~\cite{pekel2016gsw}, OpenStreetMap (OSM)~\cite{osm} & Priors of scene composition \\
Building-height distribution and its coupling to footprint area & \tlm & 3D-GloBFP~\cite{che2024globfp}, OpenStreetMap height tags~\cite{osm} & Prior of building heights; dependency of height on footprint area \\
Tree heights and crown-to-height ratios & \tlm & Canopy-height map~\cite{tolan2024chm}, Tallo allometry database~\cite{jucker2022tallo} & Prior of tree heights; dependency of crown size on height \\
Share of buildings aligned with roads (tilting scene program) & \tlm & OpenStreetMap~\cite{osm}, Overture~\cite{overture} & Prior of building placement along roads \\
Revisit cycles, urban growth, construction durations, tree mortality, and minimum mapped building area & \tlm, two-date & NAIP~\cite{naip}, urban-expansion atlas~\cite{blei2018expansion}, GHSL~\cite{ghsl2023}, Natural Earth~\cite{naturalearth}, construction statistics~\cite{census_soc,census_c30_length}, street-tree mortality~\cite{roman2011street}, building outlines~\cite{linz_buildings} & Priors of the timing, rate, and size of two-date events \\
\multicolumn{4}{@{}l}{\cellcolor{tableband}\textit{Published methods and definitions}} \\
Growth of the road network & \tld & Parish and M\"uller~\cite{parish2001cities} & Procedural method \\
River networks as drainage trees with stream order & \tld & Drainage trees~\cite{patel_rivers}, stream order~\cite{strahler1957} & Procedural method \\
Sun position for a drawn solar elevation & \tld & Solar position algorithm~\cite{reda2004spa} & Analytical relation \\
Land-cover classes of the labels & both & OpenEarthMap~\cite{openearthmap} & Class definitions \\
\bottomrule
\end{tabularx}

%% file: tables/supp_pools.tex
\begin{tabular}{@{}lrccrrrrrrrr@{}}
\toprule
 & & & & \multicolumn{8}{c}{Labeled pixels (\%)} \\
\cmidrule(l){5-12}
Pool & Samples & GSD (m), median [p5, p95] & Tilted (\%) & Bare & Rangeland & Developed & Road & Tree & Water & Agriculture & Building \\
\midrule
\tld & 30\,000 & 0.47 [0.12, 0.76] & 0.0 & 0.8 & 22.5 & 18.3 & 11.1 & 20.2 & 4.7 & 12.4 & 9.9 \\
\tlm & 26\,985 & 0.32 [0.11, 0.89] & 24.8 & 1.3 & 31.9 & 16.0 & 11.2 & 15.2 & 3.6 & 5.2 & 15.5 \\
\tlmix & 53\,970 & 0.42 [0.12, 0.80] & 12.4 & 1.1 & 27.3 & 17.1 & 11.1 & 17.7 & 4.2 & 8.8 & 12.7 \\
\bottomrule
\end{tabular}

%% file: supp/s2_protocols.tex
\section{Experimental protocols}
\label{sec:s-data}
\label{sec:s-protocols}

This section describes the evaluation datasets, training conditions, adaptation procedure, and reporting conventions used throughout the experiments.

\subsection{Evaluation data and metrics}
\label{sec:s-datasets}

Reported results are evaluated on the designated TEST splits (\cref{tab:s-datasets}), using consistent evaluation protocols across training sources.
Land-cover mapping is evaluated on the eight OpenEarthMap classes and on the six classes shared with LoveDA (building, road, water, agriculture, barren as bare land, and forest as tree), excluding LoveDA background pixels.
Height estimation is evaluated by mean absolute error (MAE), averaged equally across 13 geographic domains (\cref{tab:s-hdomains}).
Building change detection uses change-class F1 on LEVIR-CD, WHU-CD, and SECOND, with S2Looking and LEVIR-CD+ reported separately.
Building instance segmentation uses COCO mask AP~\cite{coco} with up to 1\,500 detections per image.
Task-level scores are unweighted means across their corresponding evaluation datasets.

\begin{table}[!htb]
\caption{\textbf{Real-world evaluation datasets and splits.} Evaluation subsets, sample counts, units, and metrics for the four downstream tasks and the on-demand category extension (\cref{sec:s-ondemand}). LoveDA and iSAID validation serve as TEST because official test labels are unavailable. S2Looking and LEVIR-CD+ are reported separately from the main change-detection mean.}
\label{tab:s-datasets}
\label{tab:s-cdsplits}
\centering
\footnotesize
\input{tables/supp_datasets}
\end{table}

\subsection{Training conditions and recipes}
\label{sec:s-recipes}
\label{sec:s-training}

\Cref{tab:regimes} distinguishes the training conditions by their use of real imagery and annotations.
\reg{SYN} uses synthetic supervision alone, while \reg{SYN+UD} additionally incorporates unlabeled target training images through pseudo-label distillation~\cite{noisystudent} (unsupervised domain adaptation, \uda).
\reg{REAL@p} and \reg{REAL@p+SYN} use prescribed fractions of real training annotations.
Conditional-generator conditions are distinguished by whether real data are needed only to train the generator (\reg{GEN-SO}) or also at generation time (\reg{GEN-RC}).

Within each controlled comparison, downstream architectures, training schedules, and checkpoint readouts are kept consistent, except for the intervention being evaluated (\cref{tab:recipes}).
Unless otherwise specified, synthetic training images from our sources and from the synthetic baselines of \cref{tab:main} receive one label-preserving photometric and sensor transform~\cite{tobin2017domain} per tile, seeded by the tile name: per-channel white-balance gains of 0.9--1.1, log-uniform exposure of 0.75--1.35, gamma of 0.75--1.35, contrast of 0.75--1.3, saturation of 0.6--1.4, hue rotation up to 8$^\circ$, Gaussian blur with probability 0.3 ($\sigma$ in 0.3--1.2~px), Gaussian noise with a standard deviation of up to 3 digital numbers, and JPEG compression with probability 0.3 at quality 55--90.

The procedural baselines of \cref{tab:main} are trained with the same recipes and the same number of updates as our sources.
SynRS3D and SyntheWorld release their land-cover labels in the eight OEM classes, which are used as released, without remapping; for land cover (and, for SynRS3D, height) each is trained on a draw of its tiles matched in size to \tlm's pool, with the same number of updates as every other source; \cref{tab:s-size} adds SynRS3D with its full release and \tlmix restricted to the size of \tlm's pool.
For building change, every training pair of theirs that contains labeled change is used: 61\,485 pairs for SynRS3D and 30\,062 for SyntheWorld.

The rows of \cref{tab:main} train land cover and height for 24k updates.
Controlled experiments need a separately generated training set for every condition, each trained with three seeds: the rule ablations of \cref{sec:s-mabl}, for example, generate one world of 8\,000 tiles per condition.
To keep this training budget bounded, the controlled experiments use these smaller training sets and 8k updates, the schedule that the scarce-label comparisons of \cref{sec:s-real-labels} share (\cref{tab:recipes}).

\begin{table}[!htb]
\caption{\textbf{Training conditions and real-data requirements.} Downstream supervision and the use of real images or annotations for each training condition, beyond the pretrained backbone of the downstream model (\cref{tab:recipes}).}
\label{tab:regimes}
\centering
\footnotesize
\setlength{\tabcolsep}{3pt}
\begin{tabular}{@{}l>{\raggedright\arraybackslash}p{3.05cm}>{\raggedright\arraybackslash}p{3.05cm}@{}}
\toprule
Condition & Downstream supervision & Real data used \\
\midrule
\reg{SYN} & synthetic images and labels & no real annotations \\
\reg{SYN+UD} & synthetic labels and pseudo-labels & unlabeled target training images \\
\reg{REAL@p} & $p$\,\% of real training labels & real training labels \\
\reg{REAL@p+SYN} & real and synthetic labels, jointly or by fine-tuning & real training labels \\
\reg{XREAL} & real damage labels of xBD (\cref{sec:s-damage}) & labels of another dataset \\
\reg{GEN-SO} & generated images of synthetic layouts & generator training \\
\reg{GEN-SO+UD} & as \reg{GEN-SO}, plus pseudo-labels & generator training; unlabeled target training images \\
\reg{GEN-RC} & generated images of real conditions & generator training; images or label maps at generation \\
\reg{ZS} & none (pretrained model applied directly) & pretraining of the applied model \\
\bottomrule
\end{tabular}
\end{table}

\begin{table*}[!tp]
\caption{\textbf{Downstream training and adaptation recipes.} Task-specific architectures, optimization settings, training schedules, checkpoint readouts, and pseudo-label distillation configurations. Controlled experiments and scarce-label comparisons use the indicated training schedules unless stated otherwise. $^a$\,DINOv3 ViT-L/16 weights pretrained on LVD-1689M~\cite{dinov3}, for all four tasks.}
\label{tab:recipes}
\centering
\footnotesize
\setlength{\tabcolsep}{3pt}
\input{tables/recipes}
\end{table*}

\subsection{Adaptation and reporting}
\label{sec:s-uda}
\label{sec:s-statistics}

\paragraph{Adaptation with unlabeled target images.}
\uda uses only unlabeled target images from the designated training splits; no TEST image file or TEST label is used.
In each round, frozen teachers produce pseudo-labels, which are combined with synthetic supervision to train new students; \cref{tab:recipes} specifies the rounds, teachers, and pseudo-label rules.

\paragraph{Statistics and reporting.}
Results are averaged over three training seeds unless otherwise noted (four for the Changen2-OEM rows of \cref{tab:main}).
Differences between training conditions are computed using matched seeds before rounding.

%% file: tables/supp_datasets.tex
\setlength{\tabcolsep}{3pt}
\begin{tabular}{@{}llll@{}}
\toprule
Dataset & TEST & Unit & Metric \\
\midrule
\multicolumn{4}{@{}l}{\cellcolor{tableband}\textit{Land cover}} \\
OpenEarthMap (OEM)~\cite{openearthmap} & test 1\,500 & tile & mIoU \\
LoveDA~\cite{loveda} & val 1\,669 & tile & mIoU \\
\multicolumn{4}{@{}l}{\cellcolor{tableband}\textit{Height}} \\
13 domains (\cref{tab:s-hdomains}) & test 12\,399 & tile & MAE (m) \\
\multicolumn{4}{@{}l}{\cellcolor{tableband}\textit{Building change}} \\
LEVIR-CD~\cite{levircd} & test 2\,048 & 256-px crop & F1 \\
WHU-CD~\cite{whubuildings} & test 2\,640 & 256-px crop & F1 \\
SECOND~\cite{second} & test 1\,694 & pair & F1 \\
S2Looking~\cite{s2looking} & test 1\,000 & pair & F1 \\
LEVIR-CD+~\cite{levircd,s2looking} & test 348 & pair & F1 \\
\multicolumn{4}{@{}l}{\cellcolor{tableband}\textit{Building instances}} \\
xBD~\cite{xbd} & test 780 & image & mask AP \\
WHU~\cite{whubuildings} & test 2\,220 & image & mask AP \\
BRIGHT~\cite{bright} & test 878 & image & mask AP \\
\multicolumn{4}{@{}l}{\cellcolor{tableband}\textit{On-demand categories}} \\
iSAID~\cite{isaid} & val 458 & image & IoU \\
\bottomrule
\end{tabular}

%% file: tables/recipes.tex
\begin{tabular}{@{}l>{\raggedright\arraybackslash}p{3.13cm}>{\raggedright\arraybackslash}p{3.13cm}>{\raggedright\arraybackslash}p{3.13cm}>{\raggedright\arraybackslash}p{3.13cm}@{}}
\toprule
 & Land cover & Height & Change & Instances \\
\midrule
Model$^a$ & DINOv3 ViT-L/16 + DPT~\cite{dpt} & DINOv3 ViT-L/16 + DPT & DINOv3 ViT-L/16 + ChangeMamba decoder~\cite{changemamba} & DINOv3 ViT-L/16 + Mask R-CNN~\cite{maskrcnn} \\
Input size (px) & 512 & 512 & 256 & 1\,024 \\
Batch & 16 & 16 & 8 & 2 \\
Optimizer & AdamW & AdamW & AdamW & AdamW \\
LR (encoder / head) & $10^{-5}$ / $10^{-4}$ & $10^{-5}$ / $10^{-4}$ & $10^{-5}$ / $10^{-4}$ & $5{\times}10^{-6}$ / $5{\times}10^{-5}$ \\
Weight decay & $5{\times}10^{-4}$ & $5{\times}10^{-4}$ & $5{\times}10^{-3}$ & $5{\times}10^{-2}$ \\
Updates & 24k & 24k & 9.5k & 13.4k \\
Schedule & poly (0.9), warm-up 1.5k & poly (0.9), warm-up 1.5k & constant & cosine to $10^{-6}$ \\
Readout & final weights & final weights & weights averaged over the last 6 evaluation points & weights averaged over the last 6 checkpoints \\
Updates, controlled and scarce-label & 8k & 8k & 9.5k & 13.4k \\
\addlinespace[2pt]
\multicolumn{5}{@{}l}{\cellcolor{tableband}\textit{Distillation with unlabeled target images (\uda)}} \\
Unlabeled pool & 2\,310 OEM training tiles & 34\,152 tiles of 11 domains & 14\,214 LEVIR-CD, WHU-CD and SECOND pairs & 11\,678 xBD, WHU and BRIGHT images \\
Rounds & 3 & 2 & 2 & 1 \\
Teachers & flips; round 3: committee of all seeds & flips; scales 0.75 and 1.25 & committee of the seeds, each on 8 views (flips, date orders) & committee of the seeds; 6 views (each seed with and without a horizontal flip) \\
Pseudo-labels & hard class labels, loss weight 1 & mean heights, loss weight 1 & change $\geq$0.7, no change $\leq$0.3, else ignored; components $<$16~px ignored & instances with score $\geq$0.5 matched (mask IoU $\geq$0.5) in $\geq$4 of the 6 views; instances $<$16~px dropped \\
Students & new, main recipe; 50\,\% pseudo-labeled & new, main recipe; 50\,\% pseudo-labeled & new, main recipe; about 47\,\% pseudo-labeled & new, main recipe; about 47\,\% pseudo-labeled \\
\bottomrule
\end{tabular}

%% file: supp/s3_transfer.tex
\section{Transfer and generation efficiency}
\label{sec:s-transfer}
\label{sec:s-results}

This section supports the procedural rows of \cref{tab:main} with results by dataset, height domain, and land-cover class (\cref{sec:s-transfer-data}), other downstream architectures (\cref{sec:s-readers}), photometric randomization and training-set size (\cref{sec:s-components}), predictions on real test images (\cref{sec:s-qualitative}), and generation efficiency (\cref{sec:s-efficiency}).
\Cref{sec:s-generators} compares the conditional generators of \cref{tab:main} under the same protocol.

\subsection{Results by dataset, domain, and class}
\label{sec:s-transfer-data}

\begin{table*}[!tp]
\caption{\textbf{Building-instance segmentation with real and synthetic training data (TEST).} Mask AP and object F1 (detections with a score of at least 0.5, matched one-to-one at a mask IoU of at least 0.5) on xBD, WHU, and BRIGHT, their mean, and two areas of Tokyo (the areas of the height domains Tokyo east and Tokyo west of \cref{tab:s-hdomains}, divided here into 169 image tiles each), with the building footprints of Project PLATEAU~\cite{plateau} as reference; no model was trained or adapted on Tokyo. Upper rows: models trained on one real dataset or on the pooled training splits of the three; the pooled model is the \reg{REAL@100} reference of \cref{fig:s-qual-inst}. Lower rows: the instance rows of \cref{tab:main}. Means of three seeds.}
\label{tab:s-inst-realref}
\centering
\footnotesize
\setlength{\tabcolsep}{4.5pt}
\input{tables/supp_inst_realref}
\end{table*}

\begin{table*}[!t]
\begin{minipage}[t][\dimexpr\textheight-10pt\relax][s]{\textwidth}
\caption{\textbf{Height results by domain (TEST).} Rows: training sources; \tlmix + \uda trains with unlabeled real target images (\reg{SYN+UD}), the other rows on synthetic data only (\reg{SYN}). Columns: the 13 domains with their sources and TEST tile counts, and their equal-weight mean. (a) MAE, the metric of \cref{tab:main}; (b) RMSE and (c) $R^2$, each over all valid pixels of a domain. Means over three seeds; per column and panel, the best value is bold and the second underlined.}
\label{tab:s-hdomains}
{\centering
\input{tables/h_domains}
\par}
\vfill
\centering
\includegraphics[width=\textwidth]{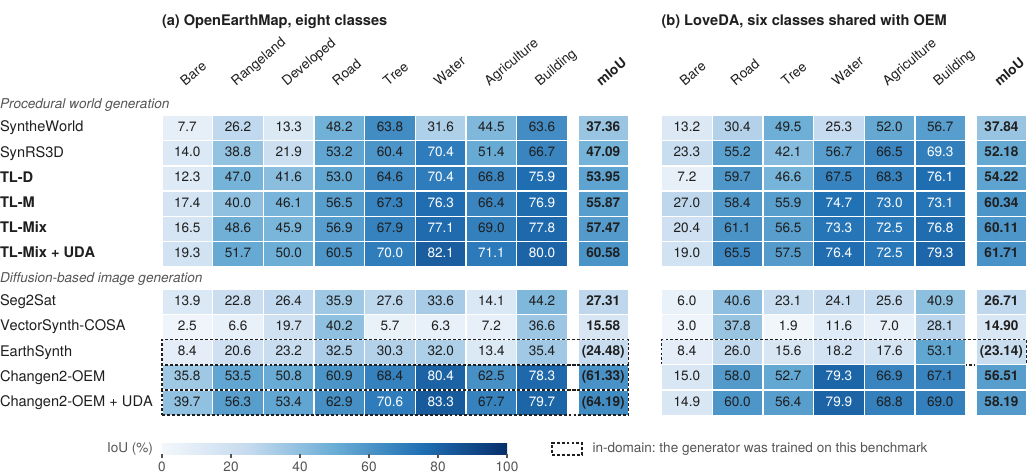}
\captionof{figure}{\textbf{Class-wise land-cover IoU (TEST).} (a) OpenEarthMap, eight classes; (b) LoveDA, the six classes it shares with OEM (background not scored). Rows: the land-cover rows of \cref{tab:main} with their training data, grouped by generation paradigm; each cell is the IoU (\%) averaged over the seeds of \cref{tab:main}, and the last column is its dataset mIoU. One 0--100 color scale serves both panels. Parentheses and dashed frames mark benchmarks used to train the generator, as in \cref{tab:main}.}
\label{fig:s-classiou}
\end{minipage}
\end{table*}

\begin{figure*}[!p]
\centering
\includegraphics[width=\textwidth]{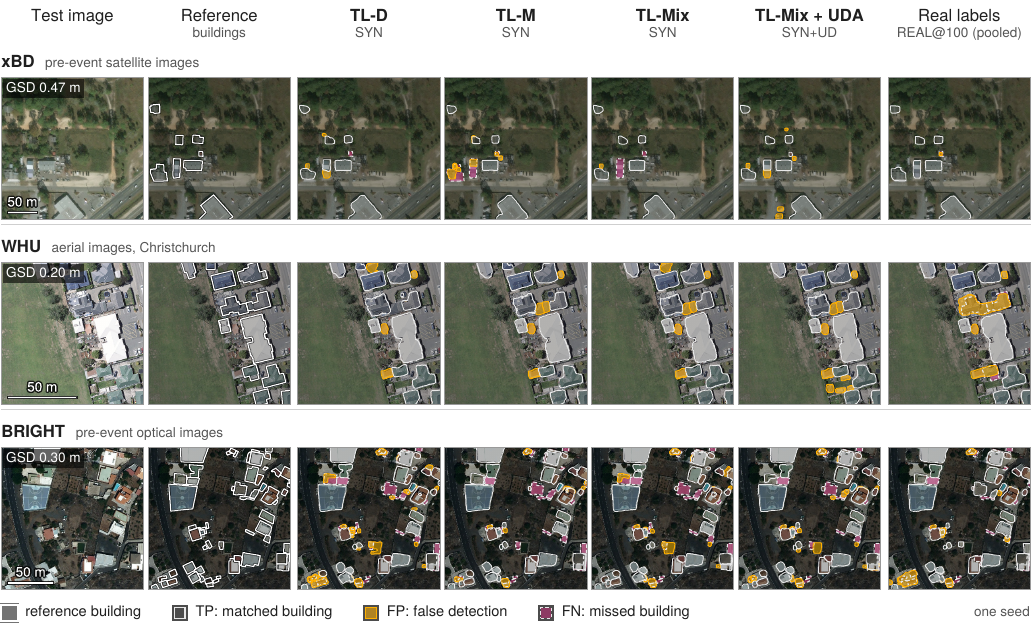}
\caption{\textbf{Building instances on real test images} (TEST, one seed): xBD, WHU, and BRIGHT. Columns: \tld, \tlm, and \tlmix (\reg{SYN}), \tlmix + \uda (\reg{SYN+UD}) and \reg{REAL@100}, trained on the pooled real training labels of the three datasets. One test image per dataset. Overlays: TP, light outline; FP, amber; FN, magenta with a dashed outline. Images: xBD~\cite{xbd} (\mbox{CC BY-NC-SA 4.0}), WHU~\cite{whubuildings} (aerial imagery from the LINZ Data Service, \mbox{CC BY 4.0}) and BRIGHT~\cite{bright} (La Palma: IGN Spain, \mbox{CC BY 4.0}); cropped and dimmed, with our overlays.}
\label{fig:s-qual-inst}
\par\vspace{20pt}
\captionof{table}{\textbf{Land cover by region (TEST, mIoU\,$\uparrow$).} The procedural land-cover rows of \cref{tab:main} with their training data, on the test tiles of each continent of the OEM test split; means over the seeds of \cref{tab:main}. \tlmix + \uda uses \reg{SYN+UD}, the other synthetic rows \reg{SYN}. Last row: the same model trained on all OEM training labels (\reg{REAL@100}, \cref{sec:s-real-labels}), the split's own domain.}
\label{tab:s-continent}
{\centering
\footnotesize
\input{tables/supp_lc_continent}
\par}
\end{figure*}

\begin{figure*}[!p]
\centering
\includegraphics[width=\textwidth]{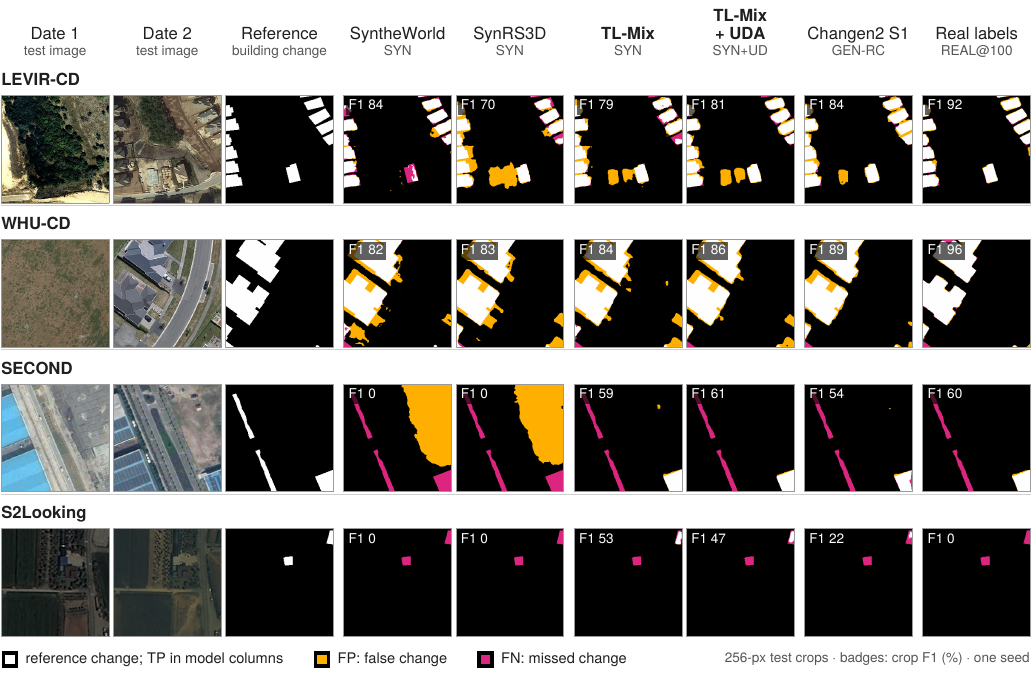}
\caption{\textbf{Building change on real test pairs} (TEST, one seed): LEVIR-CD~\cite{levircd}, WHU-CD~\cite{whubuildings}, SECOND~\cite{second}, and S2Looking~\cite{s2looking}. Columns: SyntheWorld~\cite{syntheworld}, SynRS3D~\cite{synrs3d}, and \tlmix (\reg{SYN}), \tlmix + \uda (\reg{SYN+UD}), Changen2 S1~\cite{changen2} (\reg{GEN-RC}) and \reg{REAL@100}, trained on the real training labels of the row's dataset. One test crop per dataset. Badges: change F1 on the crop, which differs from the pooled dataset F1 of \cref{tab:main}. White, TP; amber, FP; magenta, FN. Images: LEVIR-CD (Google Earth imagery), WHU-CD (second date: aerial imagery from the LINZ Data Service, \mbox{CC BY 4.0}) and S2Looking (GaoFen, SuperView, and BeiJing-2 satellite images).}
\label{fig:s-qual-cd}
\par\vspace{10pt}
\captionof{table}{\textbf{Training-free references on real TEST sets.} (a) Land-cover mapping and (b) building change detection. \reg{ZS} methods are applied without task-specific training; \tlmix trains on synthetic image--label pairs (\reg{SYN}), and \tlmix + \uda additionally uses unlabeled target images (\reg{SYN+UD}). These are different training conditions. The change mean covers LEVIR-CD, WHU-CD, and SECOND; S2Looking (S2L) and LEVIR-CD+ (LEVIR+) are reported separately; CVA, change vector analysis. $^\ast$ROSE-SAM3 uses the prompt bank of its release, which its authors selected on unlabeled OpenEarthMap images.}
\label{tab:s-gen-change}
{\centering
\footnotesize
\setlength{\tabcolsep}{3pt}
\begin{tabular}[t]{@{}lrrr@{}}
\multicolumn{4}{@{}l}{\textit{(a) Land-cover mapping, mIoU\,$\uparrow$}} \\
\toprule
Method & OEM & LoveDA & Mean \\
\midrule
SegEarth-OV~\cite{segearthov} (\reg{ZS}) & 35.70 & 53.44 & 44.57 \\
SegEarth-OV3~\cite{segearthov3} (\reg{ZS}) & 46.79 & 43.58 & 45.19 \\
ProC-SAM3~\cite{procsam3} (\reg{ZS}) & 45.74 & 52.76 & 49.25 \\
ROSE-SAM3$^\ast$~\cite{rosesam3} (\reg{ZS}) & 45.59 & 48.94 & 47.27 \\
\rowcolor{mainhl}\tlmix (\reg{SYN}) & 57.47 & 60.11 & 58.79 \\
\rowcolor{mainhl}\tlmix + \uda (\reg{SYN+UD}) & 60.58 & 61.71 & 61.15 \\
\bottomrule
\end{tabular}
\hfill
\begin{tabular}[t]{@{}lrrrrrr@{}}
\multicolumn{7}{@{}l}{\textit{(b) Building change detection, F1\,$\uparrow$}} \\
\toprule
Method & LEVIR-CD & WHU-CD & SECOND & Mean & S2L & LEVIR+ \\
\midrule
AnyChange~\cite{anychange} (\reg{ZS}) & 18.38 & 21.93 & 38.80 & 26.37 & 11.39 & 17.99 \\
DINOv2-G~\cite{dinov2} + CVA~\cite{cva} (\reg{ZS}) & 18.71 & 13.55 & 34.76 & 22.34 & 5.18 & 14.71 \\
\rowcolor{mainhl}\tlmix (\reg{SYN}) & 78.75 & 75.33 & 71.12 & 75.07 & 53.40 & 77.10 \\
\rowcolor{mainhl}\tlmix + \uda (\reg{SYN+UD}) & 83.45 & 80.94 & 73.93 & 79.44 & 56.18 & 81.45 \\
\bottomrule
\end{tabular}
\par}
\end{figure*}

\Cref{tab:s-inst-realref} sets the synthetic rows of building-instance segmentation beside models trained on real labels, by mask AP and object F1, on the three TEST sets and on two areas of Tokyo that lie outside every training set.
Per class (\cref{fig:s-classiou}), the two generators are complementary: \tlm exceeds \tld on bare land, trees, and water in both datasets, \tld exceeds \tlm on rangeland and on LoveDA's roads and buildings, and \tlmix is at or above the better of the two on 6 of the 8 OEM classes and 3 of the 6 LoveDA classes.
Across height domains (\cref{tab:s-hdomains}), RMSE and $R^2$ order the sources on the 13-domain mean as MAE does; SynRS3D's $R^2$ is negative, that is, below a constant prediction of the domain's mean height, on 5 domains and on the mean, the other sources' only on Vaihingen.
\Cref{tab:s-transfer-full,tab:s-continent} report LEVIR-CD+ and the continents of the OEM test split.
\Cref{tab:s-gen-change} sets the \tlmix models beside training-free methods; \reg{ZS}, \reg{SYN}, and \reg{SYN+UD} are different training conditions.

\begin{table}[!t]
\caption{\textbf{LEVIR-CD+ (TEST).} An additional change set~\cite{levircd,s2looking} for the procedural rows of \cref{tab:main}: change F1, means of three seeds; \tlmix + \uda uses \reg{SYN+UD}, the other rows \reg{SYN}.}
\label{tab:s-transfer-full}
\centering
\footnotesize
\input{tables/supp_transfer}
\end{table}

\subsection{Transfer across downstream architectures}
\label{sec:s-readers}
\enlargethispage{-140pt}

\tlmix achieves the strongest transfer in all eleven task--architecture configurations of \cref{tab:readers}, also against SyntheWorld and SynRS3D wherever they provide labels.
The same holds for the change models on S2Looking, which is reported separately as in \cref{tab:main}.
The relative performance of \tld and \tlm varies by task and architecture.

\begin{table*}[!p]
\caption{\textbf{Transfer across downstream architectures (TEST).} Every row trains each source on its training data of \cref{tab:main} with photometric randomization, for the task's schedule (land cover and height 24\,000 updates, change 9\,500, instances 13\,400), and is evaluated as in \cref{tab:main}; the DINOv3 rows are those of \cref{tab:main}, and R50 is a ResNet-50 encoder~\cite{resnet}, and the ResNet-50 instance model uses SGD with step decay and a batch of 8 images. S2Looking is reported separately, as in \cref{tab:main}. SyntheWorld has no height labels and neither baseline has building instances (--). Mean $\pm$ SD over three seeds; per row, the best source is bold and the second underlined.}
\label{tab:readers}
\centering
\footnotesize
\input{tables/readers}

\vspace{14pt}
\sbox{\SuppTabBoxB}{\begin{minipage}[t]{\columnwidth}
\caption{\textbf{Training-set size (TEST).} Top: the land-cover and height rows of \cref{tab:main} with the number of training samples, SynRS3D's full release, and \tlmix restricted to the size of \tlm's pool (half from each generator). Bottom: building change with the same number of training pairs from each \tl source, and the \tld and \tlmix rows of \cref{tab:main}. Metrics as in \cref{tab:main}, schedules as in \cref{tab:recipes}; means of three seeds.}
\label{tab:s-size}
\centering
\footnotesize
\input{tables/supp_size}
\end{minipage}}%
\sbox{\SuppTabBoxC}{\begin{minipage}[t]{\columnwidth}
\caption{\textbf{Generation throughput of procedural generators.} Rendered images per GPU-hour (one GB200 GPU with its 32 CPU threads), measured for all generators on the same hardware; a two-date sample counts as two images, and the two dates of a \tlm sample are rendered from one constructed scene. \tlm uses no GPU (about 23 CPU thread-seconds per image); \tlmix takes half of its images from each generator. One-time rule specification and preparation of reusable assets and textures are excluded.}
\label{tab:cost}
\label{tab:generators}
\centering
\footnotesize
\input{tables/cost}
\end{minipage}}%
\noindent\usebox{\SuppTabBoxB}\hfill\usebox{\SuppTabBoxC}
\end{table*}

\subsection{Photometric randomization and training-set size}
\label{sec:s-components}

\paragraph{Photometric randomization.}
\label{sec:s-photometric}
Training \tlmix with photometric randomization instead of without it, on the main schedules of \cref{tab:recipes} and paired by seed, changes land-cover mIoU by \ensuremath{+0.45}, height MAE by \ensuremath{-0.025}~m (lower is better), change F1 by \ensuremath{+0.95}, and instance AP by \ensuremath{+2.09} on TEST, an improvement on all four tasks.
The baselines also gain from it in land cover and building change: without photometric randomization, SynRS3D and SyntheWorld lose 2.10 and 2.41 land-cover mIoU and 3.41 and 2.55 change F1, whereas SynRS3D's height MAE is 0.027~m lower without it.

\paragraph{Training-set size.}
\Cref{tab:s-size} varies the size of the training set at the main schedules of \cref{tab:recipes}.
SynRS3D with its full release, more than twice the matched draw, reaches 49.21 mIoU and 3.881\,m (matched draw: 49.64 and 3.945\,m).
With half of its samples, the same number as \tlm's pool, \tlmix reaches 58.49 mIoU and 2.692\,m (full pool: 58.79 and 2.693\,m) and remains ahead of both \tld and \tlm.
For building change, with the same number of pairs from each \tl source, the order of \cref{tab:main} is unchanged: 68.72 F1 for \tlm, 72.51 for \tld, and 74.95 for \tlmix.

\subsection{Qualitative predictions on real test images}
\label{sec:s-qualitative}

\begin{figure*}[!tp]
\centering
\includegraphics[width=\textwidth]{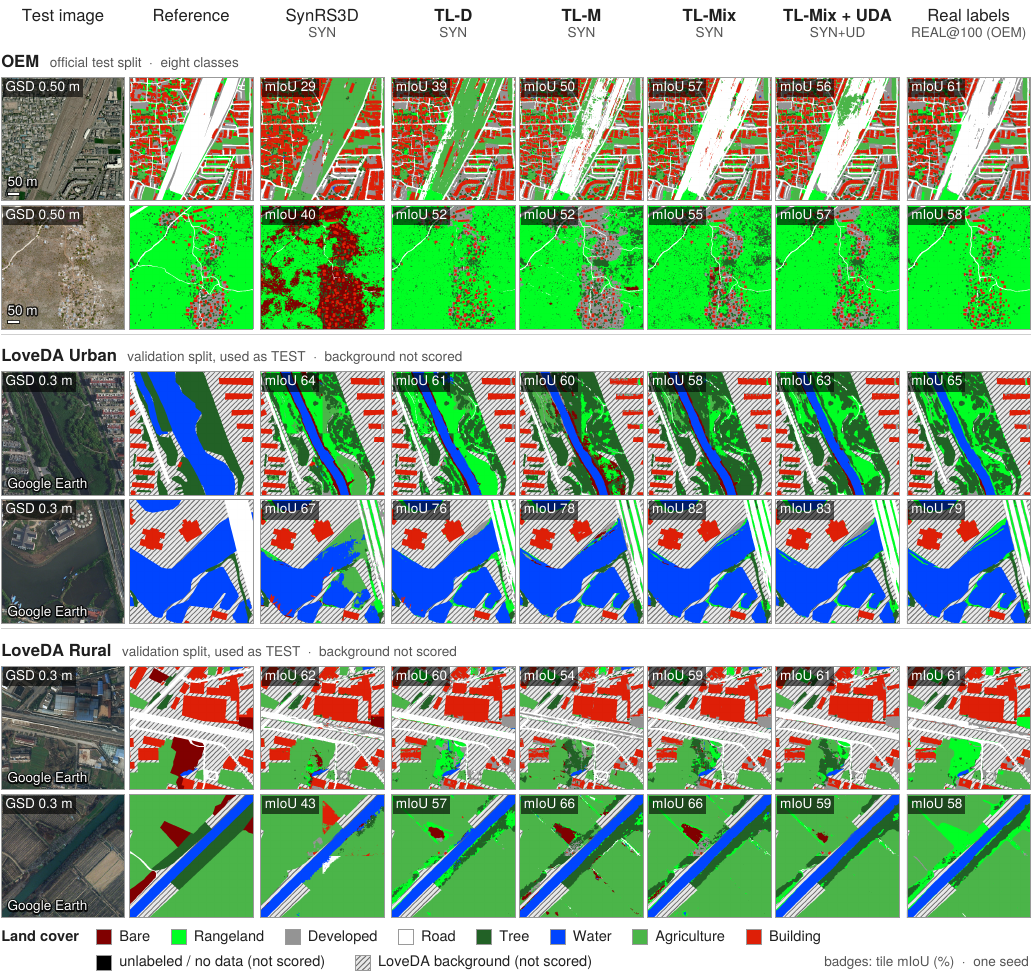}
\caption{\textbf{Land cover on real test tiles} (TEST, one seed): OEM~\cite{openearthmap} and LoveDA~\cite{loveda} (Urban, Rural). Columns: SynRS3D~\cite{synrs3d}, \tld, \tlm, and \tlmix (\reg{SYN}), \tlmix + \uda (\reg{SYN+UD}) and \reg{REAL@100}, trained on OEM's real training labels and applied to LoveDA across datasets. Two test tiles per dataset or domain. Badges: per-tile mIoU, which differs from the dataset mIoU of \cref{tab:main}. Colors: the eight OEM classes; LoveDA is scored on the six classes it shares with OEM, its background hatched (not scored). Images: OEM (Kyoto: Geospatial Information Authority of Japan, under its terms of use; Zanzibar: Commission for Lands, Revolutionary Government of Zanzibar, and GFDRR Open Cities AI, \mbox{CC BY 4.0}) and LoveDA (\mbox{CC BY-NC-SA 4.0}; Google Earth imagery); cropped and resampled.}
\label{fig:s-qual-lc}
\end{figure*}

\begin{figure*}[!p]
\centering
\includegraphics[width=\textwidth]{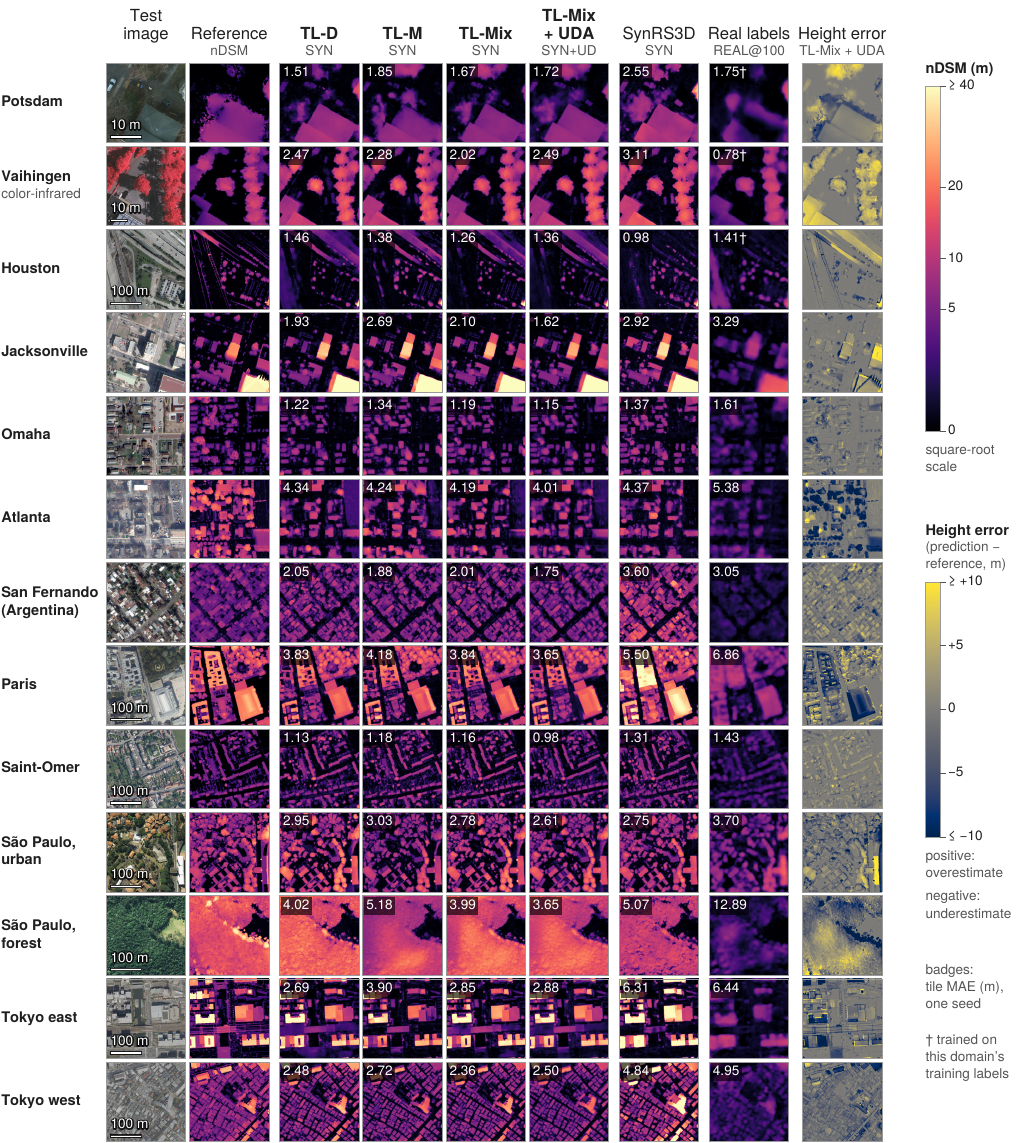}
\caption{\textbf{Height on real test images} (TEST, one seed): one test tile per domain of \cref{tab:s-hdomains}. Columns: \tld, \tlm, and \tlmix (\reg{SYN}), \tlmix + \uda (\reg{SYN+UD}), SynRS3D~\cite{synrs3d} (\reg{SYN}), and \reg{REAL@100}, trained on the real training labels of Potsdam, Vaihingen, and Houston (\dag). nDSM on a square-root scale, saturated from 40\,m. Error map: \tlmix + \uda prediction minus reference, in meters. Badges: the tile's MAE in meters, which differs from the domain MAE of \cref{tab:s-hdomains}. Scale bars are drawn on the tiles that carry a projected georeference. Test images: ISPRS~\cite{isprs}, DFC18~\cite{dfc18}, US3D~\cite{us3d}, OGC~\cite{ogc,ogc2021} (Atlanta: SpaceNet Partners, \mbox{CC BY-SA 4.0}~\cite{weir2019spacenet}) and the domains of~\cite{song2026sparse}: IGN BD~ORTHO\textsuperscript{\textregistered}~\cite{ignbdortho} (Paris, Saint-Omer; Licence Ouverte~2.0), GeoSampa Ortofoto 2020~\cite{geosampa} (S\~{a}o Paulo; \mbox{CC BY-SA 4.0}) and Tokyo Digital Twin Project~\cite{tokyodt} (Tokyo; \mbox{CC BY 4.0}); cropped and resampled.}
\label{fig:s-qual-h}
\end{figure*}

\Cref{fig:s-qual-inst,fig:s-qual-lc,fig:s-qual-h,fig:s-qual-cd} provide qualitative predictions on real TEST images across the four downstream tasks.
They show \tlmix, its adaptation with unlabeled target images (\uda), and synthetic- or real-trained references; the instance, land-cover, and height figures also show the individual \tl generators.
While the synthetic-trained models recover many objects and scene structures, their instance predictions exhibit more false positives than the real-trained reference.

%% file: tables/supp_inst_realref.tex
\begin{tabular}{@{}lrrrrrr@{\hspace{14pt}}rrrrrr@{}}
\toprule
 & \multicolumn{6}{c}{Mask AP\,$\uparrow$} & \multicolumn{6}{c}{Object F1\,$\uparrow$} \\
\cmidrule(lr){2-7}\cmidrule(l){8-13}
Training data & xBD & WHU & BRIGHT & Mean & \makecell[r]{Tokyo\\east} & \makecell[r]{Tokyo\\west} & xBD & WHU & BRIGHT & Mean & \makecell[r]{Tokyo\\east} & \makecell[r]{Tokyo\\west} \\
\midrule
\multicolumn{13}{@{}l}{\cellcolor{tableband}\textit{Real training labels (\reg{REAL@100})}} \\
xBD only & 37.57 & 59.17 & 25.77 & 40.84 & 7.60 & 12.17 & 70.00 & 64.81 & 59.58 & 64.80 & 34.94 & 49.84 \\
WHU only & 28.26 & 72.02 & 19.37 & 39.88 & 4.23 & 8.89 & 64.18 & 92.02 & 46.56 & 67.59 & 11.20 & 17.53 \\
BRIGHT only & 31.40 & 56.79 & 35.73 & 41.30 & 7.23 & 11.71 & 63.18 & 62.72 & 67.02 & 64.31 & 36.09 & 51.48 \\
All three pooled & 37.32 & 69.61 & 33.90 & 46.94 & 7.38 & 12.65 & 69.71 & 89.48 & 66.21 & 75.13 & 31.83 & 47.43 \\
\midrule
\multicolumn{13}{@{}l}{\cellcolor{tableband}\textit{Synthetic training data (\reg{SYN}; \uda: \reg{SYN+UD})}} \\
\tld & 28.77 & 56.09 & 26.33 & 37.07 & 7.15 & 11.86 & 56.96 & 68.48 & 55.98 & 60.48 & 36.47 & 50.53 \\
\tlm & 23.18 & 53.62 & 24.00 & 33.60 & 7.39 & 11.70 & 50.04 & 72.39 & 50.93 & 57.78 & 39.17 & 53.44 \\
\tlmix & 27.82 & 56.78 & 26.74 & 37.11 & 8.01 & 12.69 & 54.38 & 66.92 & 54.38 & 58.56 & 39.74 & 53.28 \\
\tlmix + \uda & 27.99 & 55.82 & 27.29 & 37.03 & 8.15 & 12.19 & 52.59 & 62.79 & 53.51 & 56.30 & 40.68 & 53.58 \\
\bottomrule
\end{tabular}

%% file: tables/h_domains.tex
\footnotesize
\setlength{\tabcolsep}{3pt}
\begin{tabular}{@{}lrrrrrrrrrrrrr@{\hspace{8pt}}r@{}}
\toprule
 & \rotatebox{90}{\makecell[l]{Potsdam\\\cite{isprs}}} & \rotatebox{90}{\makecell[l]{Vaihingen\\\cite{isprs}}} & \rotatebox{90}{\makecell[l]{Houston\\\cite{dfc18}}} & \rotatebox{90}{\makecell[l]{Jacksonville\\\cite{us3d}}} & \rotatebox{90}{\makecell[l]{Omaha\\\cite{us3d}}} & \rotatebox{90}{\makecell[l]{Atlanta\\\cite{ogc}}} & \rotatebox{90}{\makecell[l]{San Fernando\\(Argentina)\\\cite{ogc2021}}} & \rotatebox{90}{\makecell[l]{Paris\\\cite{song2026sparse,ignbdortho,ignlidarhd}}} & \rotatebox{90}{\makecell[l]{Saint-Omer\\\cite{song2026sparse,ignbdortho,ignlidarhd}}} & \rotatebox{90}{\makecell[l]{S\~{a}o Paulo, urban\\\cite{song2026sparse,geosampa}}} & \rotatebox{90}{\makecell[l]{S\~{a}o Paulo, forest\\\cite{song2026sparse,geosampa}}} & \rotatebox{90}{\makecell[l]{Tokyo east\\\cite{song2026sparse,tokyodt}}} & \rotatebox{90}{\makecell[l]{Tokyo west\\\cite{song2026sparse,tokyodt}}} & \rotatebox{90}{\makecell[l]{\textbf{Mean of}\\\textbf{the 13}}} \\
TEST tiles & 1\,152 & 148 & 23 & 1\,083 & 1\,946 & 2\,251 & 5\,407 & 56 & 76 & 63 & 65 & 48 & 81 & \\
\midrule
\multicolumn{15}{@{}l}{\cellcolor{tableband}\textit{(a) MAE (m)\,$\downarrow$}} \\
SynRS3D~\cite{synrs3d} & 5.264 & 4.310 & 1.864 & 3.225 & 2.175 & 6.092 & 2.845 & 5.361 & 1.971 & 3.797 & 4.887 & 4.947 & 4.552 & 3.945 \\
\tld & 2.094 & 2.629 & 1.317 & 2.492 & 1.599 & 5.168 & 2.100 & 4.027 & 1.589 & 3.223 & 4.955 & \underline{3.198} & 2.878 & 2.867 \\
\tlm & 2.587 & 2.640 & 1.346 & 2.447 & 1.510 & 4.925 & 2.098 & \underline{3.937} & 1.664 & 3.461 & 4.503 & 3.511 & 3.062 & 2.899 \\
\tlmix & \underline{2.070} & \textbf{2.338} & \underline{1.247} & \underline{2.291} & \underline{1.451} & \underline{4.769} & \underline{1.977} & 3.966 & \underline{1.536} & \underline{3.180} & \underline{4.157} & 3.228 & \underline{2.800} & \underline{2.693} \\
\tlmix + \uda & \textbf{2.042} & \underline{2.558} & \textbf{1.192} & \textbf{2.144} & \textbf{1.370} & \textbf{4.635} & \textbf{1.811} & \textbf{3.817} & \textbf{1.469} & \textbf{3.023} & \textbf{3.820} & \textbf{3.052} & \textbf{2.700} & \textbf{2.587} \\
\multicolumn{15}{@{}l}{\cellcolor{tableband}\textit{(b) RMSE (m)\,$\downarrow$}} \\
SynRS3D~\cite{synrs3d} & 10.962 & 7.971 & 5.204 & 7.339 & 5.167 & 11.299 & 4.777 & 10.189 & 3.998 & 6.983 & 7.294 & 9.311 & 8.290 & 7.599 \\
\tld & 4.184 & 4.307 & 2.804 & 5.408 & 3.777 & 9.785 & 3.433 & \underline{6.399} & 3.549 & \underline{6.156} & 6.800 & \underline{6.779} & 5.662 & 5.311 \\
\tlm & 4.909 & 4.303 & 2.893 & 5.375 & 3.570 & 9.808 & 3.446 & 6.463 & 3.606 & 6.586 & 6.677 & 7.969 & 6.469 & 5.544 \\
\tlmix & \underline{4.045} & \textbf{3.823} & \underline{2.693} & \underline{5.111} & \underline{3.492} & \underline{9.335} & \underline{3.297} & 6.416 & \underline{3.395} & 6.212 & \underline{6.084} & 7.035 & \underline{5.623} & \underline{5.120} \\
\tlmix + \uda & \textbf{3.990} & \underline{4.110} & \textbf{2.572} & \textbf{4.805} & \textbf{3.318} & \textbf{8.979} & \textbf{3.077} & \textbf{6.053} & \textbf{3.236} & \textbf{5.763} & \textbf{5.748} & \textbf{6.763} & \textbf{5.425} & \textbf{4.911} \\
\multicolumn{15}{@{}l}{\cellcolor{tableband}\textit{(c) $R^2$\,$\uparrow$}} \\
SynRS3D~\cite{synrs3d} & $-$2.88 & $-$3.78 & $-$0.23 & 0.18 & 0.10 & 0.33 & $-$0.19 & $-$0.11 & 0.63 & 0.58 & 0.07 & 0.40 & 0.22 & $-$0.36 \\
\tld & 0.44 & $-$0.40 & 0.64 & 0.55 & 0.52 & 0.50 & 0.39 & \underline{0.56} & 0.71 & \underline{0.68} & 0.19 & \textbf{0.68} & \underline{0.64} & 0.47 \\
\tlm & 0.22 & $-$0.39 & 0.62 & 0.56 & 0.57 & 0.49 & 0.38 & 0.55 & 0.70 & 0.63 & 0.22 & 0.56 & 0.53 & 0.43 \\
\tlmix & \underline{0.47} & \textbf{$-$0.10} & \underline{0.67} & \underline{0.60} & \underline{0.59} & \underline{0.54} & \underline{0.43} & \underline{0.56} & \underline{0.73} & 0.67 & \underline{0.36} & \underline{0.66} & \underline{0.64} & \underline{0.53} \\
\tlmix + \uda & \textbf{0.49} & \underline{$-$0.27} & \textbf{0.70} & \textbf{0.65} & \textbf{0.63} & \textbf{0.58} & \textbf{0.51} & \textbf{0.61} & \textbf{0.76} & \textbf{0.72} & \textbf{0.42} & \textbf{0.68} & \textbf{0.67} & \textbf{0.55} \\
\bottomrule
\end{tabular}

%% file: tables/supp_transfer.tex
\setlength{\tabcolsep}{3pt}
\begin{tabular}{@{}lr@{}}
\toprule
Training data & LEVIR-CD+ F1\,$\uparrow$ \\
\midrule
SyntheWorld~\cite{syntheworld} & 69.40 \\
SynRS3D~\cite{synrs3d} & 66.49 \\
\rowcolor{mainhl}\tld & 76.86 \\
\rowcolor{mainhl}\tlm & 68.92 \\
\rowcolor{mainhl}\tlmix & 77.10 \\
\rowcolor{mainhl}\tlmix + \uda & 81.45 \\
\bottomrule
\end{tabular}

%% file: tables/supp_lc_continent.tex
\begin{tabular}{@{}lcccccc@{}}
\toprule
Training data & Africa & Asia & Europe & \makecell{North\\America} & Oceania & \makecell{South\\America} \\
\midrule
SyntheWorld~\cite{syntheworld} & 35.74 & 37.42 & 38.81 & 36.15 & 36.41 & 35.67 \\
SynRS3D~\cite{synrs3d} & 42.35 & 48.19 & 48.57 & 45.21 & 49.01 & 45.35 \\
\midrule
\rowcolor{mainhl}\tld & 50.75 & 54.74 & 55.33 & 53.63 & 55.41 & 49.32 \\
\rowcolor{mainhl}\tlm & 52.20 & 56.37 & 56.60 & 53.44 & 55.88 & 52.51 \\
\rowcolor{mainhl}\tlmix & 53.95 & 58.05 & 58.16 & 56.16 & 56.79 & 54.13 \\
\rowcolor{mainhl}\tlmix + \uda & 56.68 & 62.22 & 60.84 & 59.79 & 59.89 & 56.79 \\
\midrule
OEM training labels & 66.08 & 69.33 & 70.80 & 67.92 & 65.71 & 65.91 \\
\bottomrule
\end{tabular}

%% file: tables/readers.tex
\setlength{\tabcolsep}{6pt}
\begin{tabular}{@{}lrrrrr@{}}
\toprule
Downstream architecture & SyntheWorld~\cite{syntheworld} & SynRS3D~\cite{synrs3d} & \tld & \tlm & \tlmix \\
\midrule
\multicolumn{6}{@{}l}{\cellcolor{tableband}\textit{Land cover, mIoU\,$\uparrow$ (OEM and LoveDA)}} \\
DINOv3-L~\cite{dinov3} + DPT~\cite{dpt} & \mbox{37.60\,{\scriptsize$\pm$0.66}} & \mbox{49.64\,{\scriptsize$\pm$1.00}} & \mbox{54.08\,{\scriptsize$\pm$0.41}} & \mbox{\underline{58.10}\,{\scriptsize$\pm$0.22}} & \mbox{\textbf{58.79}\,{\scriptsize$\pm$0.14}} \\
DINOv2-L~\cite{dinov2} + DPT & \mbox{26.98\,{\scriptsize$\pm$1.08}} & \mbox{42.34\,{\scriptsize$\pm$0.95}} & \mbox{49.56\,{\scriptsize$\pm$1.54}} & \mbox{\underline{53.54}\,{\scriptsize$\pm$0.68}} & \mbox{\textbf{55.48}\,{\scriptsize$\pm$0.28}} \\
DeepLabv3+~\cite{deeplabv3plus} R50 & \mbox{22.67\,{\scriptsize$\pm$0.53}} & \mbox{27.23\,{\scriptsize$\pm$0.49}} & \mbox{37.06\,{\scriptsize$\pm$0.87}} & \mbox{\underline{38.43}\,{\scriptsize$\pm$0.65}} & \mbox{\textbf{41.51}\,{\scriptsize$\pm$0.48}} \\
\multicolumn{6}{@{}l}{\cellcolor{tableband}\textit{Height, MAE (m)\,$\downarrow$ (13 domains)}} \\
DINOv3-L + DPT & -- & \mbox{3.945\,{\scriptsize$\pm$0.043}} & \mbox{\underline{2.867}\,{\scriptsize$\pm$0.028}} & \mbox{2.899\,{\scriptsize$\pm$0.007}} & \mbox{\textbf{2.693}\,{\scriptsize$\pm$0.008}} \\
DINOv2-L + DPT & -- & \mbox{4.826\,{\scriptsize$\pm$0.318}} & \mbox{\underline{2.885}\,{\scriptsize$\pm$0.033}} & \mbox{2.950\,{\scriptsize$\pm$0.020}} & \mbox{\textbf{2.656}\,{\scriptsize$\pm$0.009}} \\
U-Net~\cite{unet} R50 & -- & \mbox{5.027\,{\scriptsize$\pm$0.096}} & \mbox{4.396\,{\scriptsize$\pm$0.024}} & \mbox{\underline{4.040}\,{\scriptsize$\pm$0.025}} & \mbox{\textbf{3.824}\,{\scriptsize$\pm$0.036}} \\
\multicolumn{6}{@{}l}{\cellcolor{tableband}\textit{Building change, F1\,$\uparrow$ (LEVIR-CD, WHU-CD, SECOND)}} \\
DINOv3-L + ChangeMamba~\cite{changemamba} & \mbox{67.82\,{\scriptsize$\pm$2.44}} & \mbox{64.08\,{\scriptsize$\pm$3.03}} & \mbox{\underline{72.93}\,{\scriptsize$\pm$0.33}} & \mbox{68.72\,{\scriptsize$\pm$1.78}} & \mbox{\textbf{75.07}\,{\scriptsize$\pm$1.90}} \\
ChangeMamba-Tiny & \mbox{52.09\,{\scriptsize$\pm$2.23}} & \mbox{52.80\,{\scriptsize$\pm$2.21}} & \mbox{\underline{55.75}\,{\scriptsize$\pm$1.03}} & \mbox{45.85\,{\scriptsize$\pm$2.37}} & \mbox{\textbf{55.87}\,{\scriptsize$\pm$1.00}} \\
SiamCRNN~\cite{siamcrnn} R50 & \mbox{43.29\,{\scriptsize$\pm$0.71}} & \mbox{46.65\,{\scriptsize$\pm$1.82}} & \mbox{\underline{48.83}\,{\scriptsize$\pm$1.39}} & \mbox{39.36\,{\scriptsize$\pm$0.88}} & \mbox{\textbf{50.70}\,{\scriptsize$\pm$1.94}} \\
\multicolumn{6}{@{}l}{\cellcolor{tableband}\textit{Building change, F1\,$\uparrow$ (S2Looking)}} \\
DINOv3-L + ChangeMamba & \mbox{39.73\,{\scriptsize$\pm$3.05}} & \mbox{36.96\,{\scriptsize$\pm$8.54}} & \mbox{\underline{50.81}\,{\scriptsize$\pm$2.32}} & \mbox{47.73\,{\scriptsize$\pm$1.12}} & \mbox{\textbf{53.40}\,{\scriptsize$\pm$0.28}} \\
ChangeMamba-Tiny & \mbox{12.09\,{\scriptsize$\pm$6.07}} & \mbox{18.69\,{\scriptsize$\pm$3.99}} & \mbox{\underline{32.59}\,{\scriptsize$\pm$2.17}} & \mbox{25.16\,{\scriptsize$\pm$3.82}} & \mbox{\textbf{37.20}\,{\scriptsize$\pm$0.15}} \\
SiamCRNN R50 & \mbox{12.33\,{\scriptsize$\pm$1.19}} & \mbox{22.87\,{\scriptsize$\pm$5.29}} & \mbox{\underline{29.75}\,{\scriptsize$\pm$0.77}} & \mbox{20.96\,{\scriptsize$\pm$2.02}} & \mbox{\textbf{30.11}\,{\scriptsize$\pm$1.25}} \\
\multicolumn{6}{@{}l}{\cellcolor{tableband}\textit{Building instances, mask AP\,$\uparrow$ (xBD, WHU, BRIGHT)}} \\
DINOv3-L + Mask R-CNN~\cite{maskrcnn} & -- & -- & \mbox{\underline{37.07}\,{\scriptsize$\pm$0.13}} & \mbox{33.60\,{\scriptsize$\pm$0.56}} & \mbox{\textbf{37.11}\,{\scriptsize$\pm$0.23}} \\
Mask R-CNN R50 & -- & -- & \mbox{\underline{22.12}\,{\scriptsize$\pm$0.46}} & \mbox{18.66\,{\scriptsize$\pm$0.60}} & \mbox{\textbf{24.04}\,{\scriptsize$\pm$0.15}} \\
\bottomrule
\end{tabular}

%% file: tables/supp_size.tex
\setlength{\tabcolsep}{4.5pt}
\begin{tabular}{@{}lrrr@{}}
\toprule
Training data & \makecell[r]{Samples\\($\times 10^{3}$)} & \makecell[r]{Land cover\\mIoU\,$\uparrow$} & \makecell[r]{Height\\MAE (m)\,$\downarrow$} \\
\midrule
SynRS3D~\cite{synrs3d} & 27.0 & 49.64 & 3.945 \\
SynRS3D~\cite{synrs3d}, full release & 65.3 & 49.21 & 3.881 \\
\midrule
\tld & 30.0 & 54.08 & 2.867 \\
\tlm & 27.0 & 58.10 & 2.899 \\
\tlmix & 27.0 & 58.49 & 2.692 \\
\tlmix & 54.0 & 58.79 & 2.693 \\
\midrule
Training data & \makecell[r]{Pairs\\($\times 10^{3}$)} & \multicolumn{2}{r}{\makecell[r]{Building change\\F1\,$\uparrow$}} \\
\midrule
\tlm & 8.2 & \multicolumn{2}{r}{68.72} \\
\tld & 8.2 & \multicolumn{2}{r}{72.51} \\
\tlmix & 8.2 & \multicolumn{2}{r}{74.95} \\
\midrule
\tld & 26.2 & \multicolumn{2}{r}{72.93} \\
\tlmix & 16.5 & \multicolumn{2}{r}{75.07} \\
\bottomrule
\end{tabular}

%% file: tables/cost.tex
\setlength{\tabcolsep}{6pt}
\begin{tabular}{@{}lrr@{}}
\toprule
Generator & Images per GPU-hour & Relative to SynRS3D \\
\midrule
\mbox{\tlm}, single-date & 5\,101 & 376$\times$ \\
\mbox{\tlm}, two-date & 5\,189 & 382$\times$ \\
\mbox{\tld} & 131.7 & 9.7$\times$ \\
\tlmix & 256.8 & 19$\times$ \\
SynRS3D & 13.58 & 1$\times$ \\
\bottomrule
\end{tabular}

%% file: supp/s6_cost.tex
\subsection{Generation efficiency}
\label{sec:s-cost}
\label{sec:s-efficiency}

\Cref{tab:cost} compares the generation throughput of the procedural generators, covering scene construction, rendering, and aligned-label export.
All three generators were measured on the same hardware: throughput is the number of rendered images per GPU-hour, where one GPU-hour is one GB200 GPU together with its 32 CPU threads for one hour.
A two-date sample counts as two images: every SynRS3D tile and 23\,\% of all tiles that \tld generated hold two dates.
\tlm requires no GPU computation during generation and uses only the CPU threads, at 74\,\% and 62\,\% CPU utilization for its single-date and two-date programs.
Its generation throughput is approximately 38.7$\times$ that of \tld (single-date program) and 382$\times$ that of SynRS3D (two-date program), and \tld's is 9.7$\times$ that of SynRS3D.
Per image, \tlm takes about 23 CPU thread-seconds and no GPU time, whereas \tld takes about 27 GPU-seconds and SynRS3D about 265 GPU-seconds, each together with the CPU threads of its slot.

SynRS3D was run with its original pipeline and settings, without modification.
All measurements exclude the one-time work of specifying and reviewing rules and of preparing reusable assets and texture libraries.

%% file: supp/s_generators.tex
\twocolumn[{%
\centering
\captionof{table}{\textbf{Generation inputs and real-data requirements.} Bottom: how the eight classes of the supplied layouts are passed to each layout-conditioned generator. $^a$\,Scene rules draw on public data and literature (\cref{sec:s-provenance}); surface textures are prepared with pretrained text-to-image models.}
\label{tab:s-gensettings}\label{tab:capability}
\par\vspace{4pt}
{\footnotesize\input{tables/supp_gensettings}}\par\vspace{16pt}
}]
\section{Comparison with conditional generators}
\label{sec:s-generators}

We compare \tl with conditional image generators in terms of their generation inputs, dependence on real data, and synthetic imagery.
Downstream transfer results are reported in \cref{tab:main}.

\begin{figure*}[!tp]
\centering
\includegraphics[width=\textwidth]{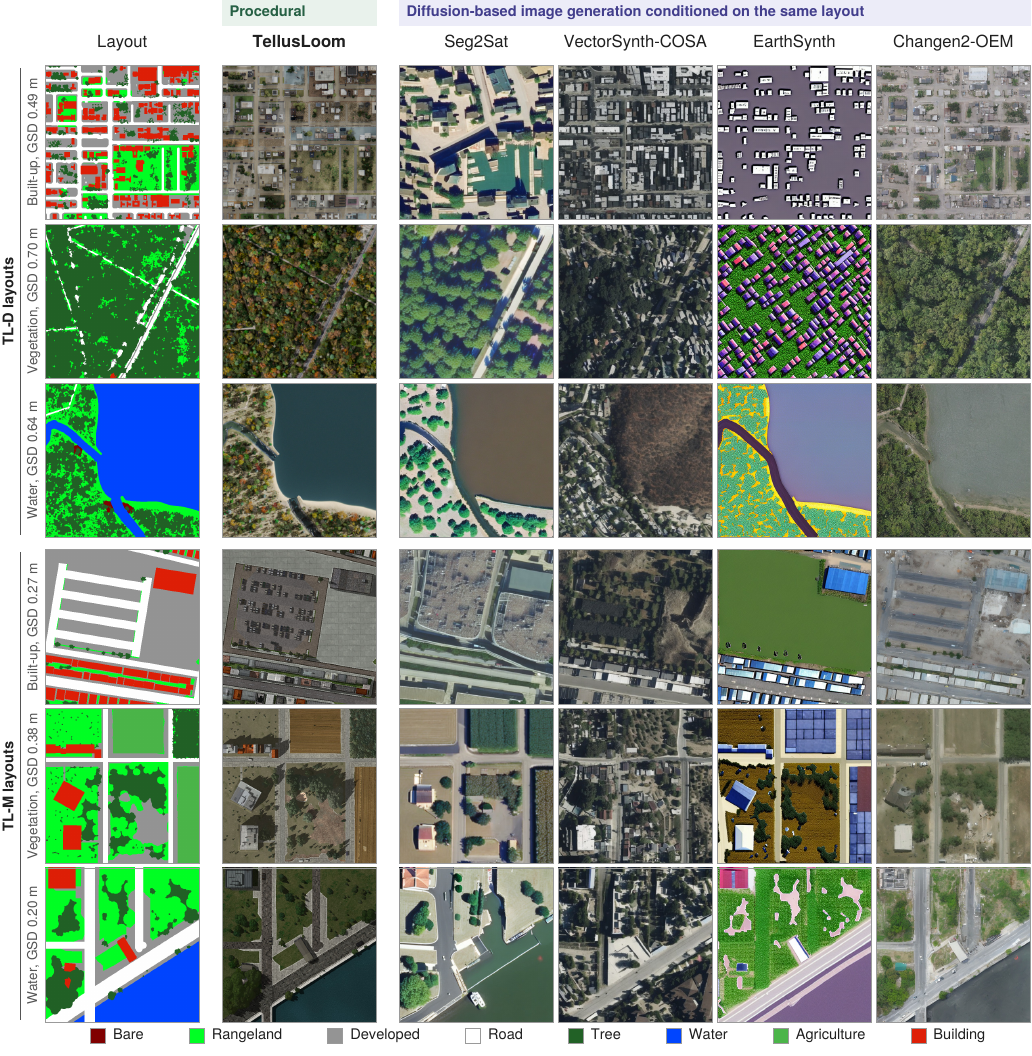}
\caption{\textbf{Synthetic imagery generated from the same semantic layouts.} Each row shows one \tl layout (row label: scene type and GSD) rendered procedurally and by four conditional image generators from conditions derived from that layout. One \tld and one \tlm layout per scene type.}
\label{fig:renderers}
\end{figure*}

\begin{figure*}[!tp]
\centering
\includegraphics[width=\textwidth]{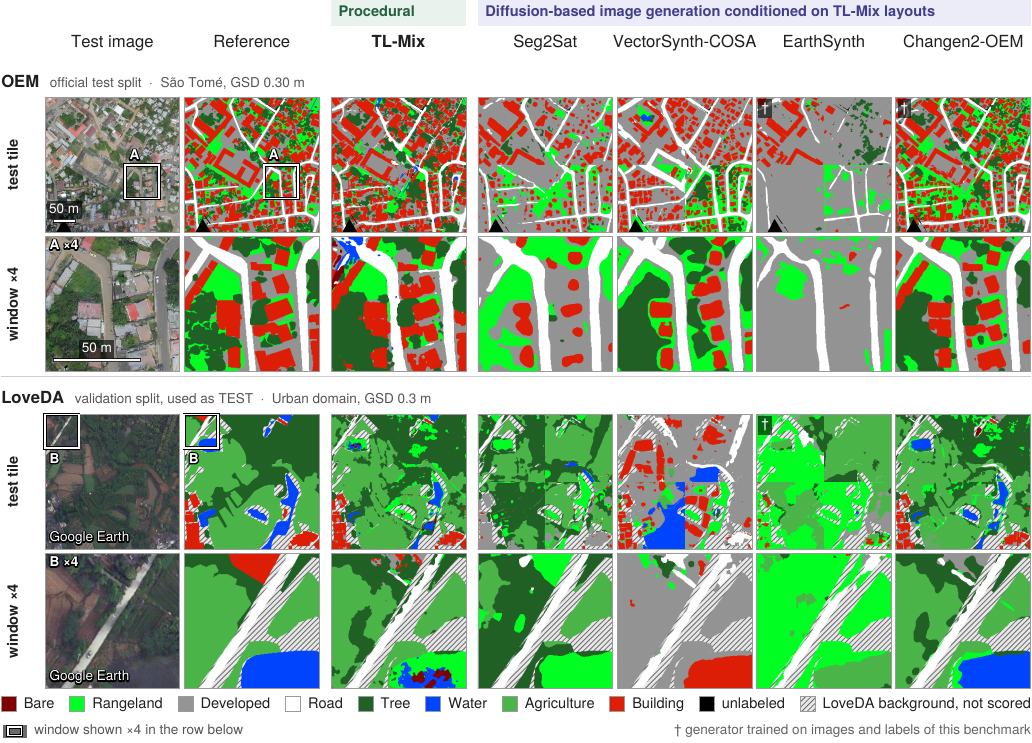}
\caption{\textbf{Land-cover predictions on real test imagery.} Models are trained on \tlmix (\reg{SYN}) or on imagery produced by conditional generators (\reg{GEN-SO}), as evaluated in \cref{tab:main}. One test tile each from OpenEarthMap~\cite{openearthmap} and LoveDA~\cite{loveda}, with an enlarged window. Images: OpenEarthMap (S\~{a}o Tom\'{e}: Drones Adventures, via OpenAerialMap; \mbox{CC BY 4.0}) and LoveDA (\mbox{CC BY-NC-SA 4.0}; Google Earth imagery).}
\label{fig:s-gen-downstream}
\end{figure*}

\begin{figure*}[!tp]
\centering
\includegraphics[height=0.88\textheight]{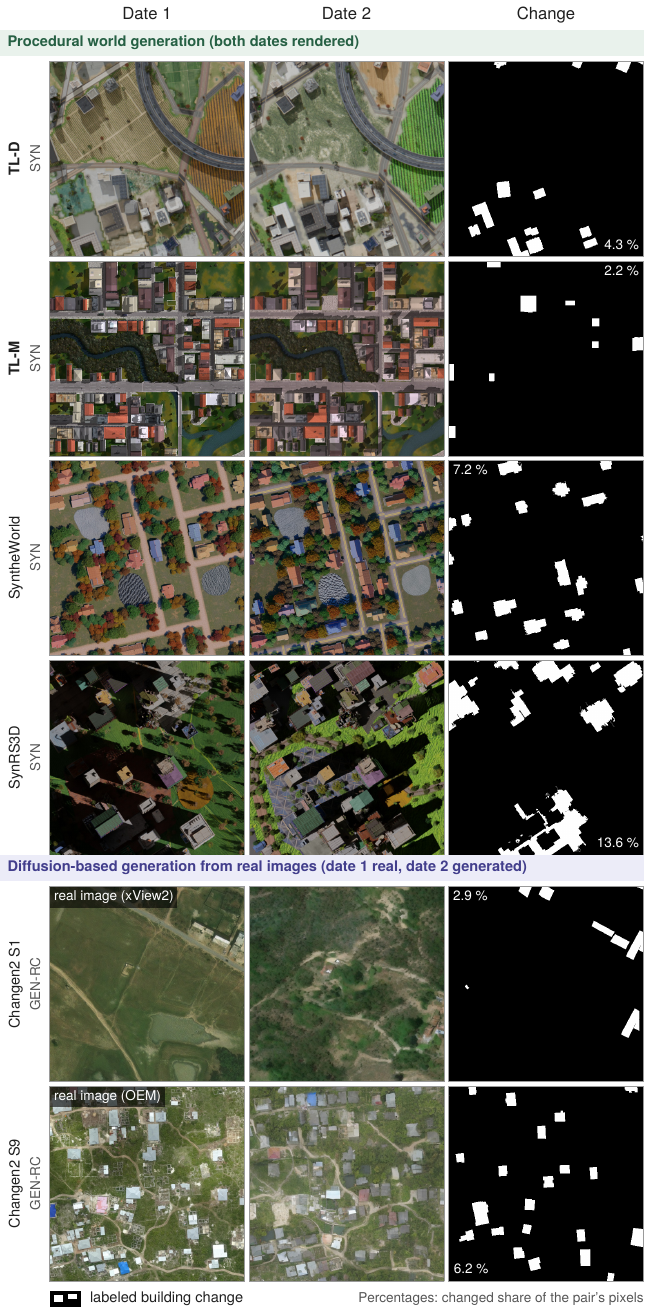}
\caption{\textbf{Building-change training pairs.} Procedural generators synthesize both dates and the change label; Changen2 S1 and S9 generate the second date from a real first-date image and spatial conditions. The examples are not layout-matched. Images: SyntheWorld~\cite{syntheworld} and SynRS3D~\cite{synrs3d} (CC BY-NC 4.0), Changen2~\cite{changen2} S1 and S9 (\mbox{CC BY-NC-SA 4.0}); first dates: xView2 (xBD~\cite{xbd}, \mbox{CC BY-NC-SA 4.0}) and OpenEarthMap~\cite{openearthmap} (Zanzibar: Commission for Lands, Revolutionary Government of Zanzibar, and GFDRR Open Cities AI; \mbox{CC BY 4.0}).}
\label{fig:s-cd-sources}
\end{figure*}

\subsection{Generation setup and real-data requirements}
\label{sec:s-gen-settings}

\Cref{tab:s-gensettings} summarizes the inputs and real-data requirements of the evaluated generators.
\tl constructs scenes and aligned labels procedurally, without training on paired real task annotations.
In contrast, the evaluated conditional generators learn from real imagery paired with semantic labels or geospatial information, and Changen2 S1 and S9 additionally use real images and corresponding conditions during generation.
For the layout-conditioned generators, we use the released models with the sampling settings of \cref{tab:s-gensettings} and adapt the supplied synthetic layouts to their respective conditioning formats: the geometry of a layout is kept, and its eight classes are passed as \cref{tab:s-gensettings} lists.
Each layout-conditioned generator renders one image per label map of a \tlmix training list, 53\,970 images in total; for Changen2 S1 and S9, we use the released datasets.

Two of the generators evaluated on land cover were trained on data from our land-cover benchmarks: Changen2-OEM on OEM images and labels, and EarthSynth on EarthSynth-180K~\cite{earthsynth,earthsynth180k}, whose training split contains OEM training tiles and all LoveDA training tiles and whose validation split contains the OEM and LoveDA validation tiles (the LoveDA validation tiles serve as our LoveDA TEST set).
\Cref{tab:main} sets their scores on these benchmarks in parentheses, and \cref{fig:s-gen-downstream} marks their predictions with a dagger.

\subsection{Qualitative comparison}
\label{sec:s-gen-layouts}
\label{sec:s-gen-change}

\paragraph{Generation from shared semantic layouts.}
\Cref{fig:renderers} compares imagery generated from spatial conditions derived from the same semantic layouts.
The training data of Changen2-OEM and EarthSynth contain the eight classes of these layouts; Seg2Sat and VectorSynth-COSA receive them through a mapping to their own classes or tags (\cref{tab:s-gensettings}).
The methods thus differ in their native conditioning formats and training distributions, and the comparison reflects the evaluated configurations rather than a universal ranking of generation paradigms.
\Cref{fig:s-gen-downstream} shows land-cover predictions on real test imagery from models trained on these sources.

\paragraph{Building-change generation.}
\Cref{fig:s-cd-sources} illustrates building-change training pairs.
Procedural generators synthesize both dates, whereas Changen2 S1 and S9 generate the second date from a real first-date image and spatial conditions.

%% file: tables/supp_gensettings.tex
\setlength{\tabcolsep}{4pt}
\begin{tabular}{@{}>{\raggedright\arraybackslash}p{1.25in}>{\raggedright\arraybackslash}p{1.7in}>{\raggedright\arraybackslash}p{2.2in}>{\raggedright\arraybackslash}p{1.38in}@{}}
\toprule
Method & Generation input & Real data for generator training & Real data at generation \\
\midrule
\tld, \tlm & procedural scene program & no paired real task labels\textsuperscript{a} & none \\
\midrule
Seg2Sat~\cite{seg2sat} & semantic map (FLAIR color map) & FLAIR~\cite{flair1} RGB + labels & none \\
VectorSynth-COSA~\cite{vectorsynth} & geospatial conditions (OSM-tag embedding map) & RGB + OSM vectors & none \\
EarthSynth~\cite{earthsynth} & semantic map (class map) & EarthSynth-180K (OEM, LoveDA, DeepGlobe, SAMRS, part of LAE-1M): RGB + annotations & none \\
Changen2-OEM~\cite{changen2} & land-cover map (one-hot label map) & OEM RGB + labels & none \\
\midrule
Changen2 S1~\cite{changen2} & real RGB + building masks & xView2 pre-disaster RGB + building footprints & xView2 RGB + building masks \\
Changen2 S9~\cite{changen2} & real RGB + land-cover map & OEM RGB + land-cover maps & OEM training RGB + land-cover maps \\
\bottomrule
\multicolumn{4}{@{}p{6.80in}@{}}{\scriptsize\textit{Sampling} with the released weights: Seg2Sat PNDM, 20 steps, guidance 7.5; VectorSynth-COSA DDIM, 40 steps, guidance 7.5; EarthSynth UniPC, 20 steps, guidance 7.5; Changen2-OEM DiT DDIM, 50 steps. Changen2 S1 and S9 are the released datasets.} \\
\end{tabular}
\par\vspace{6pt}
\setlength{\tabcolsep}{3pt}
\begin{tabular}{@{}>{\raggedright\arraybackslash}p{1.4in}>{\raggedright\arraybackslash}p{0.6in}>{\raggedright\arraybackslash}p{0.6in}>{\raggedright\arraybackslash}p{0.6in}>{\raggedright\arraybackslash}p{0.6in}>{\raggedright\arraybackslash}p{0.6in}>{\raggedright\arraybackslash}p{0.6in}>{\raggedright\arraybackslash}p{0.6in}>{\raggedright\arraybackslash}p{0.6in}@{}}
\toprule
Class of the layout & Bare & Rangeland & Developed & Road & Tree & Water & Agriculture & Building \\
\midrule
Seg2Sat: FLAIR class & bare soil & herbaceous vegetation & impervious surface & impervious surface & deciduous & water & agricultural land & building \\
VectorSynth-COSA: OSM tag & landuse brownfield & landuse grass & amenity parking & highway residential & natural wood & natural water & landuse farmland & building \\
EarthSynth, Changen2-OEM & \multicolumn{8}{l}{the same eight classes (no mapping)} \\
\bottomrule
\end{tabular}

%% file: supp/s5_real_labels.tex
\twocolumn[{%
\centering
\includegraphics[width=\textwidth]{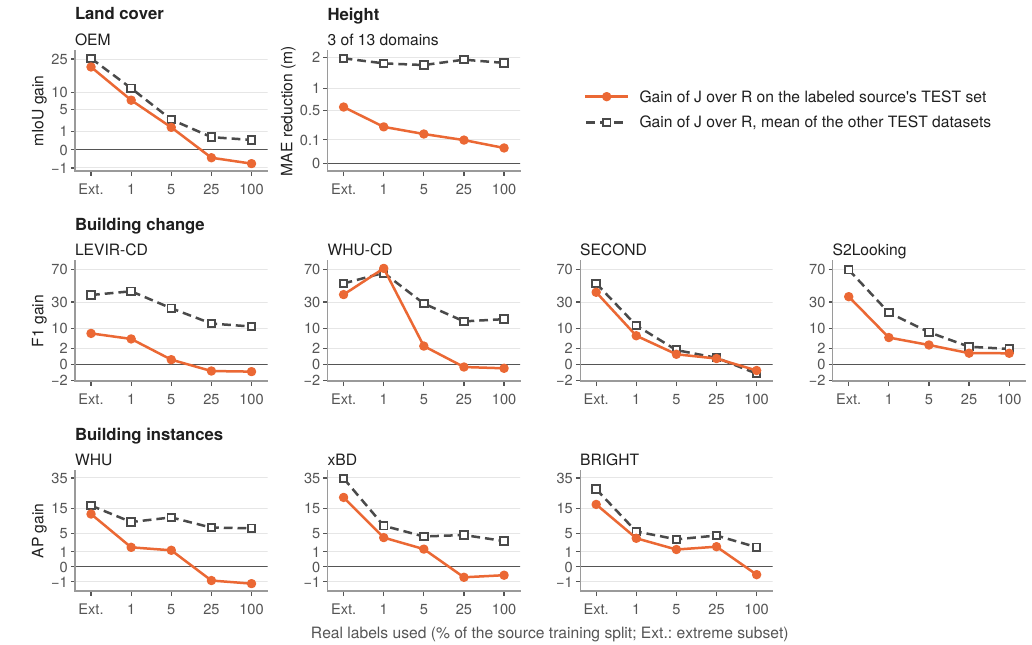}
\captionof{figure}{\textbf{Where synthetic supervision improves transfer (TEST).} Paired gains of joint training (J) over real-only training (R) on the labeled source's TEST set and averaged over the other TEST datasets of \cref{tab:s-p3-sources} (height: the three labeled domains and the other ten). For building change, the other datasets include S2Looking, which the task-level means of \cref{tab:p3} leave out. Positive values indicate improvement; height gains are MAE reductions (m). The vertical axes use a signed square-root scale, which widens the range near zero.}
\label{fig:s-scarce-sources}
\par\vspace{16pt}
}]
\section{Learning with scarce real labels}
\label{sec:s-real-labels}

We evaluate how \tlmix complements scarce real annotations across nine label sources and four downstream tasks.
The experiments compare real-only training, joint real--synthetic training, and fine-tuning from synthetic pretraining, with results evaluated across real TEST datasets.

\subsection{Training regimes and label budgets}
\label{sec:s-p3-config}

For each label source, we compare real-only training (R), joint training with \tlmix (J), and fine-tuning a synthetic-pretrained model on the same real annotations (F); R is \reg{REAL@p}, J and F are \reg{REAL@p+SYN}.
R and J train from the same initialization on the same schedule; in J, the real training list is repeated until it is at least as long as \tlmix's, so that at least half of the samples drawn are real.
F instead starts from the \reg{SYN} \tlmix model already trained on that schedule and fine-tunes it on the real annotations for another schedule with a fresh optimizer, so its total computation exceeds that of R and J.
Label budgets range from an extreme subset to 100\,\% of the source training split (\cref{tab:s-p3-sources}).
Height estimation uses annotations from Potsdam, Vaihingen, and Houston jointly; the other tasks use one labeled source at a time.
All runs follow the task-specific recipes of \cref{sec:s-protocols}, with the same number of updates at every budget (\cref{tab:recipes}), and are evaluated on the corresponding real TEST sets; every configuration is run with three seeds.
The \tlmix-only reference of \cref{fig:scarce}(a) (\reg{SYN}) is trained in separate runs with the same recipes.

\begin{table*}[!tp]
\caption{\textbf{Real-label sources and evaluation settings.} Extreme subsets are the smallest labeled training sets of each task, drawn anew for each seed, with their share of the training split (WHU-CD: of the 1\,024-px windows of its training mosaic); larger budgets are fractions of that split. For building change, the task-level means of \cref{fig:scarce} and \cref{tab:p3} average LEVIR-CD, WHU-CD, and SECOND as in \cref{tab:setup}, while S2Looking counts among the other TEST datasets of \cref{fig:s-scarce-sources}.}
\label{tab:s-p3-sources}
\centering
\footnotesize
\input{tables/supp_p3_sources}
\end{table*}

\subsection{Results across label budgets and datasets}
\label{sec:s-p3-budgets}
\label{sec:s-p3-sets}

\paragraph{Across label budgets.}
Joint training with \tlmix improves the task-level TEST mean over real-only training for all nine label sources through the 5\,\% budget (\cref{tab:p3}).
Positive mean gains remain for eight sources at 25\,\% and seven at 100\,\%, although their magnitudes decrease for most tasks.
Fine-tuning from synthetic pretraining generally offers smaller gains than joint training at the lowest label budgets, while their relative performance varies with larger training sets.
At both the 25\,\% and 100\,\% budgets, fine-tuning scores above both real-only and joint training for SECOND, S2Looking, and xBD.
The unusually large joint-training gain for WHU-CD at 1\,\% reflects a weak real-only baseline: the extreme subset is one window that contains change (\cref{tab:s-p3-sources}), whereas the uniformly sampled 1\,\% subset contains only 0.40\,\% changed pixels.

\begin{table}[!t]
\caption{\textbf{Performance across real-label budgets (TEST).} Real-only training (R), joint real--synthetic training (J), and fine-tuning from \tlmix (F) per label source; task-level means over three seeds. Label sources and extreme subsets as in \cref{tab:s-p3-sources}; shaded rows: J.}
\label{tab:p3}
\centering
\footnotesize
\setlength{\tabcolsep}{3.2pt}
\input{tables/p3}
\end{table}

\paragraph{Beyond the labeled source.}
As real-label budgets increase, the benefits of joint training often persist more strongly on evaluation datasets outside the labeled source (\cref{fig:s-scarce-sources}).
At 100\,\% labels, several sources show little improvement or a slight decline on their own TEST sets, despite positive task-level mean gains.
For height, J lowers MAE more on the ten domains without real labels than on the three labeled ones at every budget.
SECOND at 100\,\% is the exception to the pattern, with J below R both on SECOND and on the mean of the other datasets, while fine-tuning scores above R on the task-level mean (\cref{tab:p3}).
Synthetic supervision thus complements source-specific real annotations when labels are scarce and, at larger budgets, mainly by supporting transfer across datasets.

%% file: tables/supp_p3_sources.tex
\begin{tabular}{@{}llll@{}}
\toprule
Task & Real-label source & Extreme subset & Other TEST datasets \\
\midrule
Land cover & OEM & one tile per continent (0.2\,\%) & LoveDA \\
\midrule
Height & Potsdam, Vaihingen, Houston (3 of 13 domains) & one tile per domain (0.07\,\%) & other 10 domains \\
\midrule
Building change & LEVIR-CD & one pair with change (0.22\,\%) & WHU-CD, SECOND, S2Looking \\
 & WHU-CD & one window with change (0.32\,\%) & LEVIR-CD, SECOND, S2Looking \\
 & SECOND & one pair with change (0.04\,\%) & LEVIR-CD, WHU-CD, S2Looking \\
 & S2Looking & one pair with change (0.03\,\%) & LEVIR-CD, WHU-CD, SECOND \\
\midrule
Building instances & WHU & one tile with buildings (0.03\,\%) & xBD, BRIGHT \\
 & xBD & one tile with buildings (0.02\,\%) & WHU, BRIGHT \\
 & BRIGHT & one tile with buildings (0.03\,\%) & xBD, WHU \\
\bottomrule
\end{tabular}

%% file: tables/p3.tex
\begin{tabular}{@{}llccccc@{}}
\toprule
Real-label source & Regime & Extreme & 1\,\% & 5\,\% & 25\,\% & 100\,\% \\
\midrule
\multicolumn{7}{@{}l}{\cellcolor{tableband}\textit{Land cover: mIoU\,$\uparrow$, mean of OEM and LoveDA}} \\
OEM & R & 33.81 & 50.78 & 59.60 & 63.38 & 64.86 \\
\rowcolor{mainhl} & J & 56.90 & 60.26 & 61.75 & 63.52 & 64.72 \\
 & F & 43.99 & 56.02 & 59.72 & 63.18 & 64.45 \\
\midrule
\multicolumn{7}{@{}l}{\cellcolor{tableband}\textit{Height: MAE (m)\,$\downarrow$, mean of the 13 domains}} \\
3 of 13 domains & R & 4.388 & 4.150 & 4.022 & 4.182 & 4.126 \\
\rowcolor{mainhl} & J & 2.752 & 2.730 & 2.666 & 2.691 & 2.738 \\
 & F & 3.534 & 3.710 & 3.617 & 3.740 & 3.601 \\
\midrule
\multicolumn{7}{@{}l}{\cellcolor{tableband}\textit{Building change: F1\,$\uparrow$, mean of LEVIR-CD, WHU-CD, and SECOND}} \\
LEVIR-CD & R & 53.63 & 50.27 & 71.34 & 80.78 & 83.12 \\
\rowcolor{mainhl} & J & 77.49 & 78.07 & 81.90 & 84.35 & 84.70 \\
 & F & 72.09 & 64.32 & 76.39 & 81.06 & 83.07 \\
\addlinespace[2pt]
WHU-CD & R & 30.41 & 6.76 & 64.39 & 76.88 & 76.95 \\
\rowcolor{mainhl} & J & 76.89 & 78.16 & 82.58 & 84.25 & 84.60 \\
 & F & 56.99 & 29.28 & 70.35 & 79.02 & 80.24 \\
\addlinespace[2pt]
SECOND & R & 25.87 & 68.99 & 79.37 & 80.54 & 81.48 \\
\rowcolor{mainhl} & J & 74.47 & 78.29 & 80.26 & 79.98 & 79.97 \\
 & F & 47.93 & 76.29 & 81.10 & 81.80 & 82.11 \\
\addlinespace[2pt]
S2Looking & R & 4.27 & 55.53 & 70.38 & 76.38 & 76.39 \\
\rowcolor{mainhl} & J & 73.42 & 76.21 & 78.34 & 78.77 & 78.22 \\
 & F & 7.45 & 68.34 & 75.10 & 80.10 & 79.38 \\
\midrule
\multicolumn{7}{@{}l}{\cellcolor{tableband}\textit{Building instances: mask AP\,$\uparrow$, mean of xBD, WHU, and BRIGHT}} \\
WHU & R & 22.40 & 33.89 & 34.94 & 39.07 & 39.88 \\
\rowcolor{mainhl} & J & 37.56 & 40.36 & 42.53 & 43.31 & 43.85 \\
 & F & 33.82 & 37.56 & 39.74 & 42.38 & 43.13 \\
\addlinespace[2pt]
xBD & R & 5.01 & 33.69 & 38.33 & 39.65 & 40.84 \\
\rowcolor{mainhl} & J & 35.19 & 39.91 & 41.49 & 42.46 & 42.68 \\
 & F & 19.45 & 39.47 & 41.87 & 44.70 & 45.36 \\
\addlinespace[2pt]
BRIGHT & R & 12.11 & 33.67 & 37.62 & 38.50 & 41.30 \\
\rowcolor{mainhl} & J & 35.63 & 38.54 & 40.29 & 41.97 & 42.34 \\
 & F & 30.80 & 36.89 & 39.92 & 42.16 & 42.10 \\
\bottomrule
\end{tabular}

%% file: supp/s4_interventions.tex
\section{Dissecting world rules and scene properties}
\label{sec:s-atlas}
\label{sec:s-interventions}

This section gives the evidence behind \cref{sec:design}.
All models train on \tlm under \reg{SYN} with the controlled schedule of \cref{tab:recipes} (8k updates for land cover and height) and three seeds, and are evaluated on TEST; land-cover mIoU averages OEM and LoveDA, and height MAE the 13 domains.
Differences are paired by seed.
The rule ablations of \cref{sec:s-mabl} train without \photodr, so that changes in generated appearance remain observable during downstream learning; the interventions of \cref{sec:s-appearance,sec:s-composition,sec:s-observation} train with it, and each comparison is made within one batch of training runs: against the intact reference in \cref{sec:s-appearance,sec:s-observation}, and between modified training pools in \cref{sec:s-composition}.

\subsection{Ablating executable world rules}
\label{sec:s-mabl}

\paragraph{What is relaxed.}
The first three ablations of \cref{fig:properties}(a) relax selected priors (\mechP), conditional dependencies (\mechD), or constraints (\mechC) in full \tlm and keep the rest of the generator intact; the fourth relaxes all three together.
\Cref{tab:s-mabl} lists the rule groups that each ablation relaxes and what replaces them, beside the absolute scores of full \tlm and the changes shown in \cref{fig:properties}(a).
Each ablation relaxes all of its listed rule groups together and reports their joint effect; in particular, \mechC{} relaxes appearance and scene constraints together.
All worlds share scene-generation seeds and tile identifiers.

\paragraph{Scope.}
Each ablated world relaxes a selected subset of \tlm's rules.
Most of the library, including the rules that construct buildings, vegetation, roads, and terrain, stays in force in every world, as do the texture library and the renderer.
The relaxations are also bounded, so that the generated scenes do not become arbitrary: uniform sampling stays within the range that the generator allows, and the scene constraints give way to coarse global limits, not to none.
Even with \mechP, \mechD, and \mechC{} all relaxed, the generator is therefore not rule-free, and its world does not reproduce an earlier synthetic dataset such as SynRS3D~\cite{synrs3d}, an independent system built on handcrafted assets (\cref{sec:intro}) onto which these rule kinds would first have to be mapped for a matched ablation.
The ablations thus measure what relaxing selected rules changes within \tlm, not how much of the performance difference between the two in \cref{tab:main} is due to these rules.

\begin{table}[!t]
\caption{\textbf{World-rule ablations: rules relaxed and results (TEST).} The first row gives the absolute scores of full \tlm; the other rows report seed-paired changes relative to it for the four ablations of \cref{fig:properties}(a). Each ablation relaxes all of its listed rule groups together, and each change is that of the ablation as a whole; an arrow ($\rightarrow$) points to what replaces a rule group. Land-cover mIoU is the mean of OEM and LoveDA; height MAE is the mean over 13 domains. Values are means over three training seeds under the controlled synthetic-only protocol without \photodr.}
\label{tab:s-mabl}
\centering
\footnotesize
\setlength{\tabcolsep}{3pt}
\input{tables/supp_mabl}
\end{table}

\subsection{Appearance: texture strength}
\label{sec:s-appearance}

We fade within-region texture while preserving the semantic labels (\cref{fig:properties}(b)).
Texture is each pixel's deviation from the mean color of its region, the regions being building and tree-crown instances and connected class components; fading scales this deviation by a constant factor and the texture energy by the square of that factor.
Fading to 75\,\%, 50\,\%, 25\,\%, and 11\,\% of the texture energy reduces land-cover mIoU by 0.10, 0.51, 1.85, and 4.17 points; the levels use one recipe, one set of seeds, and one intact reference.

\begin{table*}[!t]
\caption{\textbf{Height estimation under different training-view configurations (TEST).} Changes in MAE (m) relative to the intact reference of each configuration's batch of runs, by reference building height; Ref.\ MAE is that of the reference in the nadir-only batch. The five bins cover building pixels (Pixels, in millions), and their values are unweighted means over the domains with at least 1\,000 building pixels in the bin (Domains); the last row covers all pixels of the 13 domains.}
\label{tab:s-observation}
\centering
\footnotesize
\setlength{\tabcolsep}{4pt}
\input{tables/supp_hcf_bins}
\end{table*}

\subsection{Composition: rare scene mixtures}
\label{sec:s-composition}

We study scene composition through the land-cover area fractions of \tlm training tiles (\cref{fig:properties}(c)).
A tile's composition is its eight class-area fractions, and its rarity the Aitchison (log-ratio) distance~\cite{aitchison1982} to its twentieth nearest neighbor among the training compositions, computed after 0.01 is added to every fraction and the fractions are renormalized, so a rare composition can consist of common classes.
With the training-pool size held fixed (removed tiles are replaced by copies of retained ones), removing the rarest 10\,\% of compositions costs 1.40~mIoU more than removing a random 10\,\% and 1.62~mIoU more than removing a random half of the pool.

The removal also changes class exposure, in particular the pixels of bare land (\ensuremath{-56}\,\%) and water (\ensuremath{-38}\,\%).
Restoring the class-pixel exposure, by choosing the replacement copies so that every class keeps its pixel share of the full pool, recovers 0.48~mIoU, only part of the loss, although this control does not isolate composition as the sole cause.

\subsection{Observation: oblique views and building height}
\label{sec:s-observation}

We re-render identical \tlm scenes under different camera configurations, regenerating the optical images and the aligned labels for each view: from nadir only, with the look angle of tilted views doubled, or with every view tilted.
Removing oblique training views increases the overall height MAE by only 0.051~m, but the increase reaches 2.781~m for buildings at least 40~m tall (\cref{fig:properties}(d), \cref{tab:s-observation}).
Without oblique views, the increase grows with building height (0.191~m at 10--20~m, 0.775~m at 20--40~m), and in the two tallest bins MAE rises in every domain: in all twelve domains at 20--40~m and in all ten at 40~m and above, where the increase is 2.293~m without Atlanta, the domain with most of the bin's pixels.
Doubling the look angle of tilted views or tilting every view lowers MAE in the three bins from 10~m upward.
The results illustrate how observation coverage can affect specific structures more strongly than the overall metric suggests.

%% file: tables/supp_mabl.tex
\begin{tabular}{@{}l@{\hspace{4pt}}>{\raggedright\arraybackslash}p{1.84in}rr@{}}
\toprule
\multicolumn{2}{@{}l}{Rules relaxed $\rightarrow$ replacement} & \makecell[r]{Land-cover\\mIoU $\uparrow$} & \makecell[r]{Height\\MAE (m) $\downarrow$} \\
\midrule
\multicolumn{2}{@{}l}{Full \tlm{} (absolute scores)} & 57.67 & 2.929 \\
\midrule
\mechP{} & \textbf{Selected priors} $\rightarrow$ uniform over their valid ranges or supported categories & \ensuremath{-1.23} & \ensuremath{+0.090} \\
 & \multicolumn{3}{@{}>{\raggedright\arraybackslash}p{3.07in}@{}}{\hangindent=0.8em\hangafter=1\relax --~Scene priors, e.g., the terrain regime} \\
 & \multicolumn{3}{@{}>{\raggedright\arraybackslash}p{3.07in}@{}}{\hangindent=0.8em\hangafter=1\relax --~Appearance priors, e.g., canopy color} \\
 & \multicolumn{3}{@{}>{\raggedright\arraybackslash}p{3.07in}@{}}{\hangindent=0.8em\hangafter=1\relax --~Shares of scene types and, in the tilting scene program, of building-layout families $\rightarrow$ equal shares} \\
 & \multicolumn{3}{@{}>{\raggedright\arraybackslash}p{3.07in}@{}}{\hangindent=0.8em\hangafter=1\relax --~Presence of selected scene elements $\rightarrow$ presence and absence equally likely} \\
\midrule
\mechD{} & \textbf{Selected dependencies} $\rightarrow$ independent draws with unchanged marginals & \ensuremath{-0.89} & \ensuremath{+0.215} \\
 & \multicolumn{3}{@{}>{\raggedright\arraybackslash}p{3.07in}@{}}{\hangindent=0.8em\hangafter=1\relax --~Scene dependencies, e.g., the height of a tree crown given its neighboring trees} \\
 & \multicolumn{3}{@{}>{\raggedright\arraybackslash}p{3.07in}@{}}{\hangindent=0.8em\hangafter=1\relax --~Appearance dependencies, e.g., canopy color given the tile's ground} \\
 & \multicolumn{3}{@{}>{\raggedright\arraybackslash}p{3.07in}@{}}{\hangindent=0.8em\hangafter=1\relax --~Roof materials drawn from a small set shared within a tile $\rightarrow$ drawn independently per building} \\
\midrule
\mechC{} & \textbf{Selected constraints} $\rightarrow$ switched off or coarse global limits & \ensuremath{-4.28} & \ensuremath{+0.218} \\
 & \multicolumn{3}{@{}>{\raggedright\arraybackslash}p{3.07in}@{}}{\hangindent=0.8em\hangafter=1\relax --~Appearance constraints (metric scale of textures and tree crowns; color bounds of the classes; street-appearance rules) $\rightarrow$ switched off} \\
 & \multicolumn{3}{@{}>{\raggedright\arraybackslash}p{3.07in}@{}}{\hangindent=0.8em\hangafter=1\relax --~Scene constraints (each module's limits on building height, footprint area, and aspect ratio; its building-placement checks; detailed tree-placement rules) $\rightarrow$ a global height range and coarse overlap checks for buildings and tree crowns} \\
\midrule
\multicolumn{2}{@{}l}{\mechP{}\,+\,\mechD{}\,+\,\mechC{}\hspace{4pt}\textbf{All three together}} & \ensuremath{-7.28} & \ensuremath{+0.450} \\
\bottomrule
\end{tabular}

%% file: tables/supp_hcf_bins.tex
\begin{tabular}{@{}lrrrrrr@{}}
\toprule
 & & & & \multicolumn{3}{c}{Change in MAE (m)\,$\downarrow$} \\
\cmidrule(l){5-7}
Reference height (m) & Ref.\ MAE (m)\,$\downarrow$ & Domains & Pixels (M) & Nadir only & Look angle doubled & Every view tilted \\
\midrule
0--2 & 4.45 & 13 & 66.5 & \ensuremath{-0.262} & \ensuremath{-0.016} & \ensuremath{+0.029} \\
2--10 & 2.82 & 13 & 544.8 & \ensuremath{-0.004} & \ensuremath{-0.111} & \ensuremath{+0.006} \\
10--20 & 5.37 & 13 & 117.3 & \ensuremath{+0.191} & \ensuremath{-0.127} & \ensuremath{-0.130} \\
20--40 & 9.59 & 12 & 39.3 & \ensuremath{+0.775} & \ensuremath{-0.131} & \ensuremath{-0.287} \\
$\geq$40 & 29.14 & 10 & 14.6 & \ensuremath{+2.781} & \ensuremath{-0.258} & \ensuremath{-0.712} \\
\midrule
13-domain mean, all pixels & 2.95 & 13 & -- & \ensuremath{+0.051} & \ensuremath{-0.031} & \ensuremath{-0.021} \\
\bottomrule
\end{tabular}

%% file: supp/s7_damage.tex
\section{Practical applications}
\label{sec:s-utility}

The synthetic multi-task supervision of \cref{sec:transfer} serves two applications beyond the benchmark tasks: recognizing destroyed buildings after disasters without damage labels (\cref{sec:s-damage}), and building and height predictions that can be set against released global building products (\cref{sec:s-prod}).
Both apply the \tlmix + \uda models of \cref{tab:main} as they are, without real labels of the application: the land-cover, height, and change models in the first, the instance and height models in the second.

\subsection{Cross-event damage recognition on known building footprints}
\label{sec:s-damage}

\begin{figure*}[!tp]
\centering
\includegraphics[width=\textwidth]{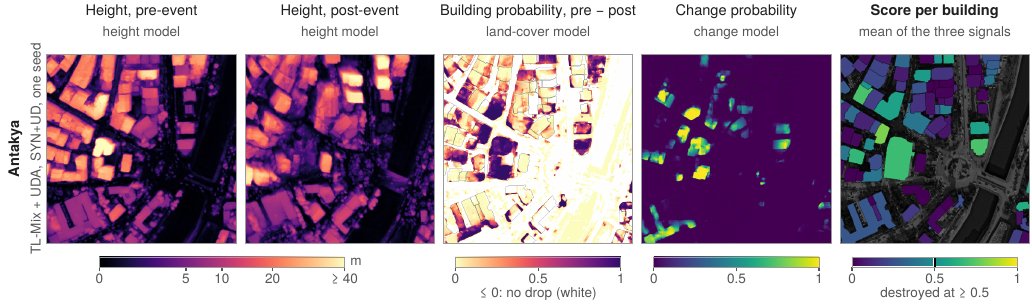}
\caption{\textbf{How the label-free damage score is formed} (Antakya, Turkey earthquake; one seed). From left: pre- and post-event height predicted by the \tlmix + \uda height model, the drop of the land-cover model's building probability, the change model's change probability, and the per-building score, the mean of the three signals; a building counts as destroyed at 0.5 and above. Gray background: pre-event image, BRIGHT~\cite{bright} (Maxar Open Data, CC BY-NC 4.0).}
\label{fig:s-damage-ingredients}
\end{figure*}

\begin{table*}[!tp]
\caption{\textbf{Destroyed-building recognition on known footprints (BRIGHT, four events).} Destroyed-class F1 (at the fixed threshold 0.5 for the label-free and fused scores; the supervised model: its class of highest mean probability within a building) and destroyed-versus-rest AUROC; means over 3 paired seeds, the mean over the events with its SD across seeds. The label-free rows use no damage labels (\tlmix + \uda, \reg{SYN+UD}); the xBD-supervised rows use real damage labels (\reg{XREAL}).}
\label{tab:s-dmg-destroyed}
\centering
\footnotesize
\setlength{\tabcolsep}{4pt}
\input{tables/supp_dmg_destroyed}
\end{table*}

\paragraph{Damage score.}
We combine changes in predicted building height, building probability, and change probability to identify destroyed buildings without damage-label training (\cref{fig:s-damage-ingredients}).
The \tlmix + \uda land-cover, height, and change models of \cref{tab:main} (\reg{SYN+UD}, 3 seeds; their unlabeled pools hold no BRIGHT image, \cref{tab:recipes}) are applied to the pre- and post-event images, and each reference building receives three scores, each clipped to the unit interval: the relative drop of its median height, $r_h=\operatorname{clip}((h_{\mathrm{pre}}-h_{\mathrm{post}})/\max(h_{\mathrm{pre}},1\,\mathrm{m}))$; the drop of its mean building probability, $r_p=\operatorname{clip}(p_{\mathrm{pre}}-p_{\mathrm{post}})$; and its mean change probability $r_c$.
The damage score is their equal-weight mean, $r=\operatorname{mean}(r_h,r_p,r_c)$, and a building counts as destroyed when $r\geq0.5$.

\paragraph{Evaluation.}
Evaluation uses the 156\,408 reference building footprints of four BRIGHT~\cite{bright} events, 5\,963 of them destroyed, so it covers damage recognition on known footprints, not end-to-end building detection and damage mapping.
Post-event optical images come from Maxar Open Data~\cite{maxaropendata} (Hawaii, Libya, and Turkey; CC BY-NC 4.0) and the Geospatial Information Authority of Japan (Noto); those of the Turkey earthquake are chosen per tile from the acquisition metadata alone (least cloud cover, then lowest off-nadir angle), and all are resampled to the pre-event grid without registration correction.
The supervised reference (\reg{XREAL}) is ChangeMamba's MambaBDA decoder~\cite{changemamba} with a DINOv3-L encoder~\cite{dinov3}, trained on the damage labels of 9\,168 xBD image pairs~\cite{xbd}; its score is the mean destroyed probability within a building, its decision the class of highest mean probability within the building, and its fusion with ours averages the two scores and therefore uses real damage labels.
Scores and the threshold were fixed before the results were read.

\paragraph{Results.}
The equal-weight combination reaches 49.9 destroyed-class F1 and 0.861 AUROC on average over the four events, above each single signal and above the combination without the change model (\cref{tab:s-dmg-destroyed}).
Without damage labels, it approaches the xBD-supervised reference (50.6 F1, 0.869 AUROC), and combining the two sources increases F1 to 54.9 and AUROC to 0.883.

\begin{figure*}[!tp]
\centering
\includegraphics[width=0.95\textwidth]{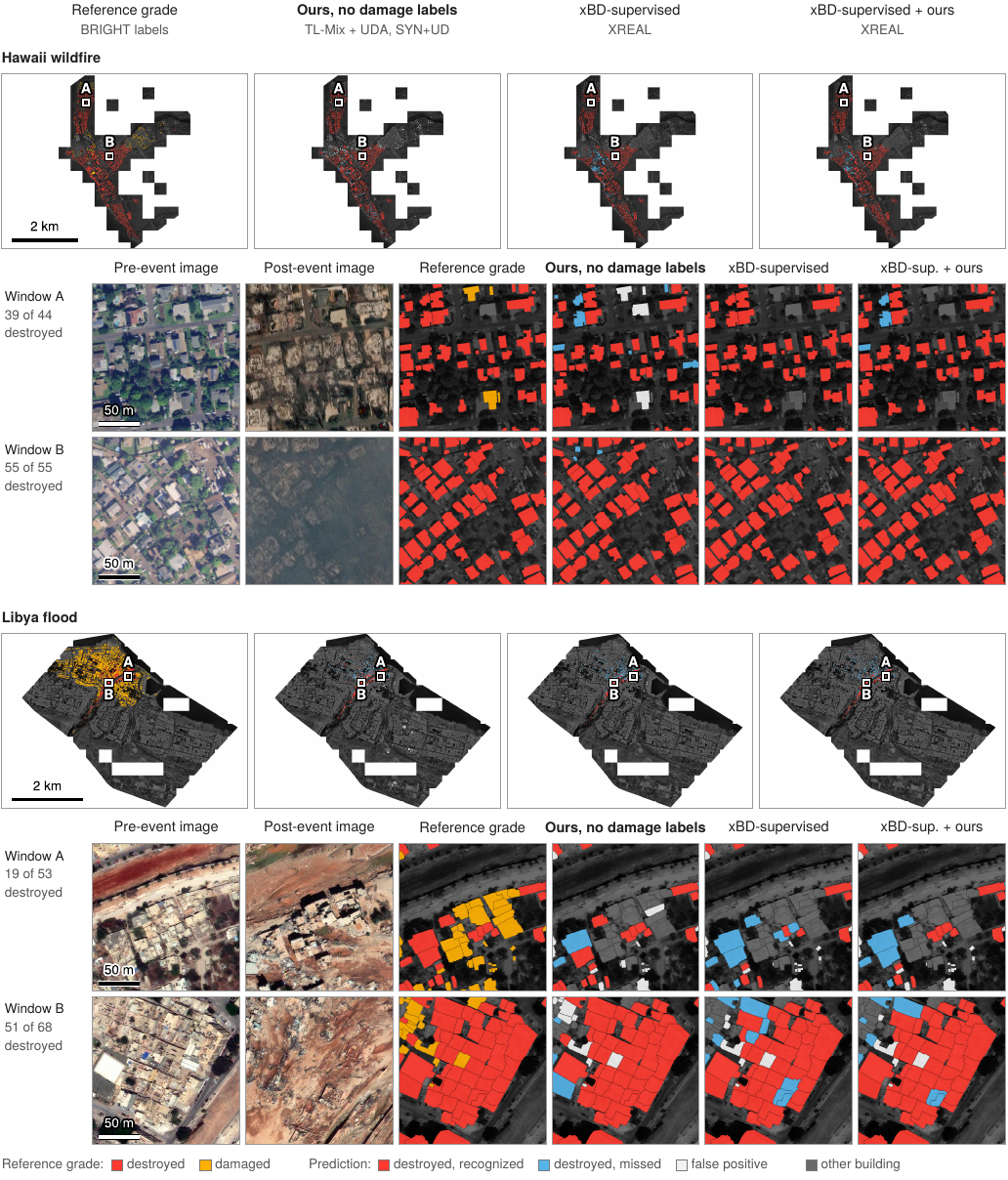}
\caption{\textbf{Destroyed-building recognition in the Hawaii wildfire and the Libya flood.} Per event, top: reference grades and the per-building decisions of ours (\tlmix + \uda, \reg{SYN+UD}, no damage labels), of the xBD-supervised model, and of their fusion (both \reg{XREAL}) on the reference footprints, over the gray pre-event image; one seed. Gray buildings are intact in the reference panels and, in the prediction panels, neither destroyed nor predicted as destroyed. Bottom: the windows A and B marked in the top panels, with the pre- and post-event images. Pre-event images: BRIGHT~\cite{bright} (Hawaii: NOAA Office for Coastal Management; Libya: Maxar Open Data, CC BY-NC 4.0). Post-event images: Maxar Open Data~\cite{maxaropendata} (CC BY-NC 4.0).}
\label{fig:s-damage-hawaii-libya}
\end{figure*}

\begin{figure*}[!tp]
\centering
\includegraphics[width=0.95\textwidth]{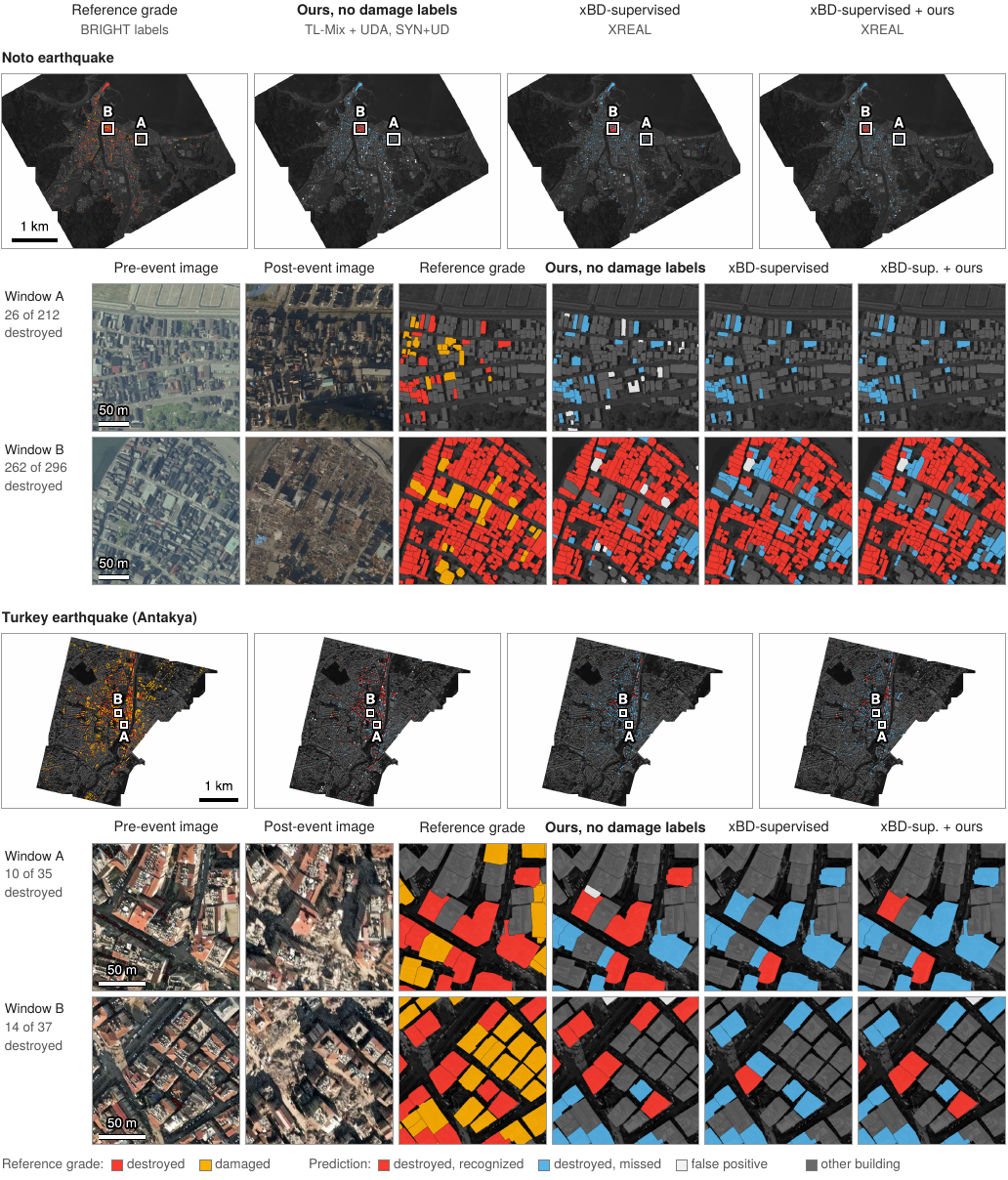}
\caption{\textbf{Destroyed-building recognition in the Noto earthquake and the Turkey earthquake (Antakya).} Layout, colors, and training conditions as in \cref{fig:s-damage-hawaii-libya}. Pre-event images: BRIGHT~\cite{bright} (Noto: Geospatial Information Authority of Japan; Turkey: Maxar Open Data, CC BY-NC 4.0). Post-event images: Geospatial Information Authority of Japan (Noto) and Maxar Open Data~\cite{maxaropendata} (Turkey; CC BY-NC 4.0).}
\label{fig:s-damage-noto-turkey}
\end{figure*}

Performance varies across events (\cref{tab:s-dmg-destroyed,fig:s-damage-hawaii-libya,fig:s-damage-noto-turkey}): in both metrics, the xBD-supervised reference is ahead on the Hawaii wildfire and the Noto earthquake, and the label-free score on the Turkey earthquake, whose post-event images were acquired 6--44$^\circ$ off nadir; on the Libya flood, the label-free score is ahead in F1 and the supervised reference in AUROC.
The events differ in disaster type and image acquisition, and the evaluation assumes known footprints and uses images that are not finely registered across dates.
Within this scope, the results show that synthetic-trained task models provide complementary signals for footprint-based damage recognition.

%% file: tables/supp_dmg_destroyed.tex
\begin{tabular}{@{}lrrrrr>{\hspace{12pt}}rrrrr@{}}
\toprule
 & \multicolumn{5}{c}{Destroyed F1 ($\times$100) $\uparrow$} & \multicolumn{5}{c}{Destroyed AUROC $\uparrow$} \\
\cmidrule(lr){2-6}\cmidrule(l{16pt}){7-11}
Score & Hawaii & Libya & Noto & Turkey & Mean & Hawaii & Libya & Noto & Turkey & Mean \\
\midrule
\multicolumn{11}{@{}l}{\cellcolor{tableband}\textit{No damage labels (\reg{SYN+UD})}} \\
Height drop $r_h$ & 81.8 & 51.9 & 33.4 & 9.4 & 44.1\,{\scriptsize$\pm$0.1} & 0.801 & 0.911 & 0.750 & 0.799 & 0.815\,{\scriptsize$\pm$0.006} \\
Building-probability drop $r_p$ & 80.4 & 54.4 & 35.6 & 16.0 & 46.6\,{\scriptsize$\pm$0.3} & 0.795 & 0.903 & 0.715 & 0.863 & 0.819\,{\scriptsize$\pm$0.006} \\
Change probability $r_c$ & 83.1 & 42.2 & 30.8 & 24.2 & 45.1\,{\scriptsize$\pm$0.7} & 0.918 & 0.871 & 0.685 & 0.848 & 0.830\,{\scriptsize$\pm$0.009} \\
Mean of $r_h$ and $r_p$ & 82.9 & 53.9 & 34.3 & 12.9 & 46.0\,{\scriptsize$\pm$0.2} & 0.828 & 0.918 & 0.756 & 0.849 & 0.838\,{\scriptsize$\pm$0.005} \\
\rowcolor{mainhl}Equal weights (ours) & 86.1 & 58.1 & 36.3 & 19.3 & 49.9\,{\scriptsize$\pm$0.4} & 0.900 & 0.922 & 0.759 & 0.864 & 0.861\,{\scriptsize$\pm$0.003} \\
\multicolumn{11}{@{}l}{\cellcolor{tableband}\textit{With xBD damage labels (\reg{XREAL})}} \\
xBD-supervised & 92.9 & 56.6 & 42.5 & 10.5 & 50.6\,{\scriptsize$\pm$1.2} & 0.975 & 0.951 & 0.764 & 0.787 & 0.869\,{\scriptsize$\pm$0.007} \\
xBD-supervised + ours & 93.4 & 61.1 & 44.0 & 21.1 & 54.9\,{\scriptsize$\pm$1.0} & 0.969 & 0.943 & 0.769 & 0.853 & 0.883\,{\scriptsize$\pm$0.003} \\
\bottomrule
\end{tabular}

%% file: supp/s8_products.tex
\subsection{Regional comparison with global building products}
\label{sec:s-prod}

\begin{table*}[!tp]
\caption{\textbf{Regional comparison with global building products.} (a) Object F1 of building footprints, pooled over the evaluated areas of each dataset (BRIGHT's Turkey cities count separately). $^\dagger$\,Partial coverage, read on the product's own areas, where \tlmix + \uda reaches 36.4 (xBD, Open Buildings v3), 53.1 (BRIGHT, Open Buildings v3), and 53.3 (BRIGHT, Microsoft); n/a, no coverage. (b) Height MAE on building areas (m) in the domains with tile coordinates, for GBA.LoD1 (GBA) and \tlmix + \uda (Ours): on the part of the building area inside a GBA.LoD1 footprint (Cov., its share), and on the whole building area, a pixel outside every footprint counting as zero height for GBA.LoD1. Bold, the best value of a column (a) and the lower value of a pair (b).}
\label{tab:s-prod}
\centering
\footnotesize
\begin{minipage}[t]{0.50\textwidth}
\vspace{0pt}
\centering
\textit{(a) Building footprints, object F1 $\uparrow$}\\[2pt]
\input{tables/supp_prod_fp}
\end{minipage}\hfill
\begin{minipage}[t]{0.48\textwidth}
\vspace{0pt}
\centering
\textit{(b) Building heights, MAE on building areas (m) $\downarrow$}\\[2pt]
\input{tables/supp_prod_height}
\end{minipage}
\end{table*}

\paragraph{Protocol.}
We compare the building-instance and height predictions of \tlmix + \uda (\reg{SYN+UD}; one seed, the first of the three) with Open Buildings v3~\cite{sirko2021continental}, Microsoft footprints~\cite{msfootprints}, Overture~\cite{overture}, and GBA.LoD1~\cite{gba} on real test imagery (\cref{tab:s-prod}).
The products are used as released and rasterized to each test tile's grid; they cover the test areas unevenly and come from other acquisitions than the test images, and xBD tile coordinates come from its building labels.
Building footprints are scored by object F1, with instances matched one-to-one at a mask IoU above one half and counts pooled over the areas with reliable coordinates, where Overture or Microsoft footprints lie within a median shift of 5~m of the reference masks; a product with partial coverage is scored on the areas it covers.
Building heights are scored on the 9 TEST domains with tile coordinates (the tiles of the other four domains carry no geographic coordinates) as MAE on building areas, the pixels of the building class in the land-cover map distributed with each tile of the evaluation sets.
MAE is given both on the part that GBA.LoD1's footprints cover and on the whole building area, where a pixel outside every footprint counts as zero height for GBA.LoD1; it is pooled over the pixels of a domain and averaged over the domains.

\begin{figure*}[!tp]
\centering
\includegraphics[width=\textwidth]{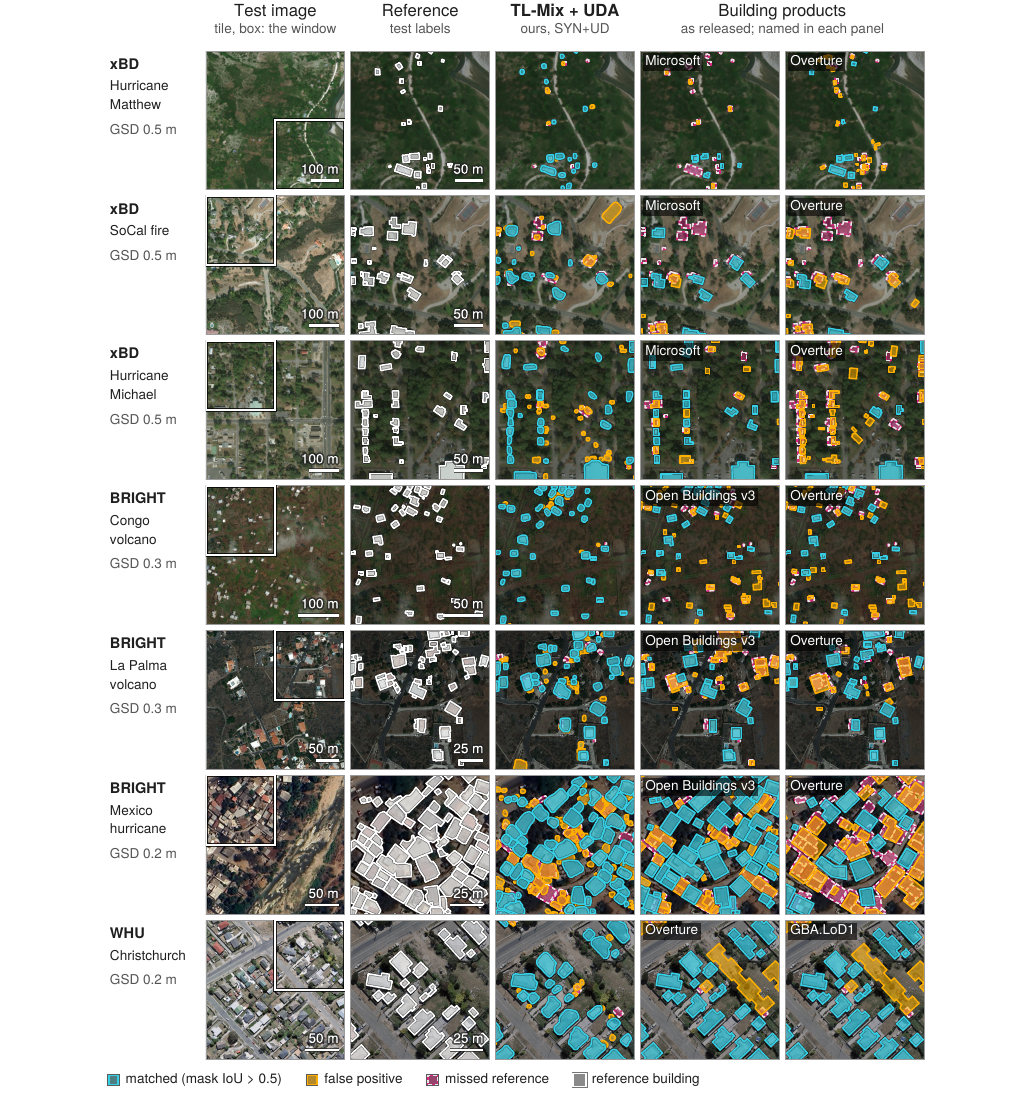}
\caption{\textbf{Building footprints of \tlmix + \uda and released building products on real test images (TEST).} Left: the test tile; the other panels show its boxed window with the reference outlines, our instances (\reg{SYN+UD}, one seed), and the two products with the highest pooled object F1 on the dataset (\cref{tab:s-prod}(a)), as released and rasterized to the tile. For xBD and BRIGHT, one test tile each from the areas with the largest, the median, and the smallest difference between our object F1 and the best product's (BRIGHT: among the areas within Open Buildings coverage); for WHU, one test tile. Images: xBD~\cite{xbd} (CC BY-NC-SA 4.0), BRIGHT~\cite{bright} (Congo: Maxar, CC BY-NC 4.0; La Palma: IGN Spain, CC BY 4.0; Mexico: Google Earth), WHU~\cite{whubuildings} (LINZ Data Service, \mbox{CC BY 4.0}). Footprints: \textcopyright\,OpenStreetMap contributors, Overture Maps Foundation (ODbL); Microsoft (CDLA Permissive 2.0); Open Buildings v3 (CC BY 4.0 / ODbL); GBA (ODbL / CC BY-NC 4.0).}
\label{fig:s-prod-instances}
\end{figure*}

\enlargethispage{-140pt}
\paragraph{Building footprints.}
The relative performance varies across regions.
On the disaster areas of xBD and BRIGHT, \tlmix + \uda reaches an object F1 of 48.7 and 52.4, above the products available there (at most 40.9 and 34.0; \cref{tab:s-prod}(a)), while in WHU's Christchurch the mature products perform better (86.3--88.5 against 61.4), our errors being mostly false positives (\cref{fig:s-prod-instances}).
Overture and GBA.LoD1 build on OpenStreetMap, where 65\,\% of the buildings in the WHU area cite Land Information New Zealand as their source, so they may share lineage with WHU's reference outlines.
Differences in coverage, acquisition dates, and georegistration affect these comparisons.

\begin{figure*}[!tp]
\centering
\includegraphics[width=\textwidth]{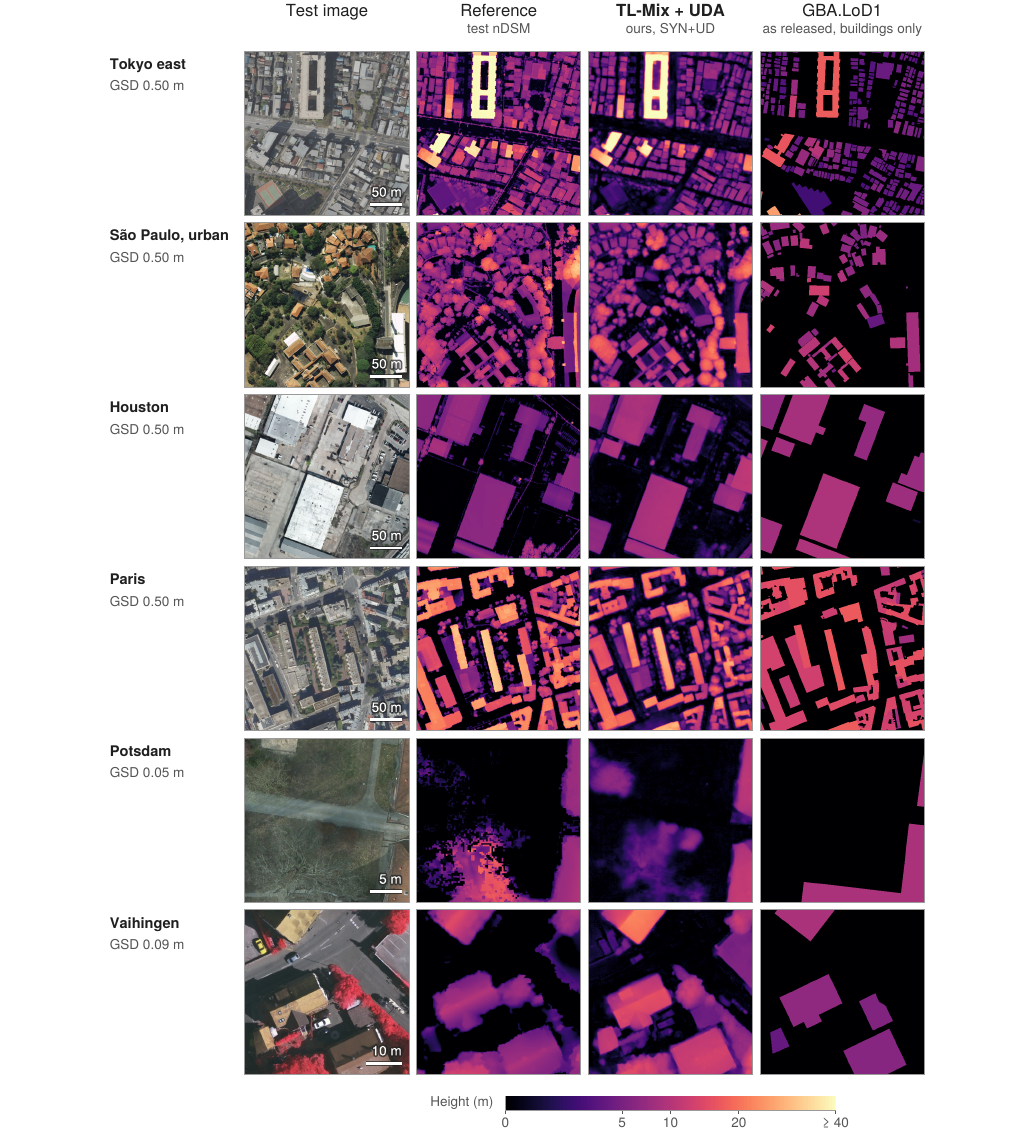}
\caption{\textbf{Building heights of \tlmix + \uda and GBA.LoD1 on real test images (TEST).} Test image, reference nDSM, our prediction (\reg{SYN+UD}, one seed), and GBA.LoD1~\cite{gba} as released, rasterized to the tile's grid; GBA.LoD1 holds one height per building footprint and nothing elsewhere. Six of the nine domains of \cref{tab:s-prod}(b), ordered by the difference in building-area MAE between ours and GBA.LoD1, from ours lower (Tokyo east) to GBA.LoD1 lower (Vaihingen); GBA.LoD1 is lower in the last two. One test tile per domain; square-root color scale. Images and reference nDSMs: Tokyo Digital Twin Project~\cite{tokyodt}, GeoSampa~\cite{geosampa}, DFC18~\cite{dfc18}, IGN BD~ORTHO\textsuperscript{\textregistered} and LiDAR~HD~\cite{ignbdortho,ignlidarhd} and ISPRS~\cite{isprs} (Vaihingen in color infrared), licensed as in \cref{fig:s-qual-h}. Footprints and heights: \textcopyright\,OpenStreetMap contributors (ODbL); GBA (ODbL / CC BY-NC 4.0).}
\label{fig:s-prod-heights}
\end{figure*}

\paragraph{Building heights.}
GBA.LoD1 provides one height per building footprint and nothing elsewhere, whereas \tlmix + \uda predicts a complete nDSM, so the two are compared on building areas only.
GBA.LoD1's footprints cover 74\,\% of the building area; on this covered part, inside the product's own footprints, where both give a height, MAE is 3.85~m for \tlmix + \uda against 4.62~m for GBA.LoD1, lower in 7 of 9 domains, with Potsdam and Vaihingen as the exceptions (\cref{tab:s-prod}(b), \cref{fig:s-prod-heights}).
Over the whole building area, with the uncovered part counted as zero height for GBA.LoD1, MAE is 3.90~m against 5.19~m, with the same two exceptions.
The building areas are not a manual annotation: against the building footprints of Tokyo's 3D city model from Project PLATEAU~\cite{plateau}, on the two Tokyo domains, their precision is 76\,\% and their recall 91\,\%.
These results illustrate that synthetic-trained predictions can complement existing building products in regions and attributes that the products cover unevenly; they do not establish a product ranking.

%% file: tables/supp_prod_fp.tex
\setlength{\tabcolsep}{7pt}
\begin{tabular}{@{}lccc@{}}
\toprule
 & xBD & BRIGHT & WHU \\
Evaluated areas & 8 & 14 & 1 \\
\midrule
Open Buildings v3 & 7.2$^\dagger$ & 34.0$^\dagger$ & \NA \\
Microsoft footprints & 40.9 & 28.6$^\dagger$ & \NA \\
Overture & 32.7 & 32.1 & \textbf{88.5} \\
GBA.LoD1 & 32.5 & 31.3 & 86.3 \\
\midrule
\rowcolor{mainhl}\tlmix + \uda & \textbf{48.7} & \textbf{52.4} & 61.4 \\
\bottomrule
\end{tabular}

%% file: tables/supp_prod_height.tex
\setlength{\tabcolsep}{3.5pt}
\begin{tabular}{@{}lccccc@{}}
\toprule
 & \multicolumn{2}{c}{Covered part} & \multicolumn{2}{c}{Whole building area} & \\
\cmidrule(lr){2-3}\cmidrule(lr){4-5}
Domain & GBA & Ours & GBA & Ours & Cov.\ (\%) \\
\midrule
Houston & 3.14 & \textbf{2.05} & 3.35 & \textbf{2.20} & 84 \\
Paris & 6.92 & \textbf{6.82} & 6.87 & \textbf{6.77} & 82 \\
Potsdam & \textbf{4.16} & 5.11 & \textbf{4.13} & 4.87 & 79 \\
S\~{a}o Paulo, forest & 2.87 & \textbf{1.83} & 3.09 & \textbf{1.88} & 52 \\
S\~{a}o Paulo, urban & 4.35 & \textbf{3.02} & 7.34 & \textbf{3.71} & 66 \\
Saint-Omer & 2.52 & \textbf{2.04} & 2.65 & \textbf{2.05} & 75 \\
Tokyo east & 8.71 & \textbf{4.77} & 9.44 & \textbf{4.80} & 73 \\
Tokyo west & 6.49 & \textbf{3.26} & 6.99 & \textbf{3.34} & 73 \\
Vaihingen & \textbf{2.44} & 5.77 & \textbf{2.81} & 5.50 & 79 \\
\midrule
Mean of 9 & 4.62 & \textbf{3.85} & 5.19 & \textbf{3.90} & 74 \\
\bottomrule
\end{tabular}

%% file: supp/s_ondemand.tex
\section{On-demand category extension}
\label{sec:s-ondemand}
\setlength{\textfloatsep}{12pt plus 2pt minus 2pt}\setlength{\dbltextfloatsep}{12pt plus 2pt minus 2pt}

\subsection{Protocol}
\label{sec:s-ondemand-protocol}

\paragraph{Request.}
The procedure starts from a request that names the target categories, their scope, and the target imagery.
Here the request names the 15 categories of iSAID~\cite{isaid} with the benchmark's definition of each, and color and panchromatic optical images at about 0.1--1~m GSD, as the public description of the benchmark's images states (color images from sources such as Google Earth, panchromatic images from the GF-2 and JL-1 satellites).

\paragraph{Benchmark and metric.}
iSAID~\cite{isaid} annotates 15 object categories with instance masks; we convert them to semantic masks and follow the benchmark's semantic protocol: windows of 896 pixels with an overlap of 384 pixels, each window predicted on its own.
All results are on the official validation split (458 images, 11\,644 windows); the labels of the test split are not public.
The foreground IoU of a category is $\mathrm{TP}/(\mathrm{TP}+\mathrm{FP}+\mathrm{FN})$ over all windows of the split, including those of images without the category; means are over the 15 categories.
We also report the benchmark's 16-class mIoU, which includes the background class.
The specialists and the unified model are trained on generated images and labels only; \cref{tab:s-datasets} lists the benchmark's splits.

\paragraph{Specialists.}
Each category has one binary segmenter: DINOv3 ViT-L/16 with a DPT head, trained with the optimizer and \photodr of the land-cover models (\cref{sec:s-training}) for 8k updates.
Task-specific sampling and augmentation: positive tiles come from the category's own generated set and negative tiles, without the category, from the other generated sets; tiles that contain the category are drawn preferentially; crops are zoomed log-uniformly; and no-data image borders and off-nadir views are simulated.
Training also uses gradient-norm clipping and an exponential moving average of the weights.
We report three seeds and the ensemble of their foreground probabilities.
For the 16-class mIoU, the three-seed ensembles of the specialists are fused per pixel by the highest foreground probability, with background where no specialist predicts its category.

\paragraph{Unified model.}
One 16-class model with the same network is trained for 24k updates on the union of the generated sets, with three seeds.
Every set labels the objects of all 15 categories, so an object of another category is never background.

\begin{table}[!t]
\caption{\textbf{Generated data and cost per category.} Objects and tiles: instances of the category and tiles that contain it among the 7\,500 generated tiles of its set. Generation: wall-clock minutes to render and index the set on up to 192 CPU cores. Training: minutes on one GPU for one specialist, one seed. The medians are the values of \cref{sec:ondemand}; the development of a card and its asset module (about 90~min per category) is not part of the table. Ship and harbor share one generated set and its generation time.}
\label{tab:s-ondemand-cards}
\centering
\footnotesize
\input{tables/supp_ondemand_cards}
\end{table}

\begin{table*}[!t]
\caption{\textbf{On-demand category extension: all results on the iSAID validation split (foreground IoU\,$\uparrow$, \%).} Open-vocabulary methods: the better of the two readings per category (\cref{sec:s-ondemand-protocol}); SegEarth-OV3$^\ast$ writes out one class name of the released class file. \tlm specialists, the \tlm unified model, and real labels (the specialists' network trained on the real training labels): mean $\pm$ SD of three seeds, and their ensemble (Ens.); the means are the values of \cref{tab:ondemand}. Bold: the best result without real labels in a row, among the open-vocabulary methods and the means of our two models; values level at the printed precision are all bold. Underlined: the best open-vocabulary method of the category (Best OV in \cref{tab:ondemand}), the two variants of SegEarth-OV3 counting as one method.}
\label{tab:s-ondemand-full}
\centering
\footnotesize
\input{tables/supp_ondemand_full}
\end{table*}

\paragraph{Open-vocabulary baselines.}
SegEarth-OV3~\cite{segearthov3}, ProC-SAM3~\cite{procsam3}, and ROSE-SAM3~\cite{rosesam3} (\reg{ZS}) run with their released code, class names, prompts, and thresholds.
SegEarth-OV3$^\ast$ differs from the release only in writing out the class name of the storage tank, which raises its IoU on that category; one other value changes in the first decimal (\cref{tab:s-ondemand-full}).
ROSE-SAM3 builds its prompt bank from unlabeled images of the target benchmark, as released.
From one forward pass we read each method twice---its 16-class map, and a single-category request in which only the category and the background compete---and keep the better reading per category.
The best open-vocabulary baseline of \cref{sec:ondemand} (Best OV in \cref{tab:ondemand}) is the highest of these over the methods for each category, the two variants of SegEarth-OV3 counting as one method.

\paragraph{Real-label reference.}
The same binary network is trained with the same schedule on the benchmark's training split, with three seeds (\reg{REAL@100}).

\subsection{Category cards}
\label{sec:s-ondemand-cards}

A card has five blocks: a definition and labeling rule; the observation range of the request (image types and GSD range); reuse; placement items; and switches.
The base world already draws some objects of the requested kinds at low rates, as unlabeled parts of its eight land-cover classes (for example vehicles, vessels, and simple sports grounds).
With a card loaded, the generator records the category and an instance identifier of these objects, and the switches redraw them with the card's geometry and appearance, so that no object of the category stays unlabeled; the reuse block raises how often the scene types that contain them are drawn, and the placement items add the assets, contexts, and constraints that the base world does not have.
A placement item names an asset and gives its geometry (types, size ranges, layouts), appearance (materials and colors relative to the tile's illumination), contexts with weights, relations to nearby objects, constraints (host surfaces, clearances, slope), and frequency.
Unlabeled look-alikes are placement items as well: round structures for storage tanks, containers for large vehicles, or floating platforms for ships.
\Cref{fig:ondemand}(a) shows part of one card, and \cref{tab:s-ondemand-cards} lists the generated objects and the cost of every category.

\begin{figure}[!t]
\centering
\footnotesize
\input{tables/supp_listing_knowledge}
\caption{\textbf{Knowledge text of one category (basketball court).} Verbatim: researched and written by an LLM agent under the four-section schema that the experts defined, and reviewed by them; the fourth section states the label extent that the request gives. The category card and its asset module implement this text.}
\label{fig:s-ondemand-knowledge}
\vspace{18pt}
\end{figure}

\begin{figure*}[!t]
\centering
\includegraphics[width=\textwidth]{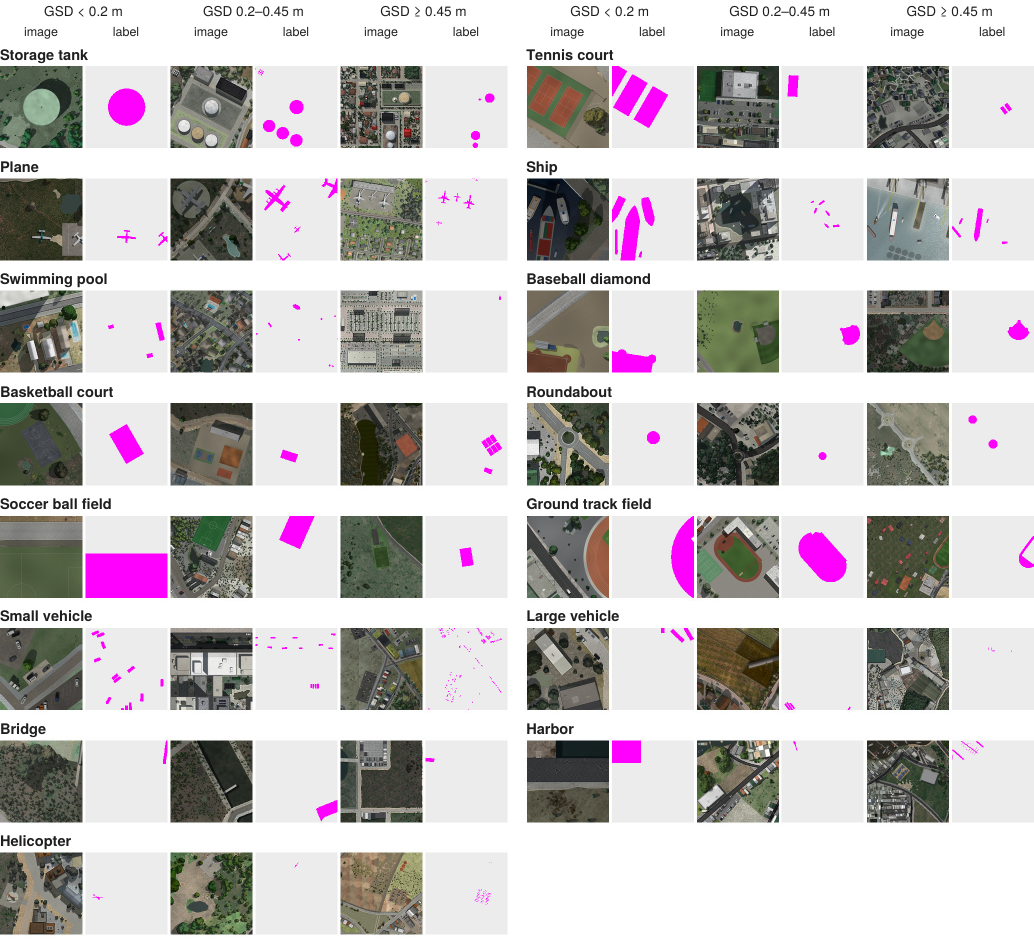}
\caption{\textbf{Generated images and labels of every category.} For each category, one tile of its generated set from each of three GSD ranges, drawn at random with a fixed seed among the tiles that contain the category; labels show the category only.}
\label{fig:s-ondemand-generated}
\end{figure*}

Each card and its asset module implement a category knowledge text that the LLM agents researched and wrote under the experts' schema (\cref{fig:s-ondemand-knowledge}); the agents used Claude Opus~5.5.
Cards and asset modules are loaded through a registration and placement layer that all categories share; with no card loaded, the generator's output is byte-identical to that of the version used in \cref{sec:supervision} on a fixed set of tiles.
\Cref{fig:s-ondemand-generated} shows generated images and labels of every category.

\subsection{Results and scope}
\label{sec:s-ondemand-results}

\Cref{tab:s-ondemand-full} gives every open-vocabulary method, the specialists, the unified model, and the real-label reference.
The specialists exceed ProC-SAM3 in 13 categories, ROSE-SAM3 in 12, and SegEarth-OV3$^\ast$ in 11.
The 16-class mIoU of the three-seed ensembles is 52.4 for the fused specialists and 48.6 for the unified model, against 29.5 (ProC-SAM3), 35.0 (ROSE-SAM3), and 31.2 (SegEarth-OV3$^\ast$).
One unified model processes a window in one forward pass (0.054~s) for all classes; the 15 specialists need one each (0.84~s together).
Its training takes 65~minutes on one GPU, against 5.3~hours for the 15 specialists, one seed each.

\paragraph{Color and panchromatic images.}
75 of the 458 validation images are panchromatic.
On the 10 categories that occur on both image types, the three-seed ensembles of the specialists reach 53.2 on color and 29.8 on panchromatic images, and each category's best open-vocabulary method 32.8 and 15.3, scored separately on the color and the panchromatic images.

\begin{figure*}[p]
\centering
\odxhead
\odxlabel{Storage tank}{ProC-SAM3}
\includegraphics[width=\odxw]{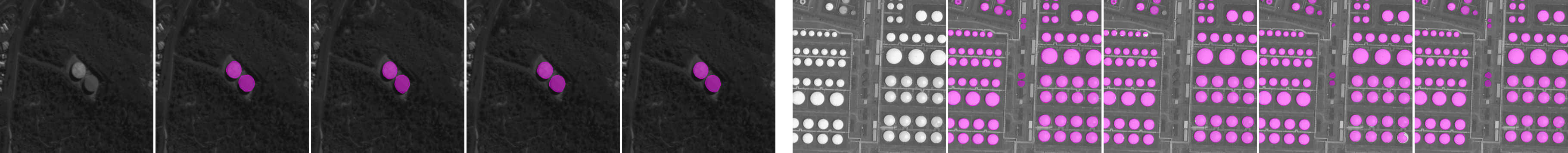}\\[4pt]
\odxlabel{Tennis court}{ROSE-SAM3}
\includegraphics[width=\odxw]{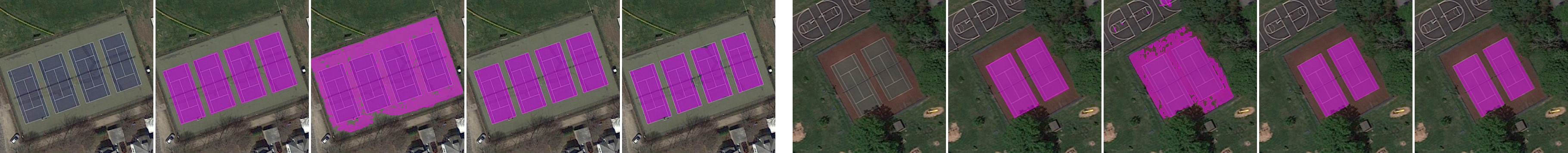}\\[4pt]
\odxlabel{Plane}{ROSE-SAM3}
\includegraphics[width=\odxw]{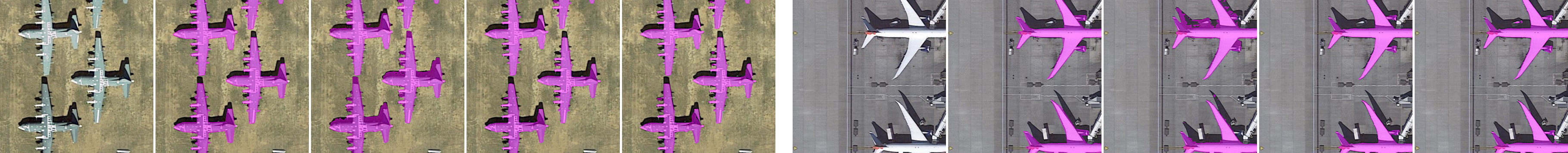}\\[4pt]
\odxlabel{Ship}{ProC-SAM3}
\includegraphics[width=\odxw]{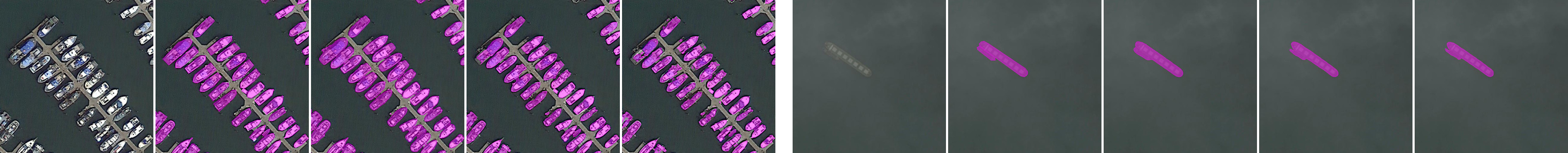}\\[4pt]
\odxlabel{Swimming pool}{SegEarth-OV3}
\includegraphics[width=\odxw]{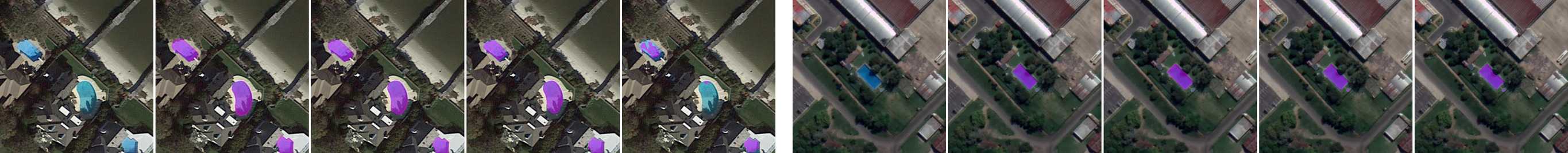}\\[4pt]
\odxlabel{Baseball diamond}{ProC-SAM3}
\includegraphics[width=\odxw]{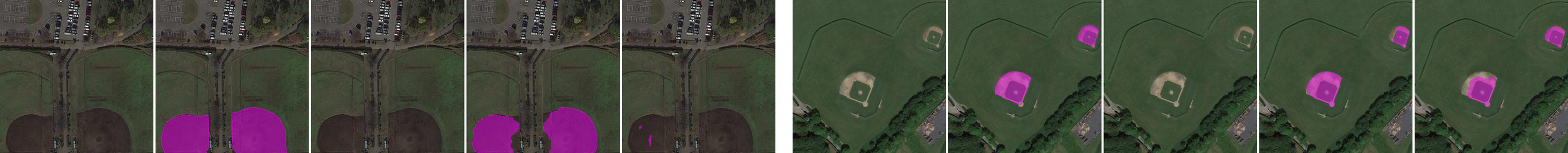}\\[4pt]
\odxlabel{Basketball court}{ROSE-SAM3}
\includegraphics[width=\odxw]{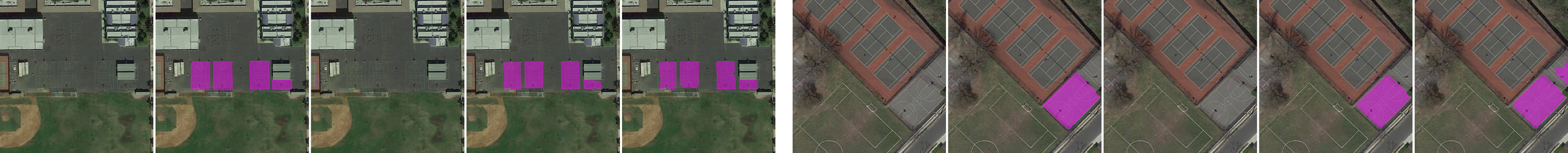}\\[4pt]
\odxlabel{Roundabout}{ROSE-SAM3}
\includegraphics[width=\odxw]{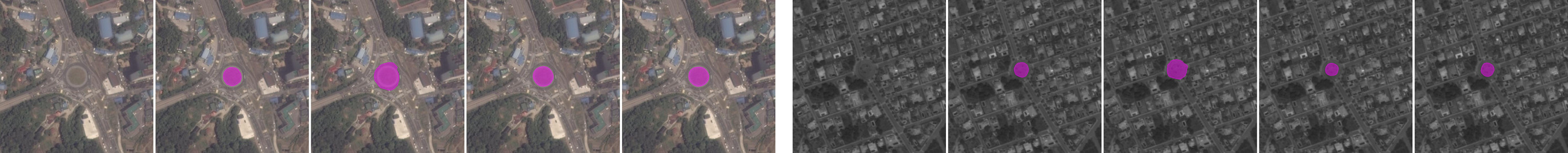}
\caption{\textbf{Real validation windows of every category.} Two windows per category, each with the image, the ground truth, the best open-vocabulary method of the category (Best OV), the \tlm specialist, and the \tlm unified model; the predictions of the specialist and the unified model are three-seed ensembles. Small objects are best viewed magnified. Images: iSAID validation split~\cite{isaid} (Google Earth and the JL-1 and GF-2 satellites; for academic use only).}
\label{fig:s-ondemand-val}
\end{figure*}

\paragraph{Qualitative results.}
\label{sec:s-ondemand-figures}
\Cref{fig:s-ondemand-val,fig:s-ondemand-val2} show two validation windows of every category, with the ground truth, the best open-vocabulary method of the category, the specialist, and the unified model.

\paragraph{Scope.}
This study evaluates one benchmark and one task, semantic segmentation.
Every generated tile also records the object category and an instance identifier per pixel, and a further category is a further card with its asset module; the same mechanism therefore applies to other categories and to instance-level tasks, which are not evaluated here.

\twocolumn[{%
\centering
\odxhead
\odxlabel{Soccer ball field}{ROSE-SAM3}
\includegraphics[width=\odxw]{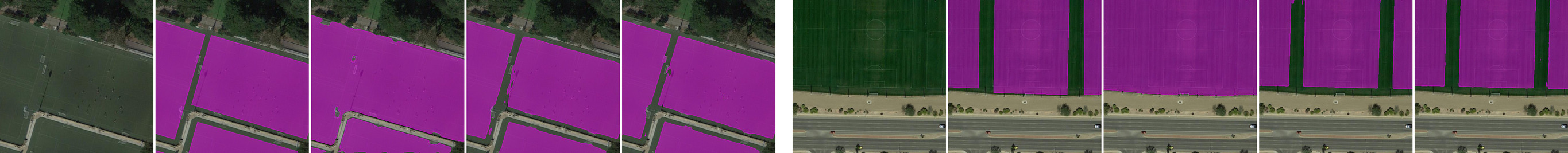}\\[4pt]
\odxlabel{Ground track field}{ProC-SAM3}
\includegraphics[width=\odxw]{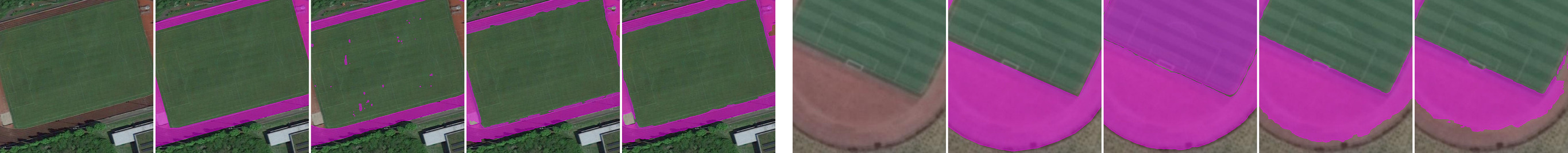}\\[4pt]
\odxlabel{Small vehicle}{ProC-SAM3}
\includegraphics[width=\odxw]{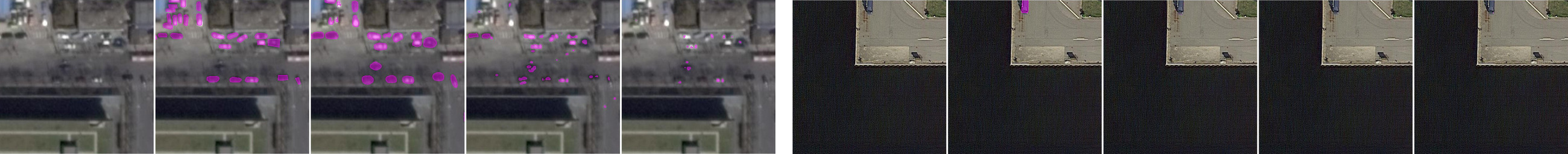}\\[4pt]
\odxlabel{Large vehicle}{SegEarth-OV3$^\ast$}
\includegraphics[width=\odxw]{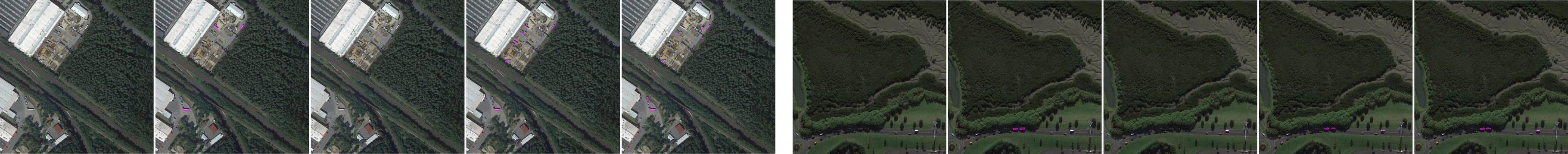}\\[4pt]
\odxlabel{Bridge}{SegEarth-OV3}
\includegraphics[width=\odxw]{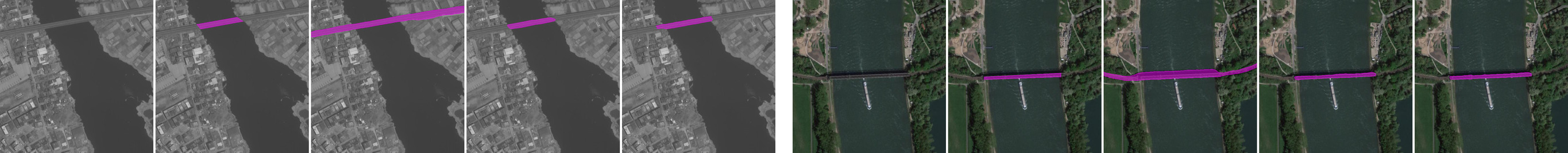}\\[4pt]
\odxlabel{Harbor}{ROSE-SAM3}
\includegraphics[width=\odxw]{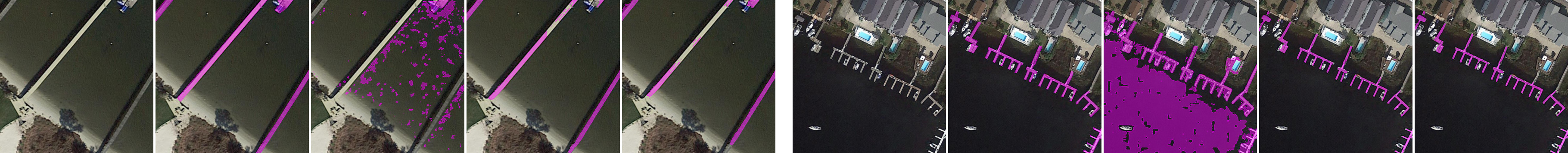}\\[4pt]
\odxlabel{Helicopter}{ProC-SAM3}
\includegraphics[width=\odxw]{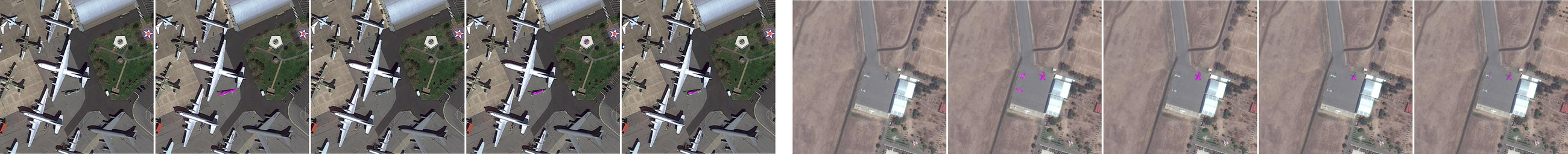}
\captionof{figure}{\textbf{Real validation windows of every category, continued from \cref{fig:s-ondemand-val}.} SegEarth-OV3$^\ast$ as in \cref{tab:s-ondemand-full}.}
\label{fig:s-ondemand-val2}
\par\vspace{16pt}
}]
\setlength{\textfloatsep}{20pt plus 2pt minus 4pt}\setlength{\dbltextfloatsep}{20pt plus 2pt minus 4pt}

%% file: tables/supp_ondemand_cards.tex
\setlength{\tabcolsep}{4pt}
\begin{tabular}{@{}l r r r r@{}}
\toprule
Category & Objects & Tiles & \makecell[r]{Generation\\(min)} & \makecell[r]{Training\\(min)} \\
\midrule
Storage tank & 32\,424 & 4\,086 & 27 & 21 \\
Tennis court & 18\,115 & 5\,726 & 28 & 22 \\
Plane & 12\,601 & 3\,322 & 34 & 22 \\
Ship & 71\,568 & 3\,465 & 34 & 21 \\
Swimming pool & 64\,581 & 5\,736 & 27 & 21 \\
Baseball diamond & 4\,690 & 3\,912 & 44 & 21 \\
Basketball court & 17\,837 & 6\,112 & 28 & 21 \\
Roundabout & 7\,976 & 4\,669 & 37 & 22 \\
Soccer ball field & 5\,952 & 4\,382 & 28 & 21 \\
Ground track field & 6\,801 & 3\,157 & 28 & 21 \\
Small vehicle & 660\,654 & 7\,296 & 40 & 21 \\
Large vehicle & 108\,560 & 6\,773 & 28 & 22 \\
Bridge & 11\,376 & 4\,556 & 28 & 21 \\
Harbor & 13\,704 & 3\,028 & 34 & 22 \\
Helicopter & 14\,581 & 3\,302 & 27 & 22 \\
\midrule
Median & -- & -- & 28 & 21 \\
\bottomrule
\end{tabular}

%% file: tables/supp_ondemand_full.tex
\setlength{\tabcolsep}{3pt}
\begin{tabular}{@{}l cccc c c c c c c@{}}
\toprule
 & \multicolumn{4}{c}{Open-vocabulary} & \multicolumn{2}{c}{\tlm specialists (ours)} & \multicolumn{2}{c}{\tlm unified (ours)} & \multicolumn{2}{c}{Real labels} \\
\cmidrule(lr){2-5}\cmidrule(lr){6-7}\cmidrule(lr){8-9}\cmidrule(l){10-11}
Category & ProC-SAM3 & ROSE-SAM3 & SegEarth-OV3 & SegEarth-OV3$^\ast$ & Mean & Ens. & Mean & Ens. & Mean & Ens. \\
\midrule
Storage tank & \underline{62.3} & 61.4 & 2.7 & 61.2 & 69.5\,{\scriptsize$\pm$0.8} & 71.2 & \textbf{69.8}\,{\scriptsize$\pm$0.4} & 70.5 & 67.7\,{\scriptsize$\pm$5.3} & 70.9 \\
Tennis court & 39.8 & \underline{44.6} & 27.5 & 27.5 & \textbf{81.1}\,{\scriptsize$\pm$1.4} & 82.5 & 72.7\,{\scriptsize$\pm$10.1} & 75.3 & 89.5\,{\scriptsize$\pm$0.2} & 90.0 \\
Plane & 66.7 & \underline{72.8} & 72.6 & 72.6 & \textbf{82.6}\,{\scriptsize$\pm$0.1} & 83.0 & 81.1\,{\scriptsize$\pm$0.2} & 81.6 & 83.6\,{\scriptsize$\pm$4.8} & 86.0 \\
Ship & \underline{49.9} & 34.3 & 34.6 & 34.6 & \textbf{60.2}\,{\scriptsize$\pm$1.2} & 60.7 & 54.0\,{\scriptsize$\pm$1.1} & 54.5 & 77.9\,{\scriptsize$\pm$3.1} & 78.9 \\
Swimming pool & 11.8 & 26.8 & \textbf{\underline{27.2}} & \textbf{\underline{27.2}} & \textbf{27.2}\,{\scriptsize$\pm$2.9} & 28.3 & 26.0\,{\scriptsize$\pm$2.3} & 26.4 & 31.8\,{\scriptsize$\pm$5.7} & 41.9 \\
Baseball diamond & \underline{6.7} & 2.9 & 4.8 & 4.8 & \textbf{41.8}\,{\scriptsize$\pm$3.0} & 44.6 & 27.2\,{\scriptsize$\pm$0.6} & 27.9 & 63.2\,{\scriptsize$\pm$6.4} & 72.5 \\
Basketball court & 6.8 & \underline{8.0} & 7.5 & 7.5 & 47.3\,{\scriptsize$\pm$3.5} & 52.5 & \textbf{52.4}\,{\scriptsize$\pm$1.4} & 55.1 & 57.5\,{\scriptsize$\pm$10.6} & 66.8 \\
Roundabout & 1.8 & \underline{19.8} & 15.1 & 15.1 & 30.5\,{\scriptsize$\pm$9.8} & 31.1 & \textbf{32.0}\,{\scriptsize$\pm$1.7} & 31.7 & 42.0\,{\scriptsize$\pm$5.2} & 57.0 \\
Soccer ball field & 14.1 & \underline{29.6} & 8.4 & 8.4 & \textbf{67.5}\,{\scriptsize$\pm$0.4} & 69.3 & 66.7\,{\scriptsize$\pm$2.9} & 67.5 & 77.6\,{\scriptsize$\pm$3.0} & 80.6 \\
Ground track field & \underline{27.1} & 23.3 & 0.5 & 0.5 & \textbf{43.6}\,{\scriptsize$\pm$2.3} & 44.4 & 30.1\,{\scriptsize$\pm$4.0} & 29.9 & 64.9\,{\scriptsize$\pm$1.7} & 67.9 \\
Small vehicle & \textbf{\underline{47.0}} & 44.4 & 44.2 & 44.2 & 35.0\,{\scriptsize$\pm$0.8} & 35.4 & 32.5\,{\scriptsize$\pm$0.7} & 32.9 & 54.1\,{\scriptsize$\pm$0.5} & 54.5 \\
Large vehicle & 41.3 & 59.4 & 59.4 & \textbf{\underline{59.5}} & 43.5\,{\scriptsize$\pm$0.8} & 44.3 & 39.4\,{\scriptsize$\pm$2.9} & 40.0 & 69.4\,{\scriptsize$\pm$0.5} & 70.1 \\
Bridge & 10.8 & 17.7 & \underline{17.8} & \underline{17.8} & 22.1\,{\scriptsize$\pm$1.5} & 23.9 & \textbf{22.2}\,{\scriptsize$\pm$1.3} & 23.4 & 48.3\,{\scriptsize$\pm$3.0} & 49.7 \\
Harbor & 8.8 & \underline{18.8} & 18.6 & 18.6 & \textbf{40.7}\,{\scriptsize$\pm$0.9} & 41.1 & 37.3\,{\scriptsize$\pm$0.8} & 37.4 & 63.4\,{\scriptsize$\pm$0.2} & 64.1 \\
Helicopter & \textbf{\underline{24.2}} & 20.1 & 20.1 & 20.1 & 17.2\,{\scriptsize$\pm$3.6} & 20.6 & 23.4\,{\scriptsize$\pm$2.6} & 25.7 & 48.2\,{\scriptsize$\pm$1.4} & 49.0 \\
\midrule
Category mean & 27.9 & 32.3 & 24.1 & 28.0 & \textbf{47.3} & 48.9 & 44.5 & 45.3 & 62.6 & 66.7 \\
\bottomrule
\end{tabular}

%% file: tables/supp_listing_knowledge.tex
\setlength{\aboverulesep}{0pt}\setlength{\belowrulesep}{0pt}\renewcommand{\arraystretch}{1.15}%
\begin{tabular}{@{}>{\raggedright\arraybackslash}p{\linewidth}@{}}
\toprule
\multicolumn{1}{@{}>{\columncolor{ruletitle}[0pt][0pt]}p{\linewidth}@{}}{\rule{0pt}{9.5pt}\hspace{4pt}\textbf{Category knowledge: basketball court}\rule[-3.5pt]{0pt}{0pt}} \\
\midrule
\addlinespace[3pt]
\textbf{What it is and how it looks from above.} A full court 28\,$\times$\,15 m (also 28.65\,$\times$\,15.24 m, school 25.6\,$\times$\,15.24 m), half courts 14\,$\times$\,15 m. Lines 5 cm in white or yellow: boundary, halfway line and centre circle (3.6 m), the keys (4.9\,$\times$\,5.8 m) with free-throw circles, three-point arcs, restricted arcs; backboards on posts at both ends with a short shadow. Surfaces: asphalt (black to faded grey, cracks, patches), concrete (light grey with joints), acrylic or rubber coatings in green, blue, terracotta, red, grey or purple, two-tone (keys, centre circle or the surround in a second colour), sport tiles in strong colours, worn paint. Overlaid lines of volleyball, badminton, futsal or netball on multi-use courts. \\
\addlinespace[3pt]
\textbf{Where it occurs.} School yards (banks of 2-12 courts side by side and end to end sharing a slab; the commonest), parks and housing estates (one or two courts with benches, fenced), residential compounds and campuses, beside tennis courts and tracks (also inside a D area), church and community-centre lots, courts painted on car parks, rooftops of schools and malls, driveways (a half court with one hoop), military bases, street-ball cages. \\
\addlinespace[3pt]
\textbf{Look-alikes that stay unlabelled.} Tennis courts (net and its shadow, service boxes, no arcs), volleyball and badminton courts, futsal and handball courts (20\,$\times$\,40 m, D-shaped goal areas), netball (thirds, semicircles), painted playground games, helipads, car-park bays, roller rinks. \\
\addlinespace[3pt]
\textbf{Label extent.} The playing rectangle of each court (boundary lines inside the mask), one instance per court, adjacent courts separate. \\
\addlinespace[2pt]
\bottomrule
\end{tabular}

%% file: supp/s9_limitations.tex
\section{Limitations}
\label{sec:s-limitations}

\paragraph{Attribution and authoring effort.}
\tld and \tlm are independent implementations that differ in rules, assets, rendering, viewing geometry, and pool composition (\cref{tab:s-pools,tab:s-implementation}); because \tld encapsulates object-level geometry in prepared 3D assets and asset-level Blender modules, whereas \tlm expresses both scene structure and object shapes through procedural rules (\cref{sec:s-pools}), comparing them under a shared rule set is not feasible.
Their comparison therefore shows that both realizations transfer rather than isolating a single factor; evidence on world rules and on single scene properties comes from the within-generator ablations and interventions of \cref{sec:s-atlas}, which are run on \tlm under one controlled training setup.
Throughput counts generation only: the one-time work of specifying and reviewing rules and cards and of preparing assets and textures is outside it (\cref{sec:s-cost}); the development time of a category card and its asset module is reported in \cref{sec:ondemand}.

\paragraph{Downstream models and remaining gaps.}
Transfer accuracy depends on the downstream model: ResNet backbones score substantially lower than pretrained foundation backbones, most of all in building change detection (\cref{tab:readers}); with lightweight change detectors, every synthetic source transfers at low accuracy, and the ranking of the sources varies with the detector: on the mean of LEVIR-CD, WHU-CD, and SECOND, \tlm alone falls below SyntheWorld and SynRS3D, whereas \tld and \tlmix remain above both.
For building instances, models trained on synthetic data produce more false positives than the real-label reference (\cref{fig:s-qual-inst}), and pseudo-label distillation with unlabeled target images does not raise instance AP (\cref{tab:main}).
In on-demand extension, a gap to full real supervision remains (47.3 against 62.6 mean IoU), the specialists trail the best open-vocabulary method on small vehicles, large vehicles, and helicopters, and panchromatic images are harder than color images both for the specialists (29.8 against 53.2 IoU) and for the best open-vocabulary method (15.3 against 32.8; \cref{sec:s-ondemand-results}).

\paragraph{Imaging scope.}
Both generators target very-high-resolution optical imagery from nadir to moderately oblique views; other modalities and strongly oblique views are outside their present scope.

\par